\documentclass{article}
\usepackage{ijcai23}
\usepackage{times}
\usepackage{soul}
\usepackage{url}
\usepackage[hidelinks]{hyperref}
\usepackage[utf8]{inputenc}
\usepackage[small]{caption}
\usepackage{graphicx}
\usepackage{amsmath}
\usepackage{amsthm}
\usepackage{booktabs}
\usepackage{algorithm}
\usepackage{algorithmic}
\usepackage{cleveref}
\usepackage{multirow}
\usepackage{color}
\usepackage{amssymb}
\usepackage{subcaption}

\usepackage{metawriting/metawriting}
\ours{SGAT4PASS}
\usepackage{metawriting/sciabbr}

\title{\ours: Spherical Geometry-Aware Transformer for PAnoramic Semantic Segmentation}

\begin{document}
\newcommand*\samethanks[1][\value{footnote}]{\footnotemark[#1]}

\author{Xuewei Li$^{1,2}$\thanks{The first two authors contributed equally to this paper.}, Tao Wu$^1$\samethanks[1], Zhongang Qi$^2$, Gaoang Wang$^3$, Ying Shan$^2$ \And Xi Li$^{1,4}$\thanks{Corresponding author.}
\affiliations
$^1$College of Computer Science and Technology, Zhejiang University
\\
$^2$ARC Lab, Tencent PCG
\\
$^3$Zhejiang University-University of Illinois at Urbana-Champaign Institute, Zhejiang University
\\
$^4$Zhejiang – Singapore Innovation and AI Joint Research Lab, Hangzhou
\\
\emails
\{xueweili, taowucs\}@zju.edu.cn, zhongangqi@tencent.com, gaoangwang@intl.zju.edu.cn, yingsshan@tencent.com, xilizju@zju.edu.cn
}
\maketitle

\begin{abstract}
  As an important and challenging problem in computer vision, PAnoramic Semantic Segmentation (PASS) gives complete scene perception based on an ultra-wide angle of view. 
  Usually, prevalent PASS methods with 2D panoramic image input focus on solving image distortions but lack consideration of the 3D properties of original $360^{\circ}$ data. 
  Therefore, their performance will drop a lot when inputting panoramic images with the 3D disturbance. 
  To be more robust to 3D disturbance, we propose our Spherical Geometry-Aware Transformer for PAnoramic Semantic Segmentation (\ours), considering 3D spherical geometry knowledge. 
  Specifically, a spherical geometry-aware framework is proposed for PASS. It includes three modules, i.e., spherical geometry-aware image projection, spherical deformable patch embedding, and a panorama-aware loss, which takes input images with 3D disturbance into account, adds a spherical geometry-aware constraint on the existing deformable patch embedding, and indicates the pixel density of original $360^{\circ}$ data, respectively.
  Experimental results on Stanford2D3D Panoramic datasets show that \ours significantly improves performance and robustness, with approximately a 2\% increase in mIoU, and when small 3D disturbances occur in the data, the stability of our performance is improved by an order of magnitude.
  Our code and supplementary material are available at \url{https://github.com/TencentARC/SGAT4PASS}.
\end{abstract}
\section{Introduction}
\label{sec:intro}
There has been a growing trend of practical applications based on $360^{\circ}$ cameras in recent years, including holistic sensing in autonomous vehicles~\cite{de2018eliminating,ma2021densepass,gao2022review,summaira2021recent,li2011graph,chen2014ranking,jiang2019learning}, immersive viewing in augmented reality and virtual reality devices~\cite{xu2018predicting,xu2021spherical,ai2022deep}, etc. 
Panoramic images with an ultra-wide angle of view deliver complete scene perception in many real-world scenarios, thus drawing increasing attention in the research community in computer vision. 
Panoramic semantic segmentation (PASS) is essential for omnidirectional scene understanding, as it gives pixel-wise analysis for panoramic images and offers a dense prediction technical route acquiring $360^{\circ}$ perception of surrounding scenes~\cite{yang2021context}.

\begin{figure}[tb]
    \centering
    \begin{subfigure}{0.48\linewidth}
        \centering
        \includegraphics[width=1\linewidth]{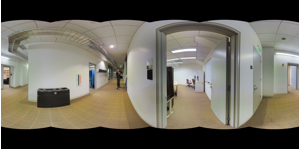}
        \caption{Original image}
        \label{sfig:original}
    \end{subfigure}
    \begin{subfigure}{0.48\linewidth}
        \centering
        \includegraphics[width=1\linewidth]{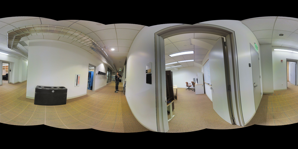}
        \caption{$5^{\circ}$ pitch rotation image}
        \label{sfig:y5}
    \end{subfigure}
    
    \begin{subfigure}{0.48\linewidth}
      \centering
      \includegraphics[width=1\linewidth]{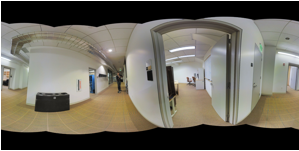}
      \caption{$5^{\circ}$ roll rotation image}
      \label{sfig:x5}
    \end{subfigure}
    \begin{subfigure}{0.48\linewidth}
        \centering
        \includegraphics[width=1\linewidth]{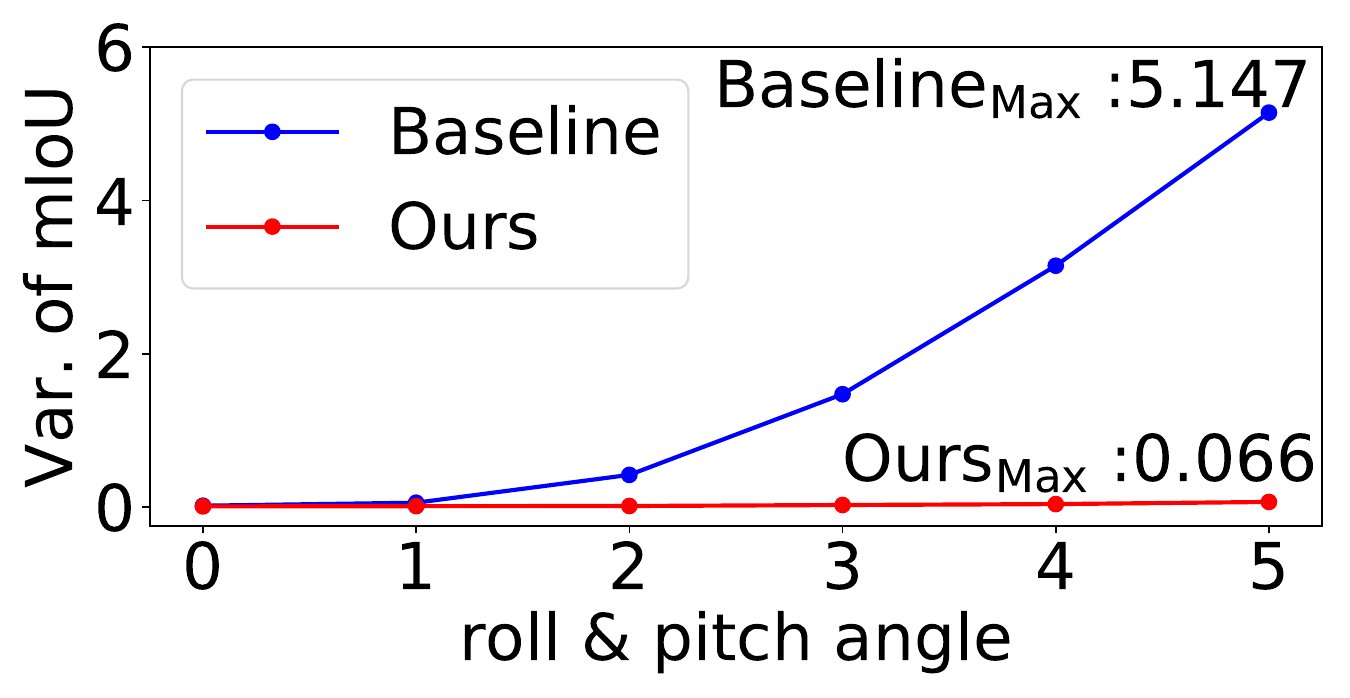}
        \caption{Angle \& Variance of mIoU}
        \label{sfig:first_var}
   \end{subfigure}
 
    \begin{subfigure}{0.48\linewidth}
        \centering
        \includegraphics[width=1\linewidth]{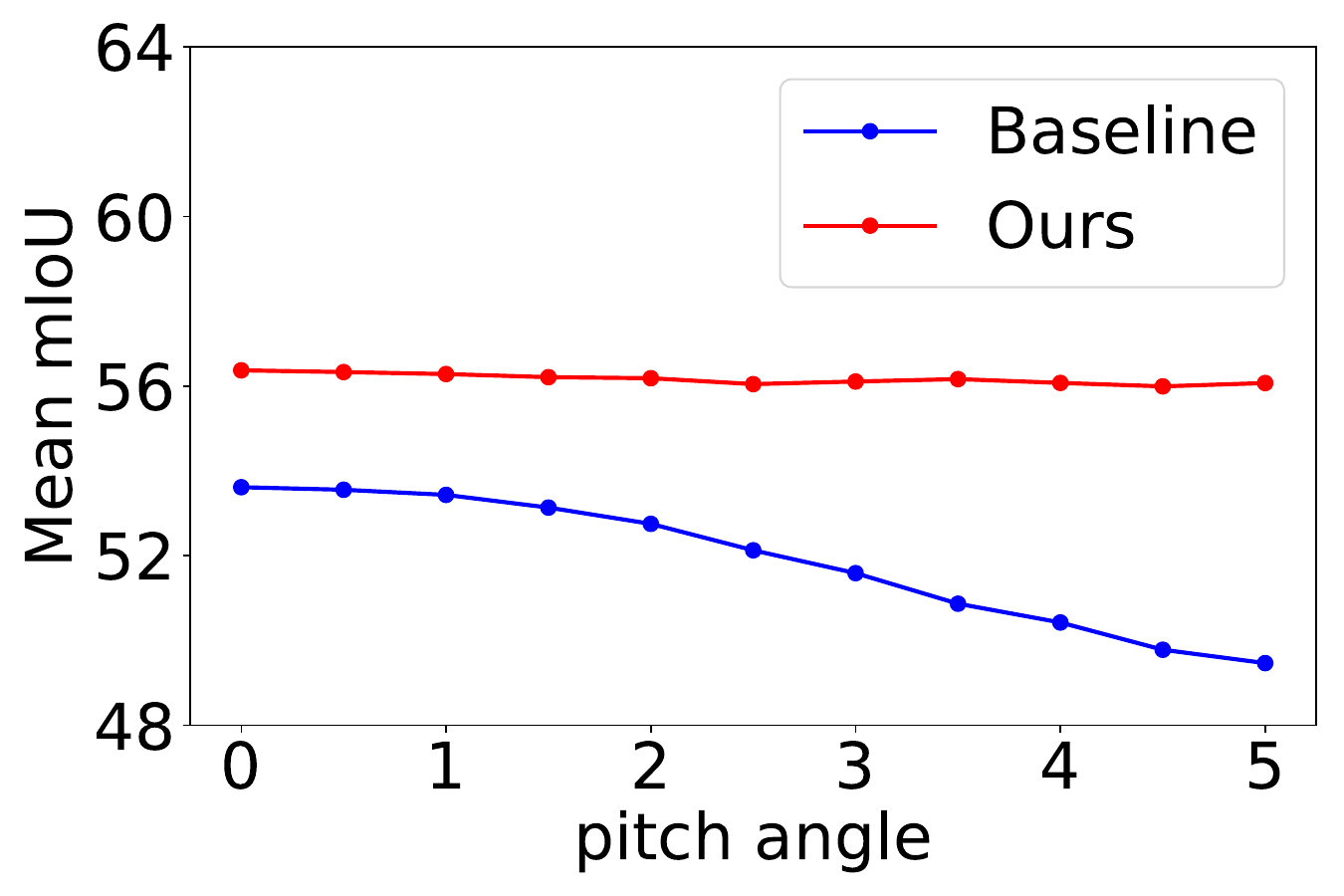}
        \caption{Pitch angle \& Mean mIoU}
        \label{sfig:pitch}
    \end{subfigure}
    \begin{subfigure}{0.48\linewidth}
        \centering
        \includegraphics[width=1\linewidth]{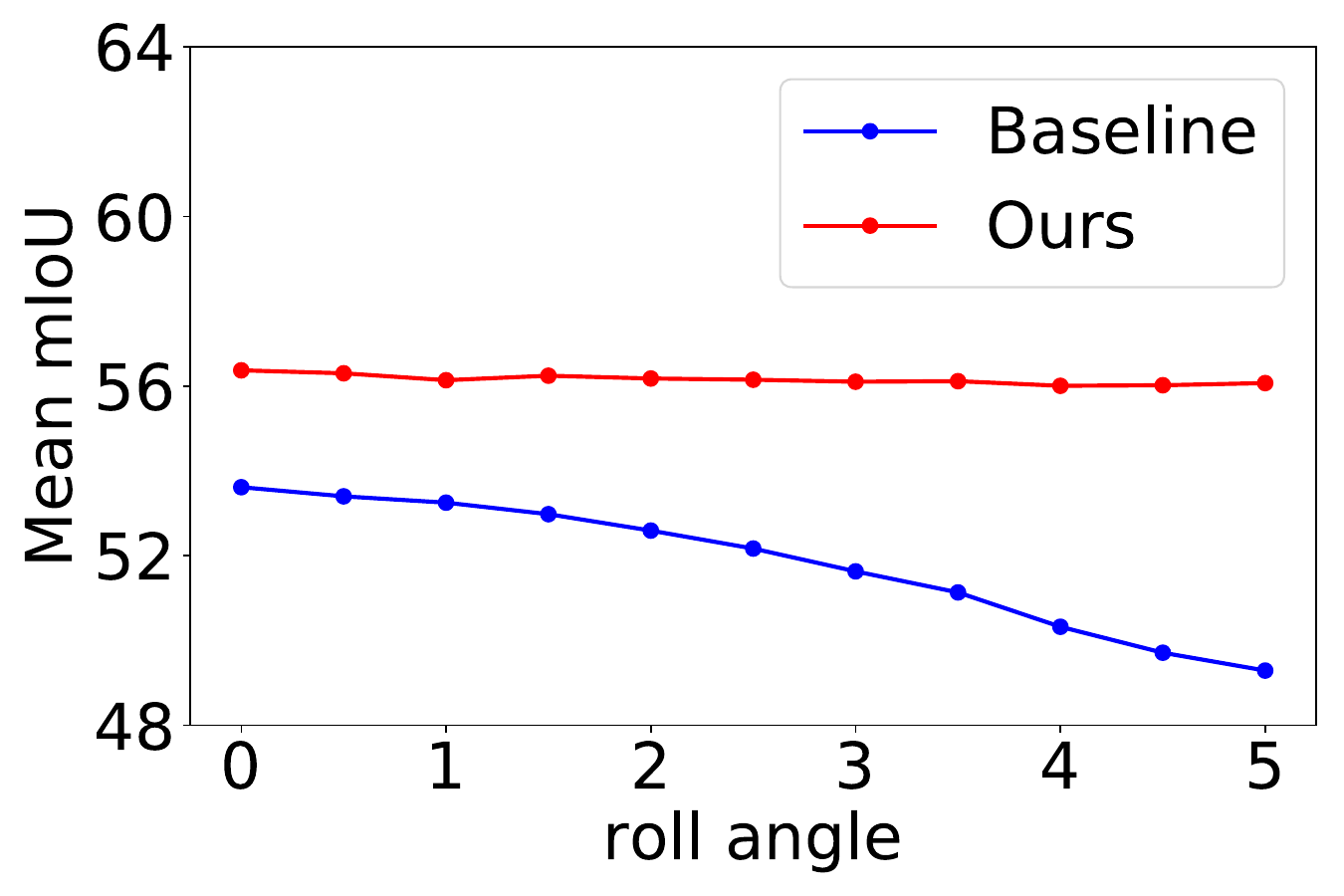}
        \caption{Roll angle \& Mean mIoU}
        \label{sfig:roll}
    \end{subfigure}
    \caption{The results with 3D disturbance input. 
    (a) is the original image, and (b) / (c) is the images rotated $5^{\circ}$ in pitch / roll axis. 
    Our baseline is Trans4PASS+. 
    Compared with the minor change in images, the huge variance / performance change in SGA validation is shown in (d) / (e) and (f). 
    ``Mean'' and ``Variance'' are defined in detail in~\Cref{ssec:datasets}. 
    }
    \label{fig:first_fig}
    \centering
\end{figure}
Most existing PASS approaches use equirectangular projection (ERP)~\cite{sun2021hohonet,yang2021capturing} to convert original $360^{\circ}$ data to 2D panoramic images. 
However, these methods often suffer from two main problems: large image distortions and lack of Spherical Geometry-Aware (SGA) robustness that resists 3D disturbance.
These problems lead neural networks to only learn suboptimal solutions for panoramic segmentation~\cite{yang2019pass,wang2021pyramid}. 
Although some recent works~\cite{zhang2022bending,zhang2022behind} take serious distortions into account in their models and become the current state-of-the-art (SOTA), they still do not pay enough attention to the SGA properties of the original $360^{\circ}$ data, resulting in performance degradation even with small projection disturbance.
As shown in~\Cref{sfig:y5} and~\Cref{sfig:x5}, applying $5^{\circ}$ rotation on the pitch or roll axis of original $360^{\circ}$ data carries only minor changes in 2D panoramic images. 
However, as shown in~\Cref{sfig:pitch}, \Cref{sfig:roll}, and~\Cref{sfig:first_var}, the performances of Trans4PASS+~\cite{zhang2022behind} (the blue lines) drop a lot (about 4\%), and the variance increases by almost 2 orders of magnitude, because the axis rotations lead to different spherical geometry relations between pixels in the projected panoramic images, which the existing methods fail to adapt. 
Besides disturbance, the ERP also introduces boundaries to panoramic images that the original $360^{\circ}$ data do not have. 
Some adjacent pixels are disconnected and some objects are separated, which is a severe issue, especially for semantic segmentation.
Furthermore, there also exists a difference in pixel sampling density between the original $360^{\circ}$ data and its corresponding projection image, e.g., pixels are over sampled in the antarctic and arctic areas of 2D panoramic images.
All these issues make panoramic semantic segmentation a challenging task, and the above characteristics should be well studied to design a robust model that adapts to disturbance, disconnection, uneven density, and other SGA properties. 

Improving robustness and taking SGA properties into account, we propose a novel model, i.e., Spherical Geometry-Aware Transformer for PAnoramic Semantic Segmentation (\ours), equipped with the SGA framework and SGA validation. 
The proposed SGA framework includes SGA image projection in the training process, Spherical Deformable Patch Embedding (SDPE), and a panorama-aware loss. 
SGA image projection provides images with 3D disturbance to improve the 3D robustness of the model. 
SDPE improves the patch embedding and makes it consider not only the image distortions with deformable operation but also spherical geometry with SGA intra- and inter-offset constraints. 
The panorama-aware loss deals with the difference in pixel density between the original $360^{\circ}$ data and its corresponding 2D panoramic images. 
Moreover, we propose a new validation method, i.e., SGA validation, to evaluate the 3D robustness of various models comprehensively, which considers different 3D disturbances for input images, and measures the average performance and the variance for comparisons.  
Extensive experimental results on popular Stanford2D3D panoramic datasets~\cite{armeni2017joint} demonstrate that our proposed approach achieves about 2\% and 6\% improvements on traditional metrics and SGA metrics, respectively.

The contributions of this paper are summarized as follows:
\begin{itemize}
    \item We propose \ours, a robustness model for the PASS task, which utilizes SGA image projection to deal with the 3D disturbance issue caused by ERP.
    \item We introduce SDPE to combine spherical geometry with deformable operation to better deal with panoramic image distortion. And we also propose panorama-aware loss to ease the oversampling problem.
    \item We evaluate \ours on the popular benchmark and perform extensive experiments with both traditional metrics and proposed SGA metrics, which demonstrate the effectiveness of each part of the framework.  
\end{itemize}
\section{Related Work}
\label{sec:related}

The two most related fields are panoramic semantic segmentation and dynamic and deformable vision transformers. 

\begin{figure*}[tb]
    \centering
    \includegraphics[width=0.9\linewidth]{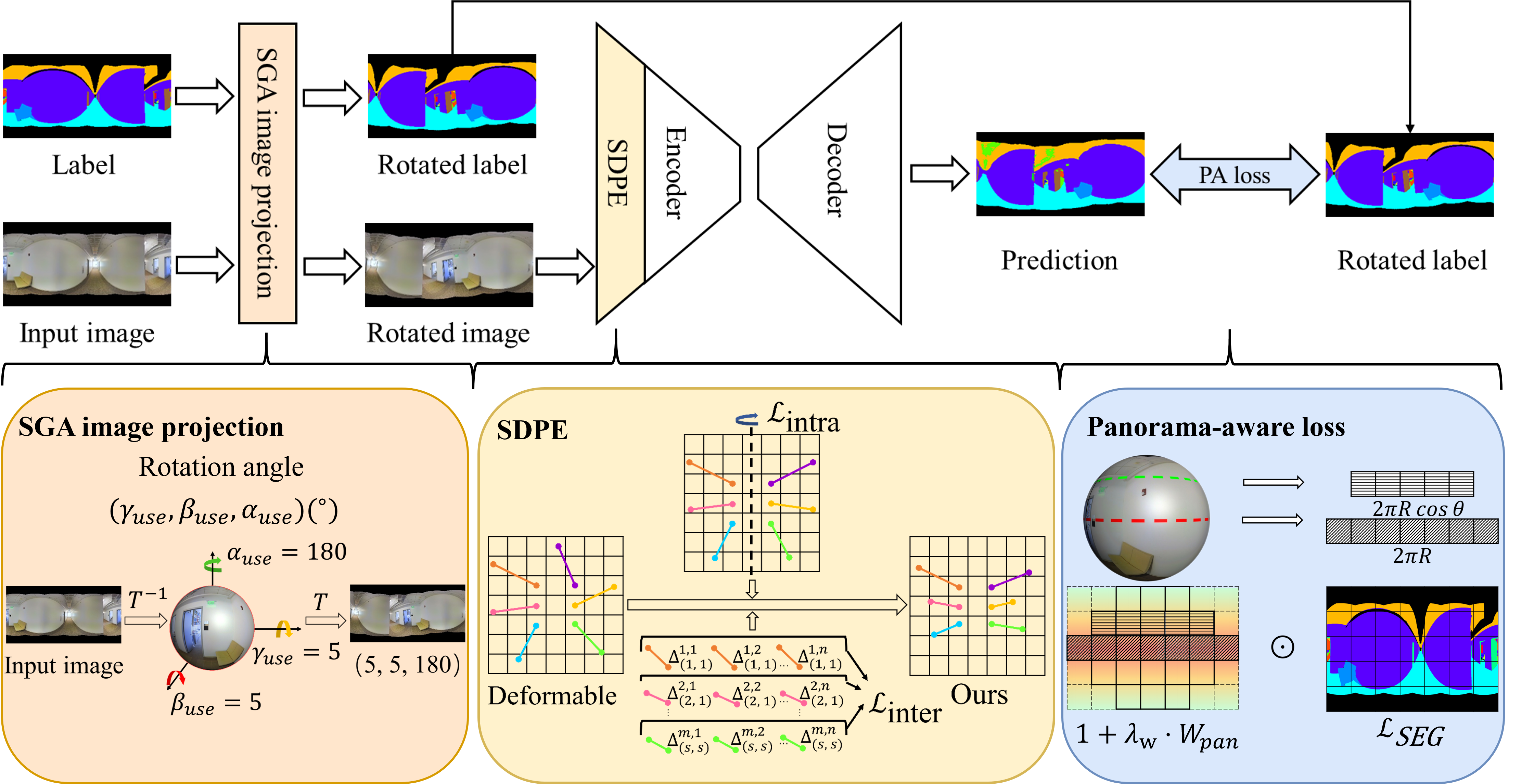}
    \caption{Overall review of \ours. 
    We borrow the network from Trans4PASS+, and add three main modules: Spherical geometry-aware (SGA) image projection, SDPE, and panorama-aware loss. 
    (Lower left) SGA image projection rotates the input panoramic images to mimic 3D disturbance. 
    (Lower middle) SDPE adds several SGA constraints on deformable patch embedding and let it consider both image distortions and spherical geometry. 
    (Lower right) Panorama-aware loss (PA loss) takes into account the pixel density of a sphere. 
    }
    \label{fig:overall}
    \centering
\end{figure*}

\subsection{Panoramic Semantic Segmentation}
\label{ssec:PASS}
Semantic segmentation of panoramic images has many applications in real-world scenarios, such as autonomous driving~\cite{ye2021ghost}, panoramic lenses safety and monitoring applications~\cite{poulin2012optical}, etc.
With the development of deep learning, many neural networks have been developed for panoramic semantic segmentation.
Deng et al.~\cite{deng2017cnn} first proposed a semantic segmentation framework for wide-angle (fish-eye) images and transformed an existing pinhole urban scene segmentation dataset into synthetic datasets. 
Yang et al.~\cite{yang2019pass} designed a semantic segmentation framework for panoramic annular images using a panoramic annular camera with an entire Field of View (FoV) for panoramic surrounding perception based on a single camera. 
Furthermore, Yang et al.~\cite{yang2020ds} proposed DS-PASS to improve it with a more efficient segmentation model with attention connections.
PASS solutions can be divided into two main fields: distortion-aware strategies and 2D-geometry-aware ones. 

For distortion-aware strategies, Tateno et al.~\cite{tateno2018distortion} proposed using specially designed distortion-aware convolutions in a fixed manner to address image distortions.
Furthermore, ACDNet~\cite{zhuang2022acdnet} combined convolution kernels with different dilation rates adaptively and used fusion-driven methods to take advantage of several projections. 
Jiang et al.~\cite{jiang2019spherical} designed a spherical convolution operation. 
Lee et al.~\cite{lee2018spherephd} used spherical polyhedrons to represent panoramic views to minimize the difference in spatial resolution of the surface of the sphere and proposed new convolution and grouping methods for the representation of spherical polyhedrons.
Hu et al.~\cite{hu2022distortion} designed and proposed a distortion convolutional module based on the image principle to solve the distortion problem caused by the distortion of the panoramic image.
Zhang et al.~\cite{zhang2022bending}~\cite{zhang2022behind} designed their Trans4PASS and Trans4PASS+ that perceived spherical distortion and solved the distortion problem of spherical images better through their Deformable Patch Embedding (DPE) and Deformable Multi-Layer Perception (DMLP) modules.
Also, Trans4PASS+ is the current SOTA panoramic semantic segmentation model and is our baseline. 
For 2D geometry-aware strategies, horizontal features are mainly used based on the ERP inherent property. 
Sun et al.~\cite{sun2021hohonet} proposed HoHoNet and Pintore et al.~\cite{pintore2021slicenet} proposed SliceNet to use the extracted feature maps in a 1D horizontal representation.

For our \ours based on the distortion-aware SOTA model, Trans4PASS+, we add SGA information from the original $360^{\circ}$ data instead of the 2D geometry prior to panoramic images to improve not only its performance but also its robustness when meeting 3D disturbance. 

\subsection{Dynamic and Deformable Vision Transformers}
\label{ssec:DDVT}
Regarding the field of vision transformers, some works have developed architectures with dynamic properties. 
Chen et al.~\cite{chen2021dpt} and Xia et al.~\cite{xia2022vision} used deformable designs in later stages of the encoder. 
Yue et al.~\cite{yue2021vision} used a progressive sampling strategy to locate discriminatory regions. 
Deformable DETR~\cite{zhu2020deformable} used deformable attention to deal with feature maps.
Some other works used adaptive optimization of the number of informative tokens to improve efficiency~\cite{wang2021not}~\cite{rao2021dynamicvit}~\cite{yin2022vit}~\cite{xu2022evo}.  
Zhang et al.~\cite{zhang2022bending}~\cite{zhang2022behind} designed their Trans4PASS and Trans4PASS+ based on DPE and Deformable Multi-Layer Perception (DMLP) modules, and we use Trans4PASS+ as our baseline. 
\section{Method}
\label{method}
We present Spherical Geometry-Aware Transformer for PAnoramic Semantic Segmentation (\ours) in this section.
First, we introduce the background of panoramic semantic segmentation in~\Cref{ssec:MPASS}. 
Second, we describe our main idea to apply different SGA properties in panoramic semantic segmentation in \Cref{ssec:SGA_framework}. 
To improve the 3D robustness of \ours, we propose SGA Image Projection, Spherical Deformable Patch Embedding (SDPE), and panorama-aware loss. 
Specifically, SGA Image Projection adds rotated samples in training; 
SDPE adds SGA constraints on the deformable patch embedding; 
and the panorama-aware loss fuses sphere pixel density to training process. 

\subsection{Background}
\label{ssec:MPASS}
We first describe a general formulation of PASS and then introduce the spherical geometry property that we focus mainly on. Panoramic images are based on original $360^{\circ}$ data formulated in the spherical coordinate system (based on longitude and latitude). 
To convert it to a rectangular image in a Cartesian coordinate system, ERP is a widely used projection in this field: $x = (\theta - \theta_0) cos \phi_1, y = (\phi-\phi_1)$,
where $\theta_0 = 0$ is the central latitude and $\phi_1 = 0$ is the central longitude. 
The ERP-processed rectangular images are used as the input sample in datasets and fed to the neural network, and the rectangular semantic segmentation results are obtained to compare with the ground truth and calculate the metrics. 
Although traditional methods can treat PASS as the conventional 2D semantic segmentation task and deal with panoramic images easily, the spherical geometry property is partly ignored.

\begin{table}[]
    \footnotesize
    \centering
    \resizebox{0.42\textwidth}{!}{
    \begin{tabular}{lcc}
    \toprule
    Method                                        & Avg mIoU      & F1 mIoU       \\
    \midrule
    StdConv~\cite{tateno2018distortion}           &  -            & 32.6          \\
    CubeMap~\cite{tateno2018distortion}           &  -            & 33.8          \\
    DistConv~\cite{tateno2018distortion}          &  -            & 34.6          \\	
    SWSCNN~\cite{esteves2020spin}                 & 43.4          &  -            \\
    Tangent (ResNet-101)~\cite{tangent}           & 45.6          &  -            \\
    FreDSNet~\cite{berenguel2022fredsnet}         & -             & 46.1          \\
    PanoFormer~\cite{shen2022panoformer}           & 48.9          & -             \\
    HoHoNet (ResNet-101)~\cite{sun2021hohonet}    & 52.0          & 53.9          \\
    Trans4PASS (Small)~\cite{zhang2022bending}    & 52.1          & 53.3          \\
    CBFC~\cite{zheng2023complementary}            & 52.2          & -             \\
    Trans4PASS+ (Small)~\cite{zhang2022behind}    & 53.7          & 53.6          \\
    \textbf{Ours (Small)}                         & \textbf{55.3} & \textbf{56.4} \\
    \bottomrule
    \end{tabular}
    }
    \caption{Comparison with the SOTA methods on Stanford2D3D Panoramic datasets. 
    We follow recent works to compare the performance of both official fold 1 and the average performance of all three official folds. respectively. 
    ``Avg mIoU'' / ``F1 mIoU'' means the mIoU performance of three official folds on average / official fold 1. 
    A considerable improvement is gained. 
    }
    \label{tab:sota}
\end{table}

\subsection{Spherical Geometry-Aware (SGA) Framework}
\label{ssec:SGA_framework}
We propose the SGA framework for PASS with SGA image projection, SDPE, and panorama-aware loss. 
To deal with the inevitable 3D disturbance during the acquisition of the input image, our SGA image projection aims to encode the original $360^{\circ}$ data spherical geometry by generating input images with different rotations. 
We design SDPE to model spatial dependencies on a sphere, making patch embedding consider both spherical geometry and image distortions.
Furthermore, a panorama-aware loss is proposed to model the pixel density of a sphere, making the loss weight distribution more similar to the original $360^{\circ}$ data. 
With these three modules, the spherical geometry is well employed in the PASS task.

\subsubsection{Spherical Geometry-Aware (SGA) Image Projection}
The original $360^{\circ}$ data follow a spherical distribution and are \textit{spherically symmetric}.
After rotating any angle along the yaw / pitch / roll axis, the transformed data are still equivalent to the original data.
Traditional strategies assume that the images are taken with the yaw / pitch / roll angle equal to zero degrees, which is too ideal in real-world scenarios and ignores the camera disturbance and random noise. When the rotation angle is disturbed, traditional strategies usually have a large degradation in the PASS task.
SGA image projection fuses this property between the inevitable equirectangular projection and regular image augmentation to make models robust to 3D disturbance.   

We use $T$ to represent the forward process of ERP transformation, which is the process of converting spherical coordinates to plane coordinates, and use $T^{-1}$ to represent the inverse process of ERP that transforms the plane back onto the sphere. 
Given an ERP-processed input panoramic image, we first transform the image $I$ originally in plane coordinates to spherical coordinates through the inverse ERP process. 
After that, we use the rotation matrix in the three-dimensional (3D) space to perform a 3D rotation in the spherical coordinate system. For a general rotation in a 3D space, the angles of yaw, pitch, and roll are $\alpha_{use}$, $\beta_{use}$, and $\gamma_{use}$, respectively.
The corresponding rotation matrix is $R(\alpha_{\mathrm{use}}, \beta_{\mathrm{use}}, \gamma_{\mathrm{use}})$. 
We multiply $R$ by the data in the spherical coordinate system to obtain the rotated data in the spherical coordinate system. 
Finally, we use the ERP forward process to convert the rotated spherical coordinate system image into a panoramic image, thus obtaining a certain rotated image of the real input of the network.
The corresponding point in input image of a pixel in rotated image may not have integer coordinates, and we select the nearest pixel as its corresponding pixel to be generic to the ground truth transformation. 
Based on these operations, we build our SGA image projection, $O_{3D}(I, \alpha_{\mathrm{use}}, \beta_{\mathrm{use}}, \gamma_{\mathrm{use}}) = T(R(\alpha_{\mathrm{use}}, \beta_{\mathrm{use}}, \gamma_{\mathrm{use}}) \cdot T^{-1}(I))$. (See Section C ``Details for SGA Image Projection'' in the supplementary material for details.)
At the beginning of the training process, we set the maximum rotation angle of the yaw / pitch / roll axis at $(\alpha_{\mathrm{train}}, \beta_{\mathrm{train}}, \gamma_{\mathrm{train}})$. 

\subsubsection{SDPE: Spherical Deformable Patch Embedding}
We first introduce DPE, and then fuse spherical geometry into DPE by SGA constraints to earn SDPE. 

Faced with image distortions in panoramic images, DPE, considering different distortions in different regions of an input panoramic image, is a popular solution~\cite{zhang2022bending}~\cite{zhang2022behind}. 
In detail, given a 2D input panoramic image, the standard patch embedding handles it into flattened patches $H \times W$, and the resolution of each patch is $(s,s)$. 
A learnable projection layer transforms each patch into out-dimensional embeddings. 
For each patch, the offsets $\Delta_{(i,j)}^{DPE}$ of the $i^{th}$ row $j^{th}$ column pixel are defined as: 
\begin{equation}
  \label{eq:DPE}
  \Delta_{(i,j)}^{DPE}={\begin{bmatrix}
  \mathrm{min}(\mathrm{max}(\text{-}k_{D} \cdot H, g(f)_{(i,j)}),k_{D} \cdot H)\\
  \mathrm{min}(\mathrm{max}(\text{-}k_{D} \cdot W, g(f)_{(i,j)}),k_{D} \cdot W)
  \end{bmatrix}}, 
\end{equation}
where $g(\cdot)$ is the offset prediction function. 
Hyperparameter $k_{D}$ puts an upper bound on the learnable offsets $\Delta_{(i,j)}^{DPE}$.  
For implementation, the deformable convolution operation~\cite{dai2017deformable} is popularly employed to realize DPE. 

\renewcommand{\arraystretch}{1.5}
\begin{table*}
    \footnotesize
    \centering
    \resizebox{0.95\textwidth}{!}{
        \begin{tabular}{c|c|c|c|c|c|c|c}
        \toprule
        \multirow{2}{*}{($\beta$,$\gamma$,$\alpha$) ($^{\circ}$)} & BL mIoU / PAcc   & \multirow{2}{*}{($\beta$,$\gamma$,$\alpha$) ($^{\circ}$)} & BL mIoU / PAcc   & \multirow{2}{*}{($\beta$,$\gamma$,$\alpha$) ($^{\circ}$)} & BL mIoU / PAcc   & \multirow{2}{*}{($\beta$,$\gamma$,$\alpha$) ($^{\circ}$)} & BL mIoU / PAcc   \\ \cline{2-2} \cline{4-4} \cline{6-6} \cline{8-8} 
                                                 & Our mIoU / PAcc &                                              & Our mIoU / PAcc &                                              & Our mIoU / PAcc &                                              & Our mIoU / PAcc \\ \midrule
        \multirow{2}{*}{(0,0,0)}                     & 53.617 / 81.483         & \multirow{2}{*}{(0,5,0)}                     & 49.292 / 78.346         & \multirow{2}{*}{(5,0,0)}                     & 49.468 / 78.500           & \multirow{2}{*}{(5,5,0)}                     & 47.234 / 77.129         \\ \cline{2-2} \cline{4-4} \cline{6-6} \cline{8-8} 
                                                    & 56.374 / 83.135         &                                              & 56.073 / 82.892         &                                              & 56.074 / 82.905         &                                              & 55.784 / 82.794         \\ \hline
        \multirow{2}{*}{(0,0,90)}                    & 53.918 / 81.590          & \multirow{2}{*}{(0,5,90)}                    & 49.861 / 78.656         & \multirow{2}{*}{(5,0,90)}                    & 49.400 / 78.373           & \multirow{2}{*}{(5,5,90)}                    & 47.589 / 77.361         \\ \cline{2-2} \cline{4-4} \cline{6-6} \cline{8-8} 
                                                    & 56.441 / 83.130          &                                              & 55.954 / 82.847         &                                              & 56.128 / 82.895         &                                              & 55.636 / 82.657         \\ \hline
        \multirow{2}{*}{(0,0,180)}                   & 53.587 / 81.476         & \multirow{2}{*}{(0,5,180)}                   & 49.344 / 78.532         & \multirow{2}{*}{(5,0,180)}                   & 49.536 / 78.585         & \multirow{2}{*}{(5,5,180)}                   & 47.458 / 77.307         \\ \cline{2-2} \cline{4-4} \cline{6-6} \cline{8-8} 
                                                    & 56.246 / 83.054         &                                              & 55.951 / 82.906         &                                              & 55.714 / 82.796         &                                              & 55.501 / 82.750          \\ \hline
        \multirow{2}{*}{(0,0,270)}                   & 53.669 / 81.459         & \multirow{2}{*}{(0,5,270)}                   & 49.462 / 78.445         & \multirow{2}{*}{(5,0,270)}                   & 49.363 / 78.485         & \multirow{2}{*}{(5,5,270)}                   & 47.726 / 77.451         \\ \cline{2-2} \cline{4-4} \cline{6-6} \cline{8-8} 
                                                    & 56.223 / 83.051         &                                              & 55.924 / 82.779         &                                              & 55.983 / 82.904         &                                              & 55.732 / 82.701         \\ \bottomrule
        \end{tabular}
    }
    \caption{
    Detail performance comparison with Tran4PASS+ on Stanford2D3D Panoramic datasets official fold 1 with SGA metrics. 
    All 18 situations are shown, and the analysis is in~\cref{tab:STA_SGAM_BIG}. 
    ``BL'' means the baseline, i.e., Tran4PASS+. 
    ``PAcc'' meas the pixel accuracy metric. 
    }
    \label{tab:SGAMBIG}
\end{table*}

When fusing spherical geometry into DPE, human photographic and ERP priors are taken into consideration, in which the plane formed by pitch and roll axes is always parallel to the ground plane and the projection cylinder is perpendicular to the ground plane. As a result, we add SGA constraints mainly on the yaw axis. 
In detail, we give intra-offset and inter-offset constraints on $\Delta_{(i,j)}^{DPE}$. 
For convenience, we use $\Delta_{(i,j)}^{m,n}$ to represent the $i^{th}$ row $j^{th}$ column pixel of the learnable offset for the $m^{th}$ row $n^{th}$ column patch.

\begin{table}
    \footnotesize
    \centering
    \resizebox{0.42\textwidth}{!}{
        \begin{tabular}{ccccc}
            \toprule
            \multirow{2}{*}{Statistics} & \multicolumn{2}{c}{Baseline} & \multicolumn{2}{c}{Ours} \\
                                        & mIoU        & PAcc      & mIoU      & PAcc    \\ \midrule
            Mean                        & 50.033      & 78.949         & 55.984 ({\color{red}{+5.951}})    & 82.887 ({\color{red}{+3.938}}) \\
            Variance                    & 5.147       & 2.413         & 0.066 ({\color{red}{-5.081}})     & 0.020 ({\color{red}{-2.393}})  \\
            Range                       & 6.684       & 4.461          & 0.940 ({\color{red}{-5.744}})     & 0.478 ({\color{red}{-3.983}})  \\ 
            \bottomrule
        \end{tabular}
    }
    \caption{
    Overall performance comparison with Tran4PASS+ on Stanford2D3D Panoramic datasets in~\cref{tab:SGAMBIG} setting. 
    ``PAcc'' means the pixel accuracy metric. 
    \ours earns considerable mean performance and significant robustness improvement.  
    }
    \label{tab:STA_SGAM_BIG}
\end{table}
\paragraph{Intra-offset constraint.}
Based on the phenomenon that the original $360^{\circ}$ data are symmetric on any longitude and the projection cylinder in ERP is symmetric in any line perpendicular to the base of the cylinder, the offset of any pixel in 2D input panoramic image $I$ should be symmetric on its perpendicular. 
To be generic to the learnable offsets $\Delta_{(i,j)}^{m, n}$ dealing with the image distortions, we use a constraint $\mathcal{L}_{intra}$: 
\begin{equation}
    \label{eq:intra}
      \mathcal{L}_{intra} = \sum_{m,n} \sum_{i,j} L^{intra}_{2}(\Delta_{(i,j)}^{m,n}, {\Delta_S}_{(i,j)}^{m,n}), 
\end{equation}
where ${\Delta_S}_{(i,j)}^{m,n}$ is the single patch offset that is formed symmetrically along the yaw axis with $\Delta_{(i,j)}^{m,n}$ as the template.
$L^{intra}_{2}(\cdot, \cdot)$ represents the element-wise L2 loss. 
\paragraph{Inter-offset constraint.}
Based on the phenomenon that the projection cylinder in ERP can be slit and expanded from any line perpendicular to the base of the cylinder, the offset of any pixel in 2D input panoramic image $I$ corresponding to the same latitude of the original $360^{\circ}$ data should be similar.  
To be generic to the learnable $\Delta_{(i,j)}^{DPE}$ dealing with the image distortions, we use a constraint, $\mathcal{L}_{inter}$, to model this property. 
For a certain pixel, we use the average offset in the whole horizontal line as its constraint: 
\begin{equation}
  \label{eq:inter}
  \mathcal{L}_{inter} = \sum_{m,n} \sum_{i,j} L_{2}^{inter}(\Delta_{(i,j)}^{m,n} , \Delta_{(i,j)}^{m,\mathrm{AVG}}), 
\end{equation}
where $\Delta_{(i,j)}^{m,\mathrm{AVG}}$ is the average of each component in $\{\Delta_{(i,j)}^{m,n}, n \in W\}$, and $L_{2}^{inter}(\cdot, \cdot)$ represents the L2 loss for each component length of the two vectors. 
Then the total SDPE loss is: $\mathcal{L}_{SDPE} = \mathcal{L}_{inter} + \mathcal{L}_{intra}$. 

\subsubsection{Panorama-Aware Loss} 
Because the panoramic images are rectangular in shape, the region of the antarctic and arctic areas in the original $360^{\circ}$ data is over sampled than the one near the equator. 
However, due to human photographic priors, the semantics of the antarctic (ground, floor, etc.) and arctic areas (sky, ceiling, etc.) are relatively simple, as seen in the sample images of~\Cref{fig:first_fig} and~\Cref{fig:overall}. 
When using traditional segmentation loss for supervised training, we treat each pixel equally, which leads to models paying relatively less attention to semantic rich regions near the equator.
To deal with this phenomenon, we design our panorama-aware loss.
For an ERP-processed panoramic image, the number of pixels in each horizontal line is the same, but the corresponding resolution density on the original sphere of each horizontal line is very different. 
For this reason, we design a loss to reweight the loss proportion of different horizontal lines depending on its height. 
For a pixel $(m,n)|m \in [1,H_I], n \in [1,W_I]$ ($W_I$ and $H_I$ are the width and height of the input image), we give a weight $w_{\mathrm{pan}}^{(m,n)}$ when calculating its per pixel loss:
\begin{equation}
  \label{eq:PPlossweight}
  w_{\mathrm{pan}}^{(m,n)} = \mathrm{cos}(\frac{|2m-H_I|}{H_I} \cdot \frac{\pi}{2}). 
  \end{equation}
We use $W_{\mathrm{pan}}$ to represent the set that includes all $w_{\mathrm{pan}}^{(m,n)}$. 

\begin{table}[]
    \footnotesize
    \centering
    \resizebox{0.42\textwidth}{!}{
    \begin{tabular}{ccccc}
    \toprule
    SGAIP        & SDPE         & PA            & mIoU   & Pixel accuracy \\
    \midrule
                 &              &               & 53.617 & 81.483      \\
    $\checkmark$ &              &               & 54.637 & 82.303      \\
                 & $\checkmark$ &               & 54.554 & 81.508      \\
                 &              &  $\checkmark$ & 54.833 & 81.733      \\  
    $\checkmark$ & $\checkmark$ &  $\checkmark$ & 56.374 & 83.135      \\
    \bottomrule
    \end{tabular}
    }
    \caption{Effect of each \ours module. 
    We validate them on Stanford2D3D Panoramic datasets official fold 1 with traditional metrics.  
    ``SGAIP'' / ``SDPE'' / ``PA'' means our SGA image projection / spherical deformable patch embedding / panorama-aware loss. 
    Using anyone, an average improvement of 1.058\% mIoU / 0.365\% pixel accuracy is gained when using three gains 2.757\% / 1.652\%. }
    \label{tab:ablation}
\end{table}

When faced with a panoramic semantic segmentation problem, we first estimate the usage scenario to determine $\beta$ and $\gamma$ used in SGA image projection when $\alpha$ is often set as $360^{\circ}$ in common condition. 
We set our total loss as:
\begin{equation}
  \label{eq:allloss}
  \mathcal{L}_{\mathrm{all}} =(1 + \lambda_w \cdot W_{\mathrm{pan}}) \odot \mathcal{L}_{SEG} + \lambda_s \cdot \mathcal{L}_{SDPE},  
  \end{equation}
where $\mathcal{L}_{SEG}$ is the common per pixel loss for semantic segmentation, $\odot$ is the element-wise matrix multiplication, $\lambda_w$ and $\lambda_s$ are hyperparameters.

\begin{figure*}[tb]
    \centering
    \begin{subfigure}{0.23\linewidth}
        \centering
        \includegraphics[width=1\linewidth]{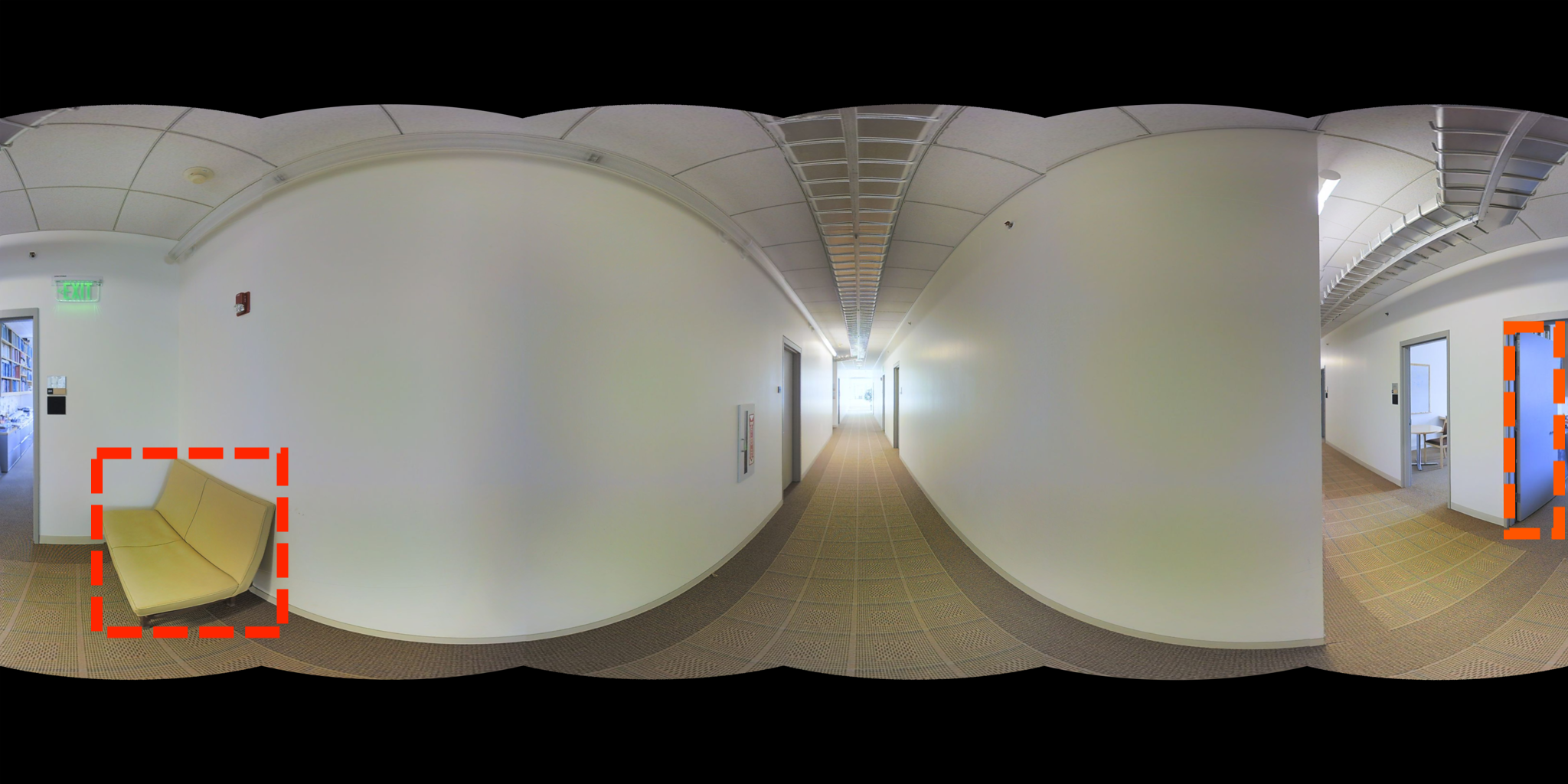}
        \caption{Original picture}
        \label{sfig:source}
    \end{subfigure}
    \begin{subfigure}{0.23\linewidth}
        \centering
        \includegraphics[width=1\linewidth]{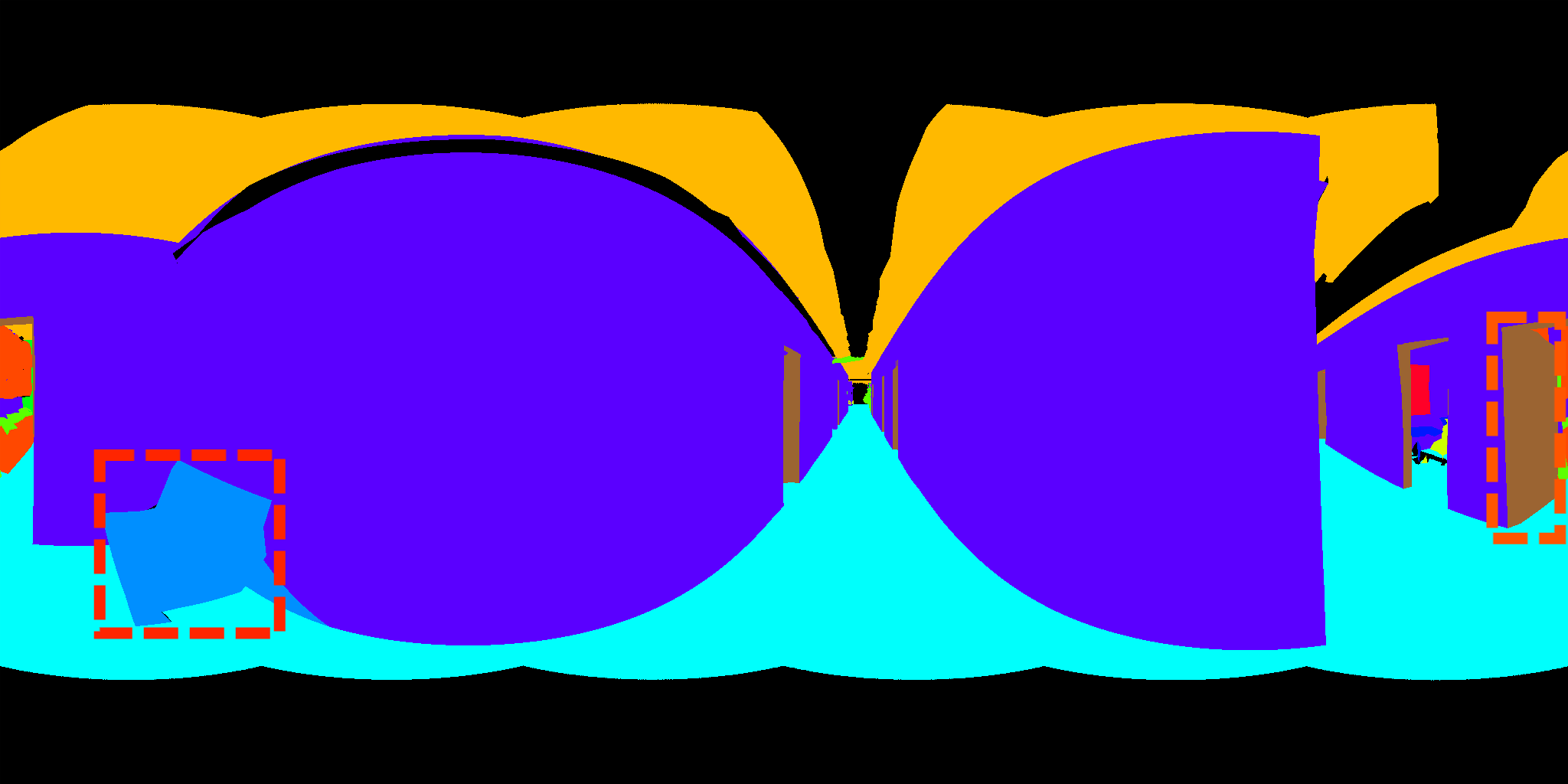}
        \caption{Label}
        \label{sfig:source_GT}
    \end{subfigure}
    \begin{subfigure}{0.23\linewidth}
        \centering
        \includegraphics[width=1\linewidth]{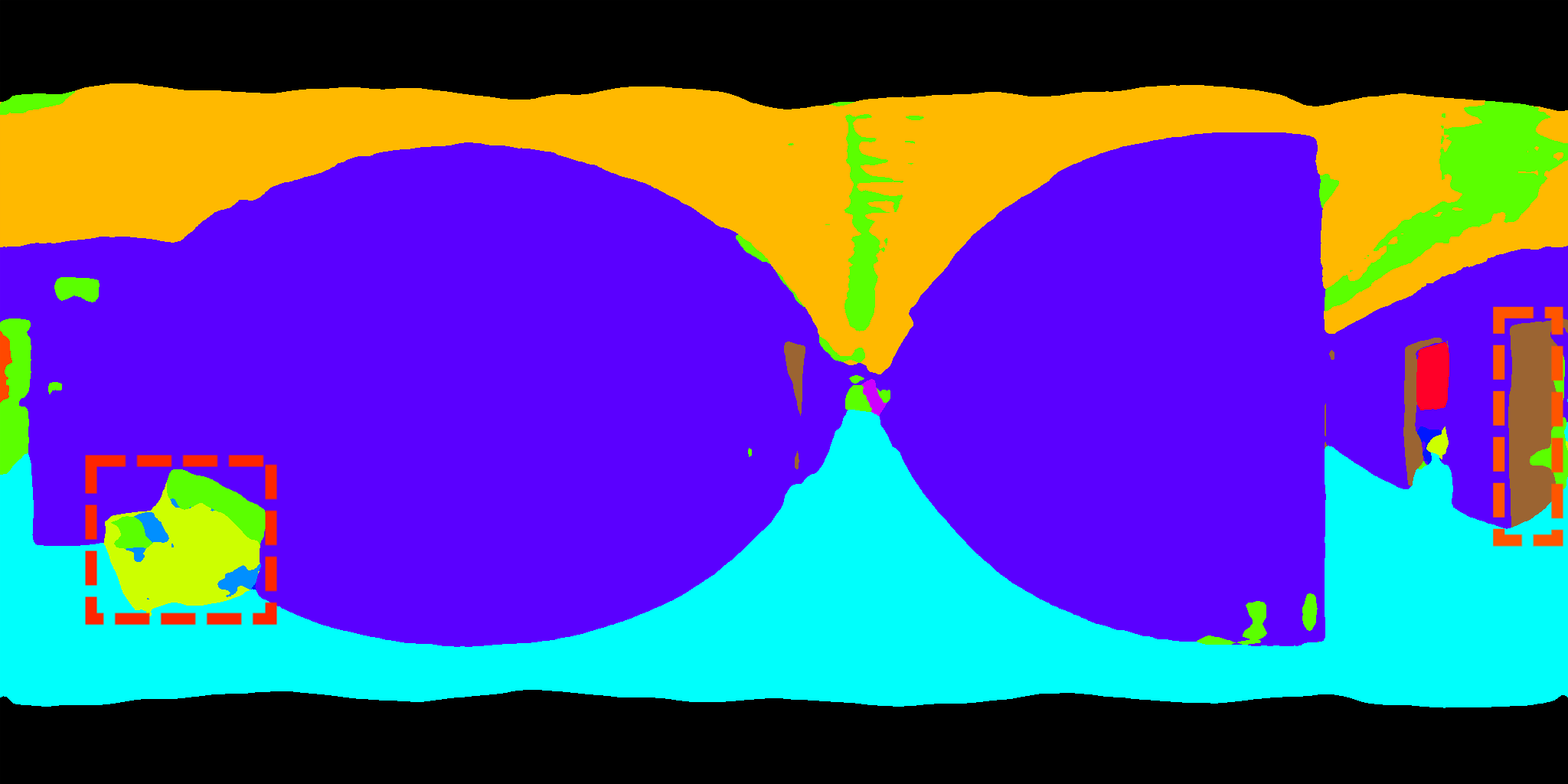}
        \caption{Baseline results}
        \label{sfig:source_BL}
    \end{subfigure}
    \begin{subfigure}{0.23\linewidth}
        \centering
        \includegraphics[width=1\linewidth]{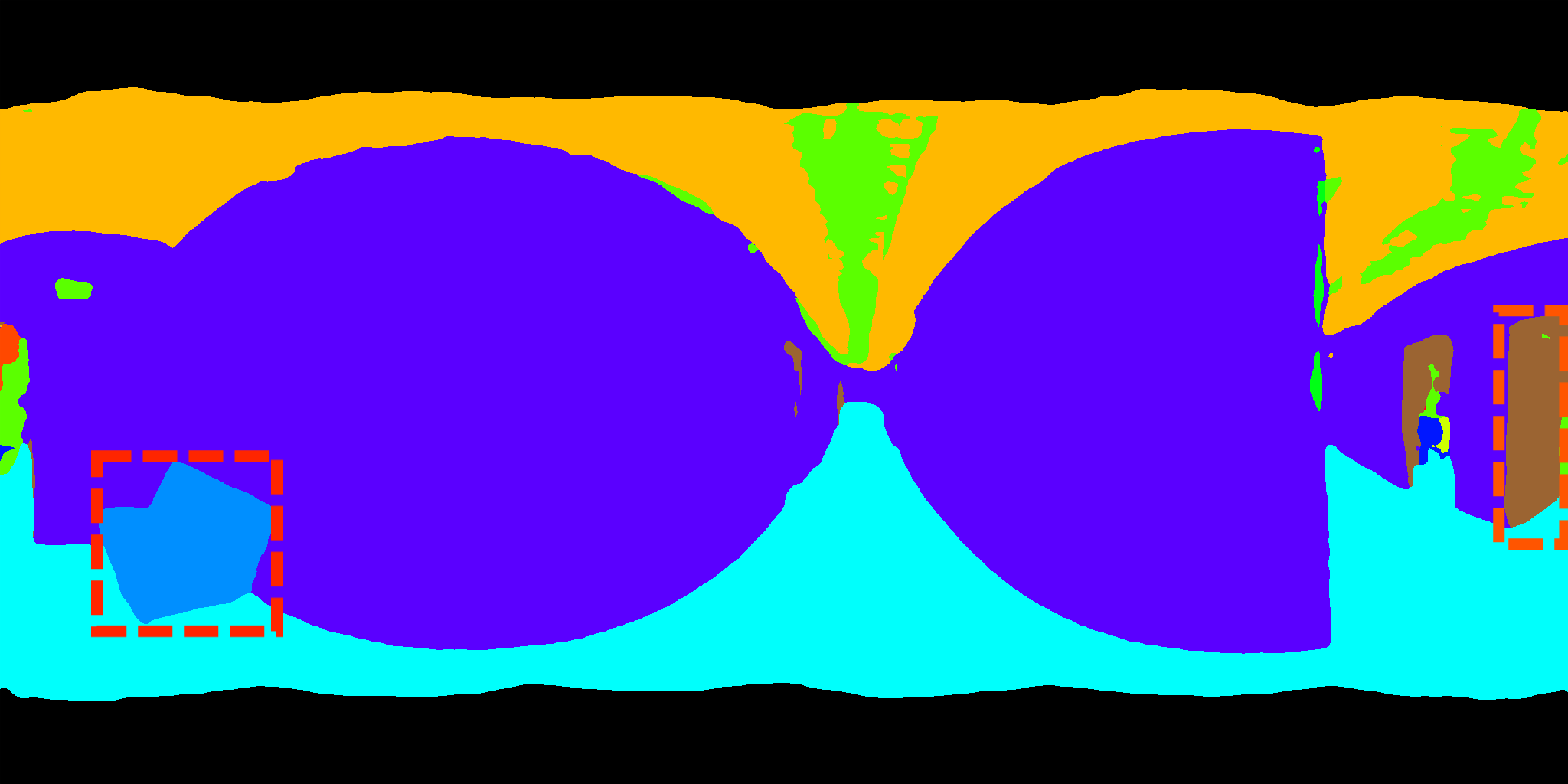}
        \caption{Our results}
        \label{sfig:source_ours}
    \end{subfigure}
    
    \begin{subfigure}{0.23\linewidth}
        \centering
        \includegraphics[width=1\linewidth]{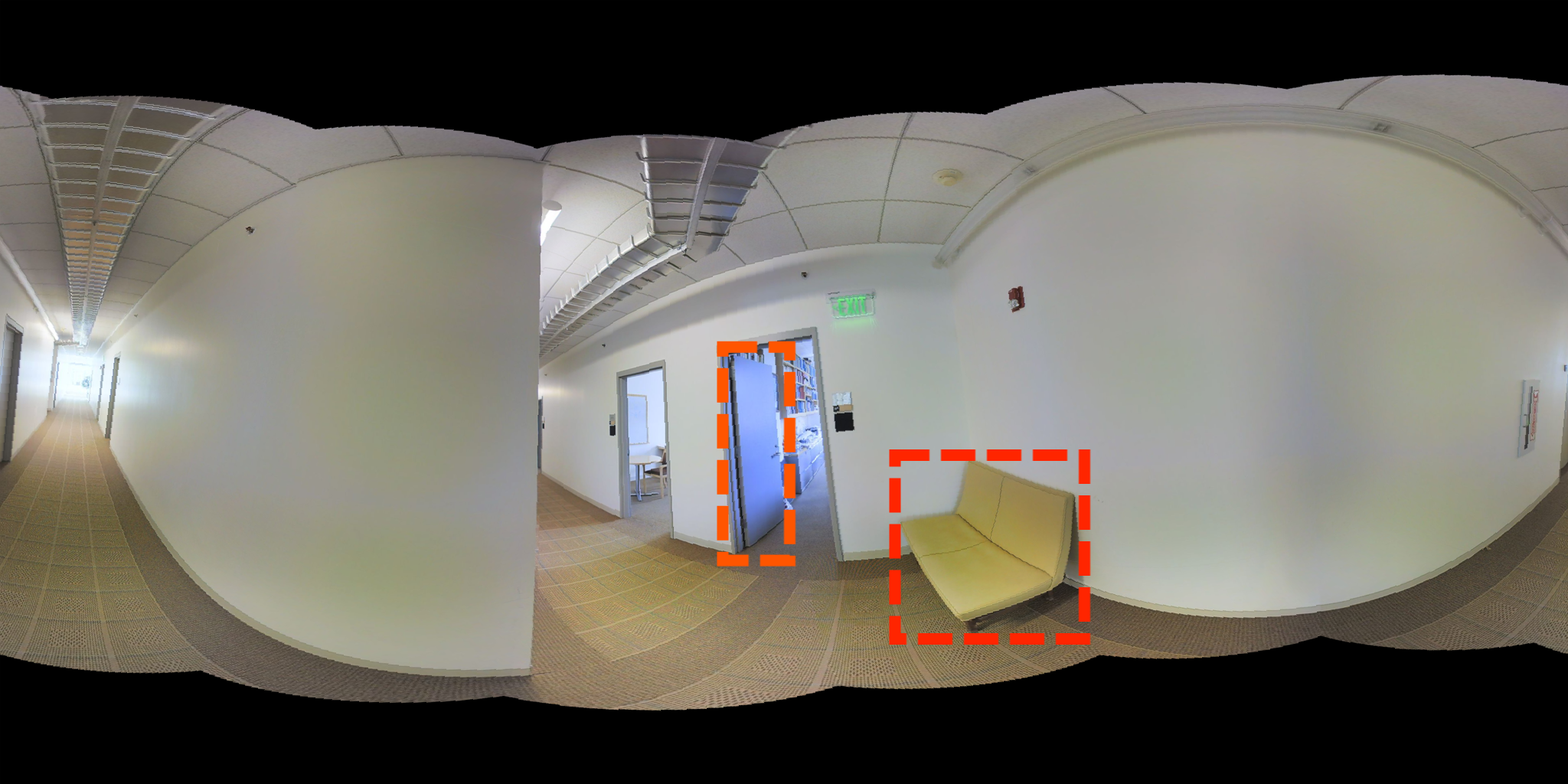}
        \caption{Rotated original picture}
        \label{sfig:rotated}
    \end{subfigure}
    \begin{subfigure}{0.23\linewidth}
        \centering
        \includegraphics[width=1\linewidth]{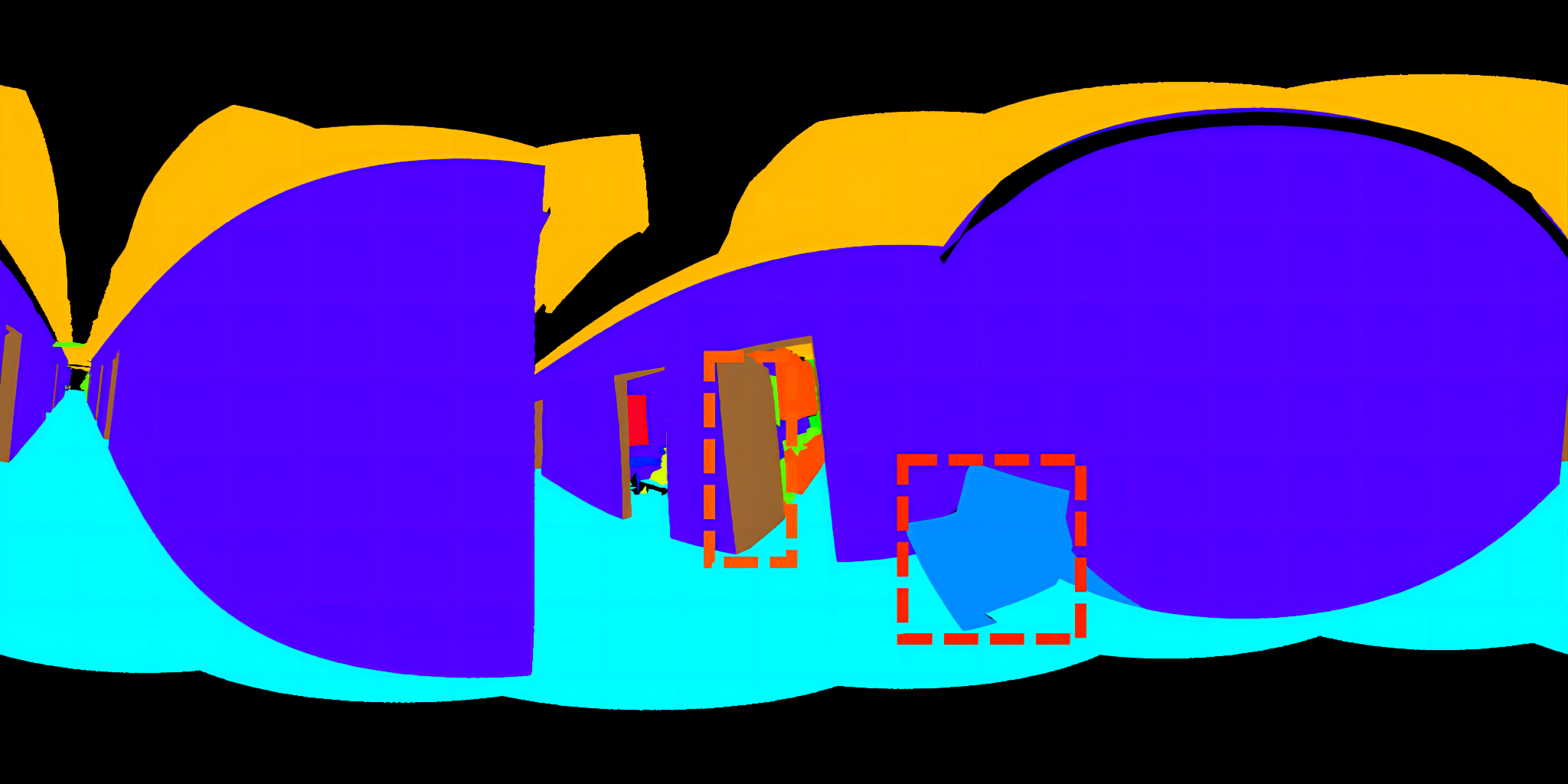}
        \caption{Rotated label}
        \label{sfig:rotated_GT}
    \end{subfigure}
    \begin{subfigure}{0.23\linewidth}
        \centering
        \includegraphics[width=1\linewidth]{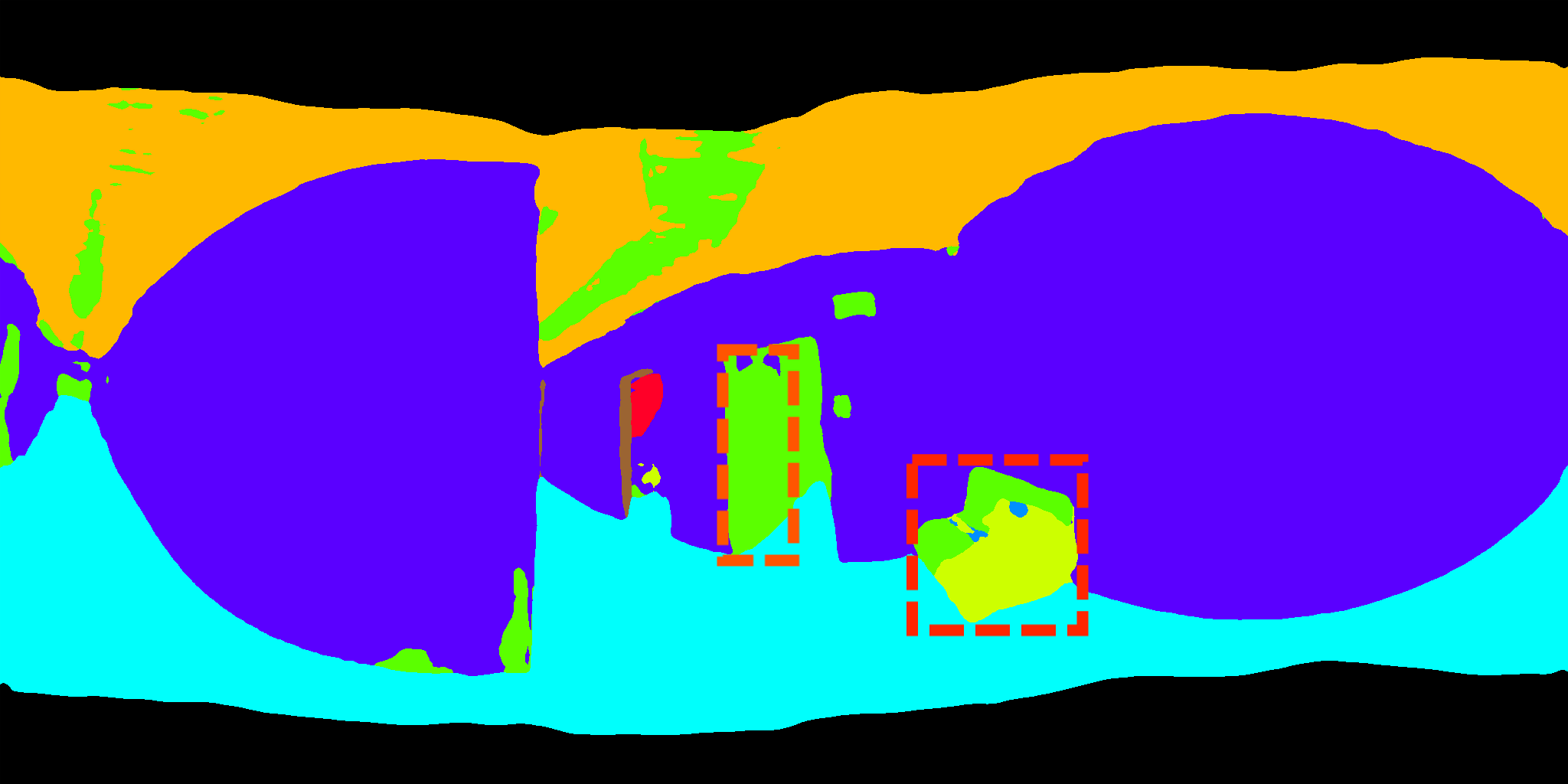}
        \caption{Baseline rotated results}
        \label{sfig:rotated_BL}
    \end{subfigure}
    \begin{subfigure}{0.23\linewidth}
        \centering
        \includegraphics[width=1\linewidth]{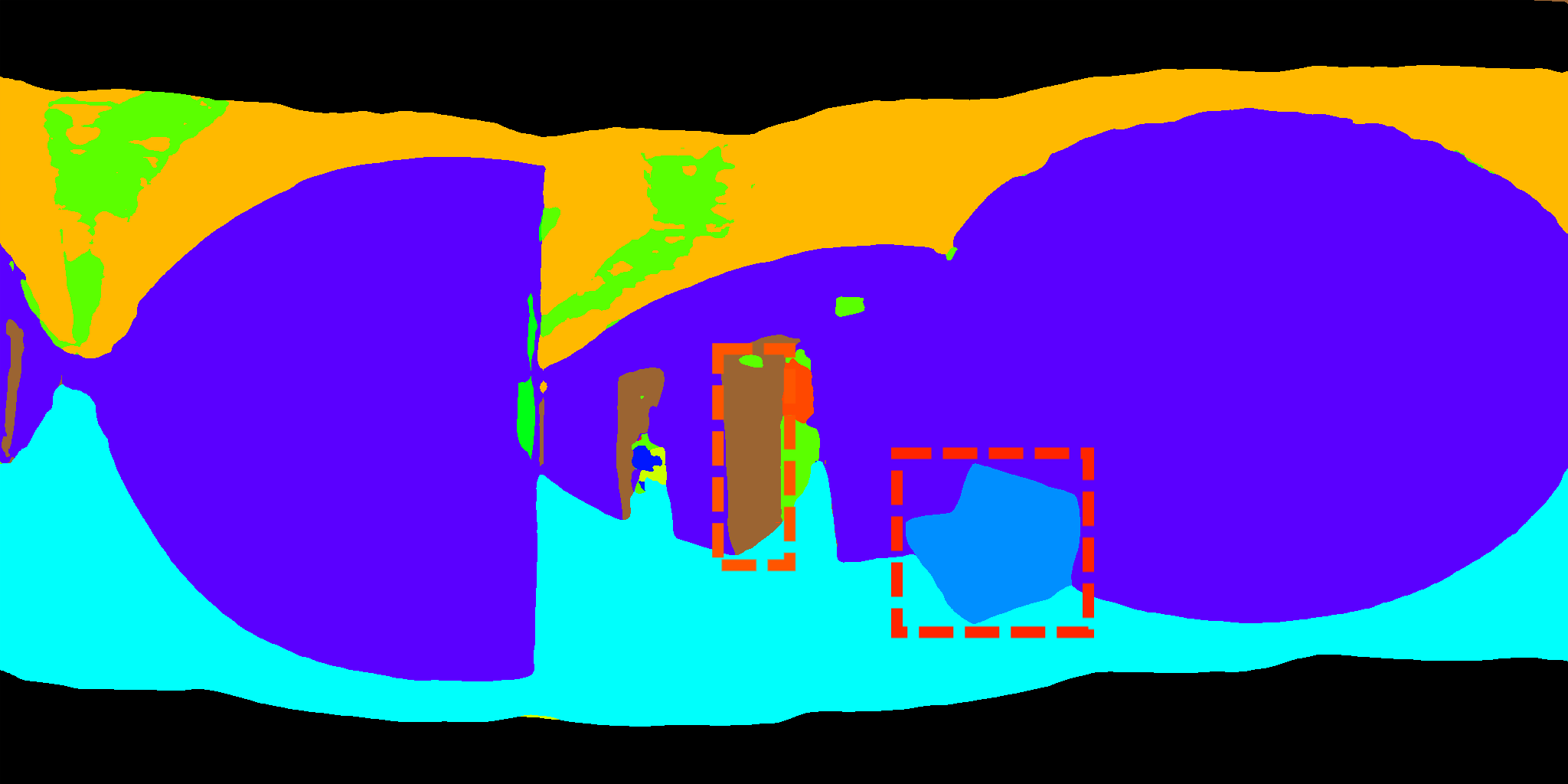}
        \caption{Our rotated results}
        \label{sfig:rotated_ours}
    \end{subfigure}
    \caption{Visualization comparison of \ours and Trans4PASS+. 
    The rotation of the pitch / roll / yaw axis is $5^{\circ}$ / $5^{\circ}$ / $180^{\circ}$. 
    \ours gains the better results of semantic class ``door'' and ``sofa'' (highlighted by red dotted line boxes). 
    }
    \label{fig:Visualization}
    \centering
\end{figure*}

\begin{table}
    \footnotesize
    \centering
    \resizebox{0.45\textwidth}{!}{
        \begin{tabular}{ccccc}
            \toprule
            \multirow{2}{*}{Statistics} & \multicolumn{2}{c}{Baseline} & \multicolumn{2}{c}{Ours} \\
                                        & mIoU        & Pixel accuracy      & mIoU      & Pixel accuracy    \\ 
            \midrule
            \multicolumn{3}{l}{$(\beta, \gamma, \alpha) = (1^{\circ},1^{\circ},360^{\circ})$} \\
            Mean                        & 53.473      & 81.251         & 56.212 ({\color{red}{+2.739}})    & 83.021 ({\color{red}{+1.770}})       \\
            Variance                    & 0.056       & 0.029          & 0.011 ({\color{red}{-0.045}})     & 0.003 ({\color{red}{-0.026}})        \\
            Range                       & 0.856       & 0.591          & 0.394 ({\color{red}{-0.462}})     & 0.192 ({\color{red}{-0.399}})         \\ 
            \midrule
            \multicolumn{3}{l}{$(\beta, \gamma, \alpha) = (0^{\circ},0^{\circ},360^{\circ})$} \\
            Mean                        & 53.698      & 81.502         & 56.321 ({\color{red}{+2.623}})    & 83.093 ({\color{red}{+1.591}})       \\
            Variance                    & 0.017       & 0.003          & 0.008 ({\color{red}{-0.009}})     & 0.002 ({\color{red}{-0.001}})        \\
            Range                       & 0.331       & 0.131          & 0.218 ({\color{red}{-0.113}})     & 0.084 ({\color{red}{-0.047}})         \\ 
            \bottomrule
        \end{tabular}
        }
    \caption{
    Overall performance comparison on Stanford2D3D Panoramic datasets in different SGA metrics in two more favorable settings for Tran4PASS+. 
    \ours also earns considerable mean performance and significant robustness improvement.      
    }
    \label{tab:STA_SGAM}
\end{table}

\section{Experiments}
\label{sec:experiments}
In this section, we evaluate our \ours against the popular benchmark, Stanford2D3D, for both traditional metrics and our SGA validation. 

\subsection{Datasets and Protocols}
\label{ssec:datasets}
We validate \ours on Stanford2D3D Panoramic datasets~\cite{armeni2017joint}. 
It has 1,413 panoramas, and 13 semantic classes are labeled, and has 3 official folds, fold 1 / 2 / 3. 
We follow the report style of previous work~\cite{zhang2022bending}~\cite{zhang2022behind}. 

Our experiments are conducted with a server with four A100 GPUs. 
We use Trans4PASS+~\cite{zhang2022behind} as our baseline and set an initial learning rate of 8e-5, which is scheduled by the poly strategy with 0.9 power over 150 epochs. 
The optimizer is AdamW~\cite{kingma2015adam} with epsilon 1e-8, weight decay 1e-4, and batch size is 4 on each GPU. 
Other settings and hyperparameters are set the same as Trans4PASS+~\cite{zhang2022behind}.
For each input panoramic image $I$ in an iteration, there is a 50\% chance of using it directly and the other 50\% chance of using it after SGA image projection, $O_{3D}(I, \alpha_{\mathrm{use}}, \beta_{\mathrm{use}}, \gamma_{\mathrm{use}})$, where $\alpha_{\mathrm{use}}$ / $\beta_{\mathrm{use}}$ / $\gamma_{\mathrm{use}}$ uniformly sampled from $0$ to $\alpha_{\mathrm{train}}$ / $\beta_{\mathrm{train}}$ / $\gamma_{\mathrm{train}}$. 
We set $(\beta_{\mathrm{train}}, \gamma_{\mathrm{train}}, \alpha_{\mathrm{train}}) = ( 10^{\circ},  10^{\circ},  360^{\circ})$.  
$\lambda_w$ and $\lambda_s$ are set as 0.3 and 0.3, respectively. 

\begin{table*}[]
    \footnotesize
    \centering
    \resizebox{0.95\textwidth}{!}{
    \begin{tabular}{c|c|c|ccccccccccccc}
    \toprule
    Network     & Test Method                   & mIoU  & beam & board & bookcase & ceiling & chair & clutter & column & door  & floor & sofa  & table & wall  & window \\ \midrule
    Trans4Pass+ & \multirow{2}{*}{Traditional} & 53.62 & 0.39 & 74.4  & 65.32    & 84.21   & 62.86 & 36.44   & 15.96  & 32.79 & 93.09 & 44.10 & 63.67 & 75.02 & 46.90  \\
    Ours        &                               & 56.37 & 0.73 & 74.05 & 65.91    & 84.20   & 64.53 & 41.24   & 19.62  & \color{red}{\textbf{52.67}} & 93.08 & \color{red}{\textbf{56.92}} & 58.86 & 76.43 & 44.62  \\ \midrule
    Trans4Pass+ & \multirow{2}{*}{SGA}  & 50.03 & 0.26 & 73.78 & 62.21    & 83.82   & 61.87 & 32.11   & 10.93  & 20.26 & 92.96 & 38.33 & 61.78 & 74.35 & 37.73  \\
    Ours        &                               & 55.98 & 0.78 & 73.94 & 65.56    & 84.08   & 64.39 & \color{red}{40.96}   & \color{red}{18.31}  & \color{red}{\textbf{51.64}} & 92.98 & \color{red}{\textbf{56.53}} & 58.14 & 76.06 & \color{red}{44.42}  \\ \bottomrule
    \end{tabular}
    }
    \caption{    
    Per-class mIoU results on Stanford2D3D Panoramic datasets according to the fold 1 data setting with traditional mIoU and per-pixel accuracy metrics. 
    No mark for the results that the gap between Trans4Pass+ and Ours less than 5\% (performance at the same level). 
    Our results will be {\color{red}red} when Ours outperforms more than 5\%. 
    If Ours outperforms more than 10\%, our results will be {\color{red}\textbf{bold and red}}.
    There is no semantic class that Trans4Pass+ outperforms Ours 5\% or more.
    }
    \label{tab:per_class_miou}
\end{table*}

\paragraph{Spherical Geometry-Aware (SGA) Validation.}
Most PASS datasets use a unified ERP way to process original $360^{\circ}$ data, PASS models have the potential to overfit the ERP way, cannot handle 3D disturbance well and have little 3D robustness.  
To validate the robustness of the PASS models, we propose a novel SGA validation. 
$n_{\alpha}$, $n_{\beta}$, and $n_{\gamma}$ are the number of different angles for the yaw / pitch / roll axis, respectively, and $n_{\alpha} \cdot n_{\beta} \cdot n_{\gamma}$ different-angle panoramic images for a certain original $360^{\circ}$ data is earned. 
Panoramic semantic segmentation models are validated in all $n_{\alpha} \cdot n_{\beta} \cdot n_{\gamma}$ settings, and their statistics are reported as SGA metrics. 
In our SGA validation, ``Mean''  means the average of all $n_{\alpha} \cdot n_{\beta} \cdot n_{\gamma}$ traditional results (e.g., mIoU, per pixel accuracy, etc.). 
``Variance''  means the variance of all $n_{\alpha} \cdot n_{\beta} \cdot n_{\gamma}$ results.  
``Range''  means the gap between the maximum and minimum results of all $n_{\alpha} \cdot n_{\beta} \cdot n_{\gamma}$ results. 
Compared to traditional validation, SGA validation avoids models gain performance by fitting the ERP way of datasets and reflects objective 3D robustness. 
In detail, we assume that the 3D rotation disturbance is at most $5^{\circ}$ / $5^{\circ}$ / $360^{\circ}$ of pitch ($\beta$) / roll ($\gamma$) / yaw ($\alpha$) angle. 
We set $n_{\alpha}=4$ ($0^{\circ}, 90^{\circ}, 180^{\circ}, 270^{\circ}$), $n_{\beta}=2$ ($0^{\circ}, 5^{\circ}$), and $n_{\gamma}=2$ ($0^{\circ}, 5^{\circ}$). 
We use the mean of them as the final performance and observe the performance difference among them to indicate the 3D robustness of models. 

\subsection{Performance Comparison}
\label{ssec:SOTA}
In this part, we first compare several recent SOTA methods with traditional metrics, and then compare the latest SOTA Trans4PASS+ in detail with SGA metrics. 

\paragraph{Traditional Metrics.}
Comparison results on Stanford2D3D Panoramic datasets with SOTA methods in traditional metrics are shown in~\Cref{tab:sota}. 
Following recent work, we report the performance of both official fold 1 and the average performance of all three official folds. 
From the results, \ours outperforms current SOTA models by  2.8\% / 1.6\% mIoU, respectively, which means that our \ours has a considerable performance margin compared to current models with traditional metrics. 

\paragraph{SGA Metrics.}
Comparison results on Stanford2D3D Panoramic datasets with our SGA validation metrics are shown in~\Cref{tab:STA_SGAM_BIG}, and~\Cref{tab:SGAMBIG} is the detailed performance of each situation. 
For mean mIoU / pixel accuracy, an improvement of nearly 6\% / 4\% is achieved, respectively. 
Furthermore, our variance is about $\frac{1}{100}$ and our fluctuation range is about $\frac{1}{10}$. 
These results show that our \ours have much better robustness than Trans4PASS+. 

\subsection{Ablation Study}
\label{ssec:ablation}
\paragraph{Effect of Three Modules in Training Process.}
The effectiveness of SGA image projection, SDPE, and panorama-aware loss are studied on Stanford2D3D Panoramic datasets official fold 1 with traditional metrics as shown in~\Cref{tab:ablation}. 
(a) SGA image projection: Using it alone improves the baseline mIoU / per pixel accuracy by 1.020\% / 0.820\%. 
(b) SDPE: Using SDPE alone outperforms the baseline by 0.937\% and 0.025\% in mIoU and per pixel accuracy. 
(b) Panorama-aware loss: Using it alone improves the baseline by 1.216\% and 0.250\% in mIoU and per pixel accuracy. 

\paragraph{Effect of SGA Validation.}
We demonstrate the effect of SGA validation, which means a stronger generalizability to resist 3D rotational perturbation. 
We carried out experiments with two smaller disturbance settings on the pitch and roll axes ($(\beta, \gamma, \alpha) = (1^{\circ},1^{\circ},360^{\circ})$ / $(0^{\circ},0^{\circ},360^{\circ})$), which are more favorable settings for Trans4PASS+~\cite{zhang2022behind}, because it is designed for the standard panoramic view image ($(\beta_{\mathrm{use}}, \gamma_{\mathrm{use}}, \alpha_{\mathrm{use}}) = (0^{\circ},0^{\circ},0^{\circ})$). 
The overall statistical results are shown in ~\Cref{tab:STA_SGAM}. 
For the $(\beta, \gamma, \alpha) = (1^{\circ},1^{\circ},360^{\circ})$ setting, an improvement of approximately 2.7 \% / 1.7 \% is obtained for the mean mIoU / pixel accuracy. 
Our variance is approximately $\frac{1}{5}$ / $\frac{1}{10}$ and our fluctuation range is approximately $\frac{1}{2}$ / $\frac{1}{3}$ in mIoU / pixel precision. 
In $(\beta, \gamma, \alpha) = (0^{\circ},0^{\circ},360^{\circ})$ setting, mean mIoU / pixel accuracy gains approximately 2.6\% / 1.6\% improvement, variances / fluctuation is approximately $\frac{1}{2}$ / $\frac{2}{3}$ for \ours. 
\ours has better robustness even with little 3D perturbations. 
The detailed performance of these two settings and the performance of several random rotation settings are shown in Section A “Detailed Performance of SGA Validation” in the supplementary material.  

\subsection{Discussion and Visualizations}
\label{ssec:discussion}
\paragraph{Performance of All Semantic Classes and Visualizations.}
We show the detailed performance of all 13 semantic classes on the Stanford2D3D Panoramic datasets with both traditional and SGA metrics in~\Cref{tab:per_class_miou}, respectively. 
We focus mainly on the classes with significant performance gaps and mark the gap larger than 5\% / 10\% as red numbers / bold red numbers, respectively. 
There is no semantic class for which the baseline is significantly better. 
From the results, we can learn that the ``sofa'' and ``door'' classes improve more. 
An image with ``door'' and ``sofa'' is visualized in~\Cref{fig:Visualization}. 
Rotation of the pitch / roll / yaw axis is $5^{\circ}$ / $5^{\circ}$ / $180^{\circ}$. 
The baseline prediction gap between the original and rotated input is large, which means less robustness. 
It predicts the door near the right boundary in~\Cref{sfig:source_BL} overall right, but it is totally wrong with rotation in~\Cref{sfig:rotated_BL} when \ours predicts both correct. 
The baseline predictions for the sofa change a lot with rotation when \ours is stable.  
More visualizations are shown in Section B “More Visualizations” in the supplementary material.  

\begin{figure}[tb]
    \centering
    \begin{subfigure}{0.42\linewidth}
        \centering
        \includegraphics[width=1\linewidth]{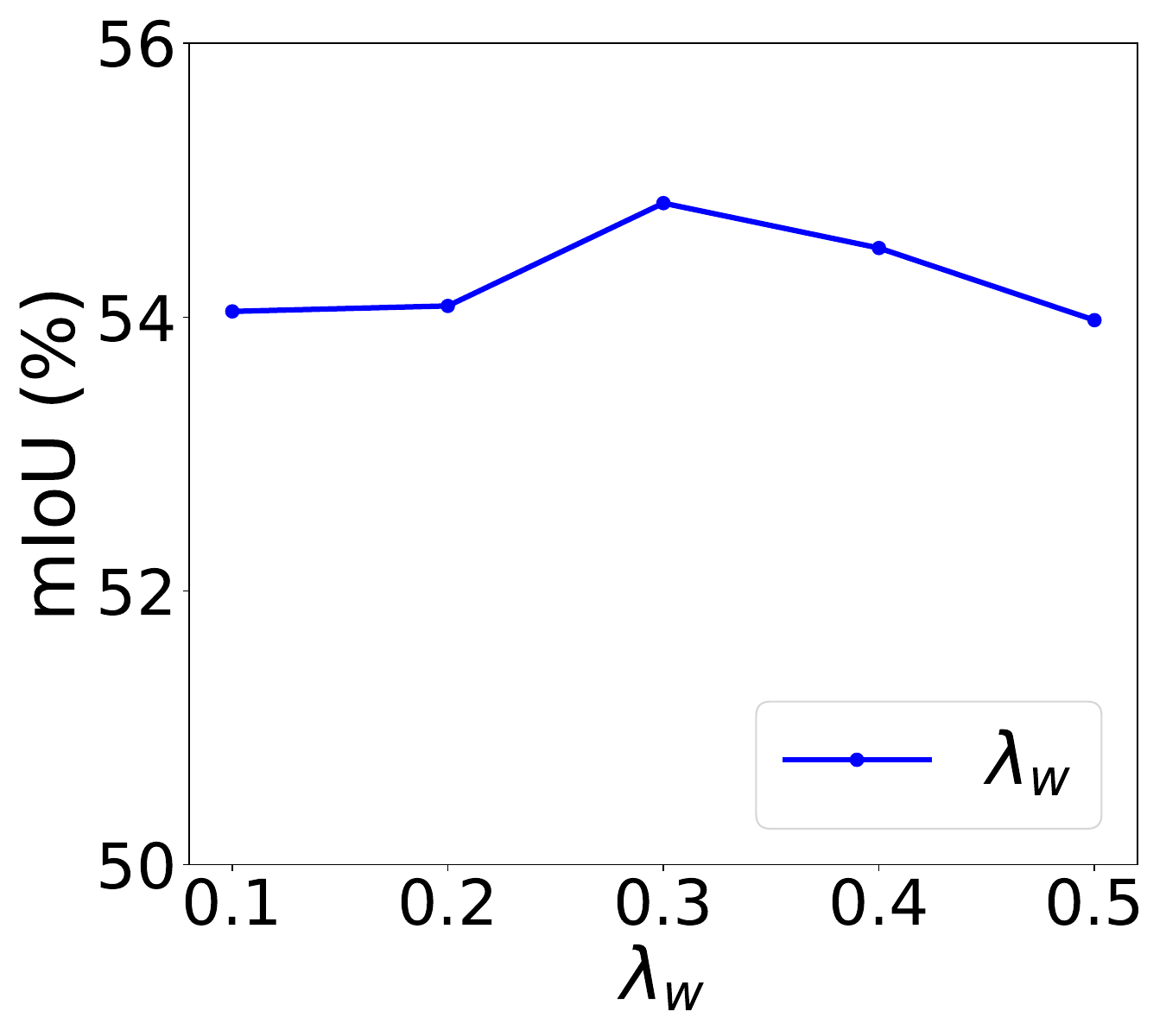}
        \caption{$\lambda_w$-mIoU}
        \label{sfig:HP1}
    \end{subfigure}
    \begin{subfigure}{0.42\linewidth}
        \centering
        \includegraphics[width=1\linewidth]{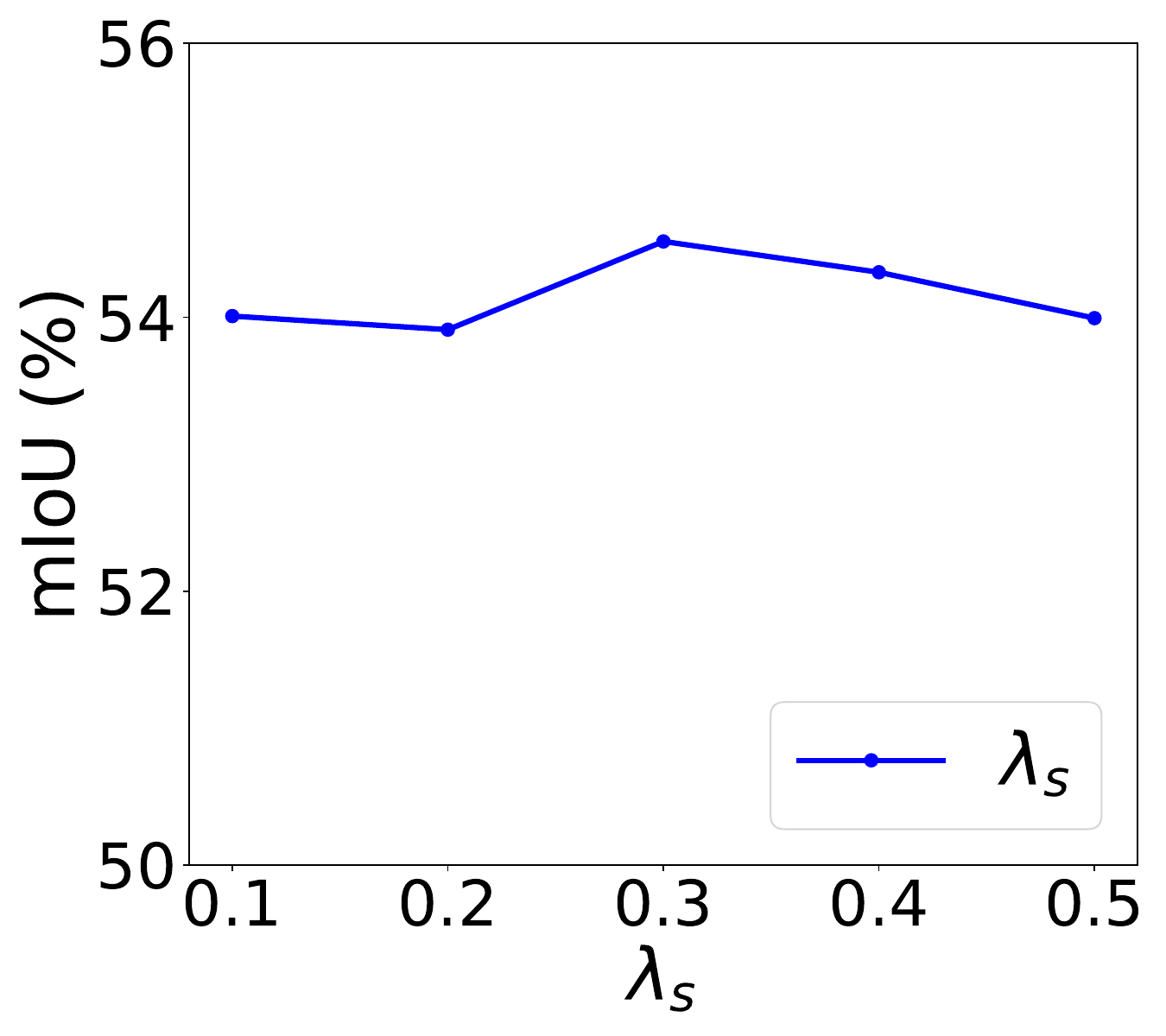}
        \caption{$\lambda_s$-mIoU}
        \label{sfig:HP2}
    \end{subfigure}
    \caption{Influence of $\lambda_s$ and $\lambda_w$ in \ours. 
    The results are carried out on Stanford2D3D Panoramic datasets official fold 1. }
    \label{fig:HP_fig}
    \centering
\end{figure}

\paragraph{Different Hyper-Parameters.}
$\lambda_w$ and $\lambda_s$ are hyperparameters in our \ours.  
$\lambda_s$ / $\lambda_w$ determines the proportion of our constraint of spherical geometry in SDPE / panorama-aware loss. 
We apply them on the baseline, respectively. 
Traditional mIoU results are shown in~\Cref{sfig:HP1} and~\Cref{sfig:HP2}, and we choose 0.3 / 0.3 as the final $\lambda_w$ / $\lambda_s$. 

\section{Conclusion}
\label{sec:conclusion}
We have studied an underexplored but important field in panoramic semantic segmentation, i.e., the robustness of dealing with 3D disturbance panoramic input images. 
We have shown that using our SGA framework is key to improving the semantic segmentation quality of 3D disturbance inputs. 
It applies spherical geometry prior to panoramic semantic segmentation and gains considerable improvement. 
In detail, the SGA framework includes SGA image projection, SDPE, and panorama-aware loss.  
We also validated the effectiveness of our \ours on popular datasets with the traditional metrics and the proposed SGA metrics, and studied its properties both empirically and theoretically. 
\section*{Acknowledgements}
This work is supported in part by National Key Research and Development Program of China under Grant 2020AAA0107400, National Natural Science Foundation of China under Grant U20A20222, National Science Foundation for Distinguished Young Scholars under Grant 62225605, Research Fund of ARC Lab, Tencent PCG, The Ng Teng Fong Charitable Foundation in the form of ZJU-SUTD IDEA Grant, 188170-11102 as well as CCF-Zhipu AI Large Model Fund (CCF-Zhipu202302).

\section*{Contribution statement}
Xuewei Li and Tao Wu contributed equally to this work.


\bibliographystyle{named}
\bibliography{main}

\end{document}


\maketitle
\renewcommand\thesection{\Alph{section}}
\renewcommand{\theequation}{S-\arabic{equation}}
\setcounter{equation}{0}
\renewcommand{\thefigure}{S-\arabic{figure}}
\setcounter{figure}{0}
\renewcommand{\thetable}{S-\arabic{table}}
\setcounter{table}{0}

\renewcommand{\arraystretch}{1.5}
\begin{table*}
    \footnotesize
    \centering
    \resizebox{1\textwidth}{!}{
        \begin{tabular}{l|cccc|cccc|cccc}
            \toprule
            \multicolumn{1}{c|}{\multirow{3}{*}{($\beta$,$\gamma$,$\alpha$) ($^{\circ}$)}}             & \multicolumn{4}{c|}{Mean}                                                                        & \multicolumn{4}{c|}{Variance}                                                                  & \multicolumn{4}{c}{Range}                                                                     \\ \cline{2-13} 
            \multicolumn{1}{c|}{}                              & \multicolumn{2}{c|}{mIoU}                                 & \multicolumn{2}{c|}{Pixel Accuracy}  & \multicolumn{2}{c|}{mIoU}                                & \multicolumn{2}{c|}{Pixel Accuracy} & \multicolumn{2}{c|}{mIoU}                                & \multicolumn{2}{c}{Pixel Accuracy} \\ \cline{2-13} 
            \multicolumn{1}{c|}{}                              & \multicolumn{1}{c|}{BL}     & \multicolumn{1}{c|}{Ours}   & \multicolumn{1}{c|}{BL}     & Ours   & \multicolumn{1}{c|}{BL}     & \multicolumn{1}{c|}{Ours}  & \multicolumn{1}{c|}{BL}     & Ours  & \multicolumn{1}{c|}{BL}     & \multicolumn{1}{c|}{Ours}  & \multicolumn{1}{c|}{BL}    & Ours  \\ \midrule
            \multirow{2}{*}{(0,0,360)}                         & \multicolumn{1}{c|}{53.698} & \multicolumn{1}{c|}{56.321} & \multicolumn{1}{c|}{81.502} & 83.093 & \multicolumn{1}{c|}{0.017}  & \multicolumn{1}{c|}{0.008} & \multicolumn{1}{c|}{0.003}  & 0.002 & \multicolumn{1}{c|}{0.331}  & \multicolumn{1}{c|}{0.218} & \multicolumn{1}{c|}{0.131} & 0.084 \\ \cline{2-13} 
                                                               & \multicolumn{2}{c|}{\color{red}{+2.623}}                                & \multicolumn{2}{c|}{\color{red}{+1.591}}           & \multicolumn{2}{c|}{\color{red}{-0.009}}                              & \multicolumn{2}{c|}{\color{red}{-0.001}}         & \multicolumn{2}{c|}{\color{red}{-0.113}}                              & \multicolumn{2}{c}{\color{red}{-0.047}}         \\ \hline
            \multirow{2}{*}{(1,1,360)}                         & \multicolumn{1}{c|}{53.473} & \multicolumn{1}{c|}{56.212} & \multicolumn{1}{c|}{81.251} & 83.021 & \multicolumn{1}{c|}{0.056}  & \multicolumn{1}{c|}{0.011} & \multicolumn{1}{c|}{0.029}  & 0.003 & \multicolumn{1}{c|}{0.856}  & \multicolumn{1}{c|}{0.394} & \multicolumn{1}{c|}{0.591} & 0.192 \\ \cline{2-13} 
                                                               & \multicolumn{2}{c|}{\color{red}{+2.739}}                                & \multicolumn{2}{c|}{\color{red}{+1.770}}           & \multicolumn{2}{c|}{\color{red}{-0.045}}                              & \multicolumn{2}{c|}{\color{red}{-0.026}}         & \multicolumn{2}{c|}{\color{red}{-0.462}}                              & \multicolumn{2}{c}{\color{red}{-0.399}}         \\ \hline
            \multirow{2}{*}{(3,3,360)}                         & \multicolumn{1}{c|}{51.862} & \multicolumn{1}{c|}{56.097} & \multicolumn{1}{r|}{80.096} & 82.972 & \multicolumn{1}{c|}{1.472}  & \multicolumn{1}{c|}{0.026} & \multicolumn{1}{c|}{0.818}  & 0.007 & \multicolumn{1}{c|}{3.787}  & \multicolumn{1}{c|}{0.603} & \multicolumn{1}{c|}{2.726} & 0.275 \\ \cline{2-13} 
                                                               & \multicolumn{2}{c|}{\color{red}{+4.235}}                                & \multicolumn{2}{c|}{\color{red}{+2.876}}           & \multicolumn{2}{c|}{\color{red}{-1.446}}                              & \multicolumn{2}{c|}{\color{red}{-0.811}}         & \multicolumn{2}{c|}{\color{red}{-3.184}}                              & \multicolumn{2}{c}{\color{red}{-2.451}}         \\ \hline
            \multirow{2}{*}{(5,5,360)}                         & \multicolumn{1}{c|}{50.033} & \multicolumn{1}{c|}{55.984} & \multicolumn{1}{c|}{78.949} & 82.887 & \multicolumn{1}{c|}{5.147}  & \multicolumn{1}{c|}{0.066} & \multicolumn{1}{c|}{2.413}  & 0.020 & \multicolumn{1}{c|}{6.684}  & \multicolumn{1}{c|}{0.940} & \multicolumn{1}{c|}{4.461} & 0.478 \\ \cline{2-13} 
                                                               & \multicolumn{2}{c|}{\color{red}{+5.951}}                                & \multicolumn{2}{c|}{\color{red}{+3.938}}           & \multicolumn{2}{c|}{\color{red}{-5.081}}                              & \multicolumn{2}{c|}{\color{red}{-2.393}}         & \multicolumn{2}{c|}{\color{red}{-5.744}}                              & \multicolumn{2}{c}{\color{red}{-3.983}}         \\ \hline
            \multirow{2}{*}{(10,10,360)}                       & \multicolumn{1}{c|}{47.317} & \multicolumn{1}{c|}{55.445} & \multicolumn{1}{c|}{77.302} & 82.531 & \multicolumn{1}{c|}{14.338} & \multicolumn{1}{c|}{0.378} & \multicolumn{1}{c|}{6.140}  & 0.151 & \multicolumn{1}{c|}{10.535} & \multicolumn{1}{c|}{1.884} & \multicolumn{1}{c|}{6.787} & 1.170 \\ \cline{2-13} 
                                                               & \multicolumn{2}{c|}{\color{red}{+8.128}}                                & \multicolumn{2}{c|}{\color{red}{+5.229}}           & \multicolumn{2}{c|}{\color{red}{-13.960}}                             & \multicolumn{2}{c|}{\color{red}{-5.989}}         & \multicolumn{2}{c|}{\color{red}{-8.651}}                              & \multicolumn{2}{c}{\color{red}{-5.617}}         \\ \bottomrule
            \end{tabular}
    }
    \caption{
    Overall performance comparison on Stanford2D3D Panoramic datasets in different SGA metrics in five different settings with Tran4PASS+, i.e., the baseline (BL) method. 
    \ours earns considerable mean performance and significant robustness improvement in each setting. 
    }
    \label{tab:STA_SGAM_sup}
\end{table*}

\section{Detailed Performance of SGA Validation} 
\label{SGA validation}
In this section, we carry additional experiments to show the detailed performance of different SGA validation settings. 

First, we show five other different settings, $(\beta,\gamma,\alpha) =  (1^{\circ},1^{\circ},360^{\circ})$ (shown in~\Cref{tab:SGAM}) / $(3^{\circ},3^{\circ},360^{\circ})$ (shown in~\Cref{tab:SGAMthree}) / $(5^{\circ},5^{\circ},360^{\circ})$ (shown in Table 2 of the manuscript) / $(10^{\circ},10^{\circ},360^{\circ})$ (shown in~\Cref{tab:SGAMten}) / $(0^{\circ},0^{\circ},360^{\circ})$ (shown in the first two columns of any table mentioned above). 
The overall statistical results for all five settings are shown in~\Cref{tab:STA_SGAM_sup}. 
We find that our performance significantly outperforms the baseline even when the test data are only slightly 3D disturbance (rotational perturbation). 
Using mIoU as an example, when $(\beta, \gamma, \alpha) =  (0^{\circ},0^{\circ},360^{\circ})$, our mIoU is approximately 2.6\% higher than the baseline and the variance of mIoU is only $\frac{1}{2}$ of the baseline. 
When the 3D rotation disturbance is expanded to a large extent ($(\beta, \gamma, \alpha) =  (5^{\circ},5^{\circ},360^{\circ})$), we find that the mIoU of the baseline has decreased greatly, compared to 53.698\% ($(\beta, \gamma, \alpha) =  (0^{\circ},0^{\circ},360^{\circ})$ ), the mIoU of the baseline has decreased to 50.033\%, a decrease of approximately 3.7\%. 
However, our performance loss is relatively small, only a drop of approximately 0.3\%, and the mIoU variance is also much better than the baseline, about $\frac{1}{100}$.
When the 3D rotation disturbance is expanded to a larger extent ($(\beta, \gamma, \alpha) =  (10^{\circ},10^{\circ},360^{\circ})$), we find that the mIoU of the baseline has declined greatly, compared to 53.698\% ($(\beta, \gamma, \alpha) =  (0^{\circ},0^{\circ},360^{\circ})$ ), the mIoU of the baseline has decreased to 47.317\%, a decrease of approximately 6.4\%. 
However, our performance loss is relatively small, only a drop of approximately 0. 9\%, and the variance of mIoU is also much better than the baseline, about $\frac{1}{40}$. 
With the growth of 3D disturbances in $\beta$ and $\gamma$, the stability of Tran4PASS+ drops extremely fast when \ours maintains a relatively stable performance. 

Second, experiments are conducted with random rotated angles. 
We apply random rotation ($\beta$ / $\gamma$ / $\alpha$ random sampled from $0^{\circ}$ / $0^{\circ}$ / $0^{\circ}$ to $\beta_{r} / \gamma_{r} / \alpha_{r}$, respectively) for each test image. 
We choose several different settings for $(\beta_{r}, \gamma_{r}, \alpha_{r})$, run each setting 20 times, measure the mIoU for each result obtained, average the 20 experimental results for each setting as the final result of this setting performance, and report the mean mIoU of 20 repetitions. 
As shown in~\cref{tab:5times}, we set the maximum $\beta_{r}$ and $\gamma_{r}$ at $10^{\circ}$ because $(\beta_{\mathrm{train}}, \gamma_{\mathrm{train}}, \alpha_{\mathrm{train}}) = (10^{\circ},  10^{\circ},  360^{\circ})$. 
SGAT4PASS exhibits increased robustness and achieves performance when exposed to a wider range of diverse and random perturbations.

\begin{table}
    \footnotesize
    \centering
    \resizebox{0.45\textwidth}{!}{
        \begin{tabular}{ccccc}
            \toprule
            $(\beta_{r}, \gamma_{r}, \alpha_{r})$   & Baseline & Ours      & Gap \\
            \midrule
            $(1^{\circ}, 1^{\circ}, 360^{\circ})$   & 53.494   & 56.234    & {\color{red}{+2.740}} \\
            $(3^{\circ}, 3^{\circ}, 360^{\circ})$   & 52.330   & 56.138    & {\color{red}{+3.808}} \\
            $(5^{\circ}, 5^{\circ}, 360^{\circ})$   & 50.697   & 56.034    & {\color{red}{+5.337}} \\ 
            $(10^{\circ}, 10^{\circ}, 360^{\circ})$ & 47.506   & 55.611    & {\color{red}{+8.105}} \\ 
            \bottomrule
        \end{tabular}
    }
    \caption{Results with more diverse and random perturbations. The rotation ($\beta$ / $\gamma$ / $\alpha$) is randomly sampled from ($0^{\circ}$ / $0^{\circ}$ / $0^{\circ}$) to ($\beta_{r} / \gamma_{r} / \alpha_{r}$). 
    Reported values is the \textbf{mean mIoU} of 20 repetitions. 
    }
    \label{tab:5times}
\end{table}

\section{More Visualizations}
\begin{figure*}[tb]
    \centering
    \begin{subfigure}{0.23\linewidth}
        \centering
        \includegraphics[width=1\linewidth]{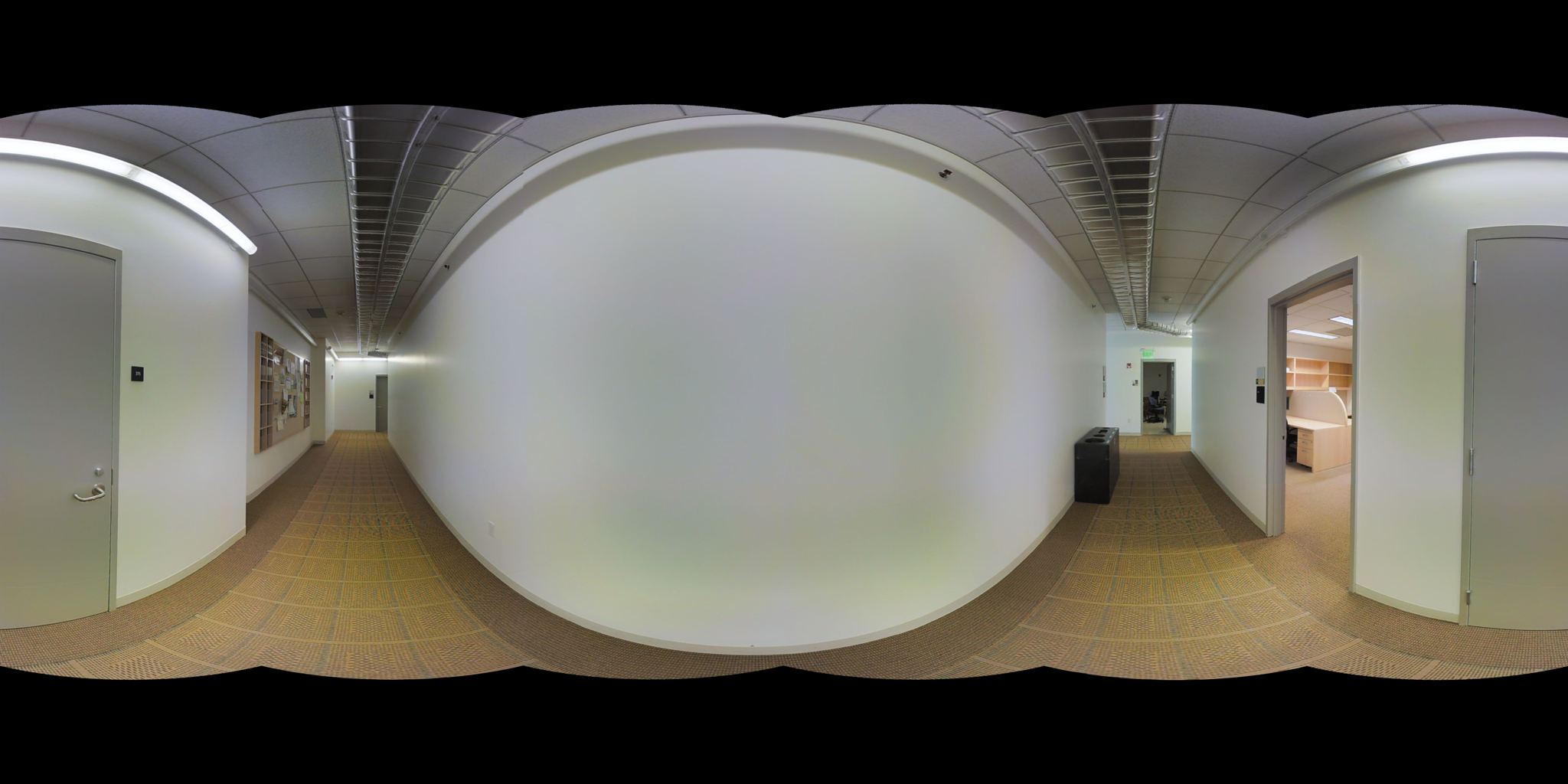}
        \caption{Original picture}
        \label{sfig:source_door1}
    \end{subfigure}
    \begin{subfigure}{0.23\linewidth}
        \centering
        \includegraphics[width=1\linewidth]{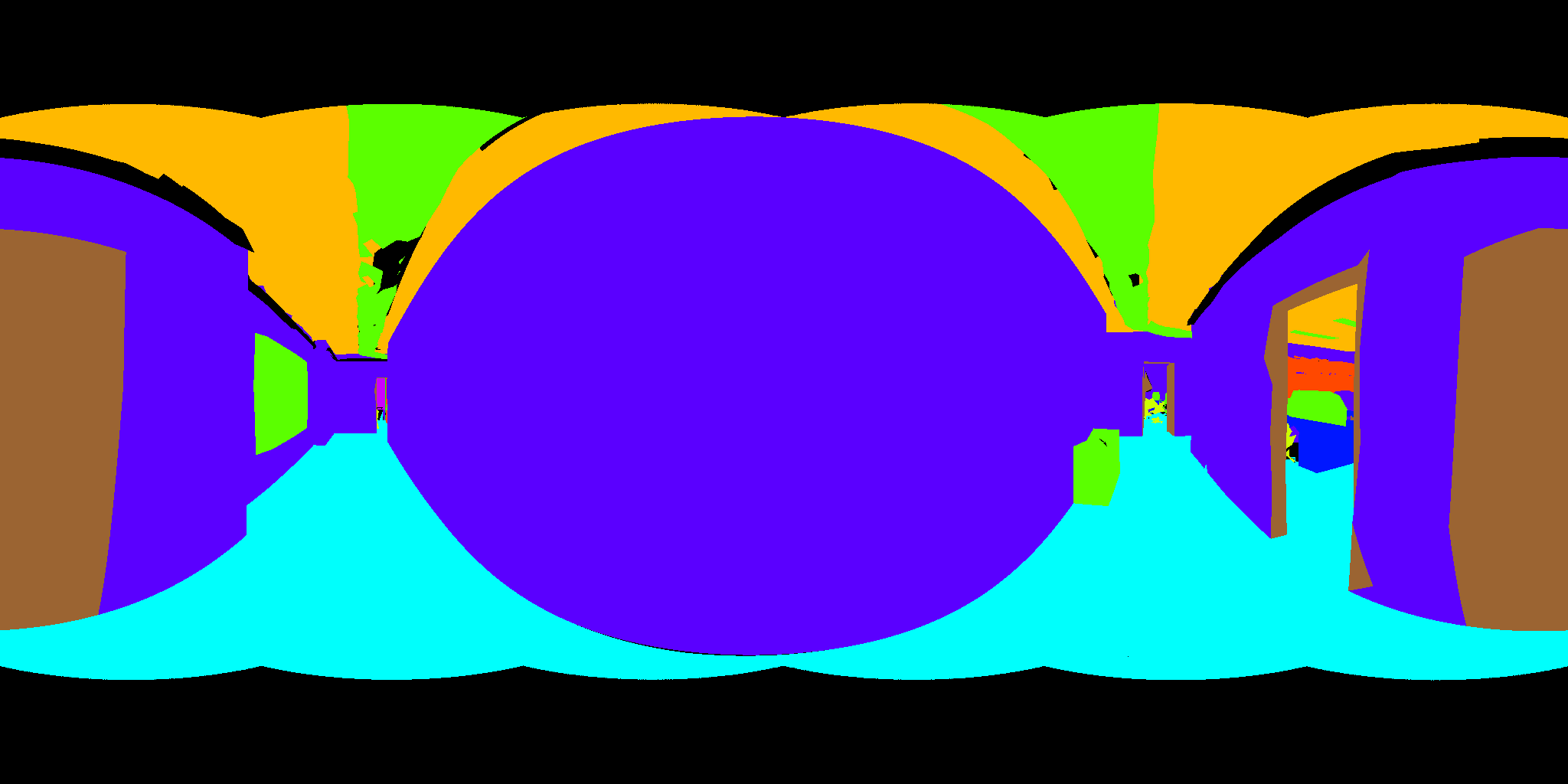}
        \caption{Label}
        \label{sfig:source_GT_door1}
    \end{subfigure}
    \begin{subfigure}{0.23\linewidth}
        \centering
        \includegraphics[width=1\linewidth]{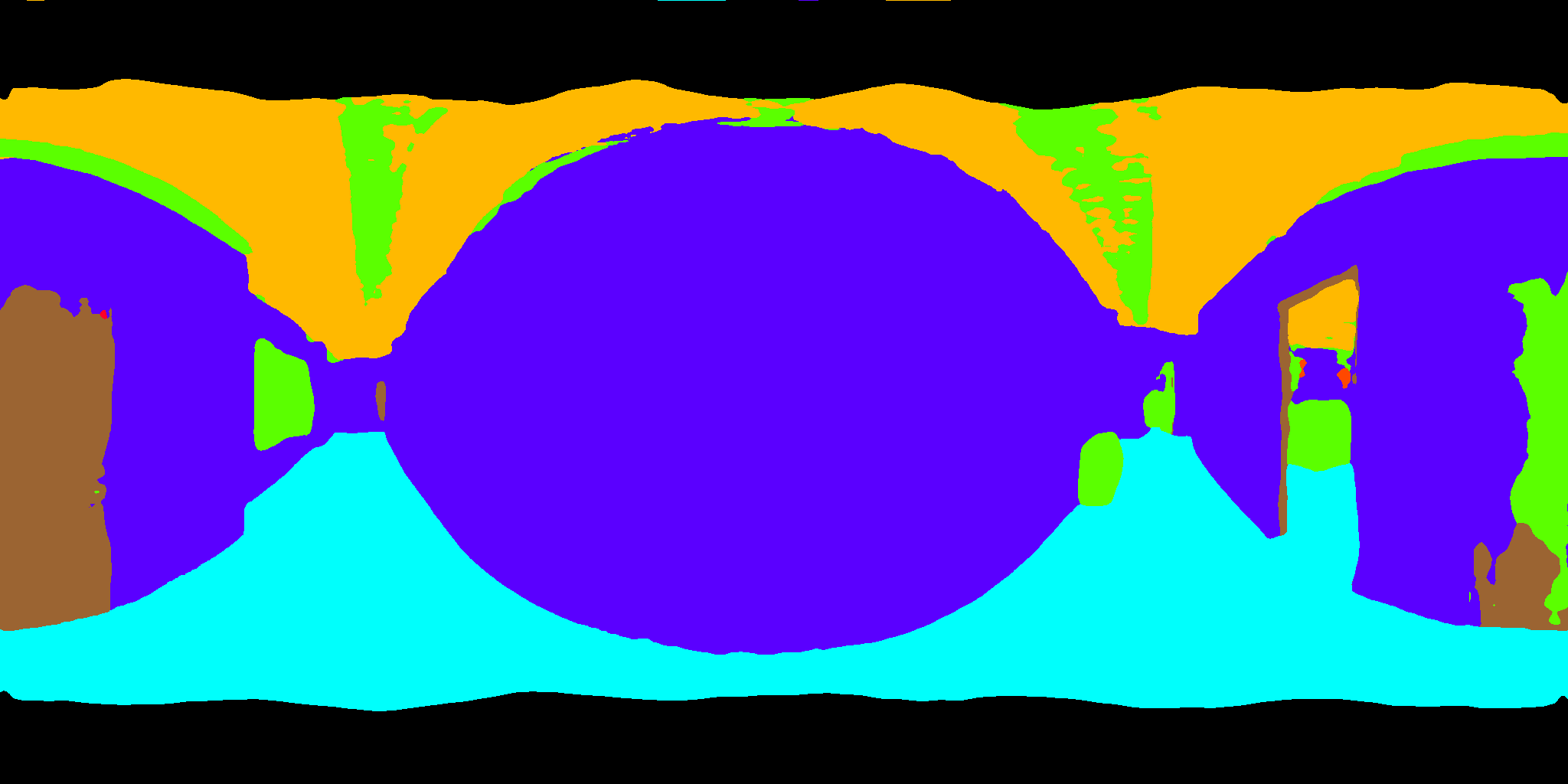}
        \caption{Baseline results}
        \label{sfig:source_BL_door1}
    \end{subfigure}
    \begin{subfigure}{0.23\linewidth}
        \centering
        \includegraphics[width=1\linewidth]{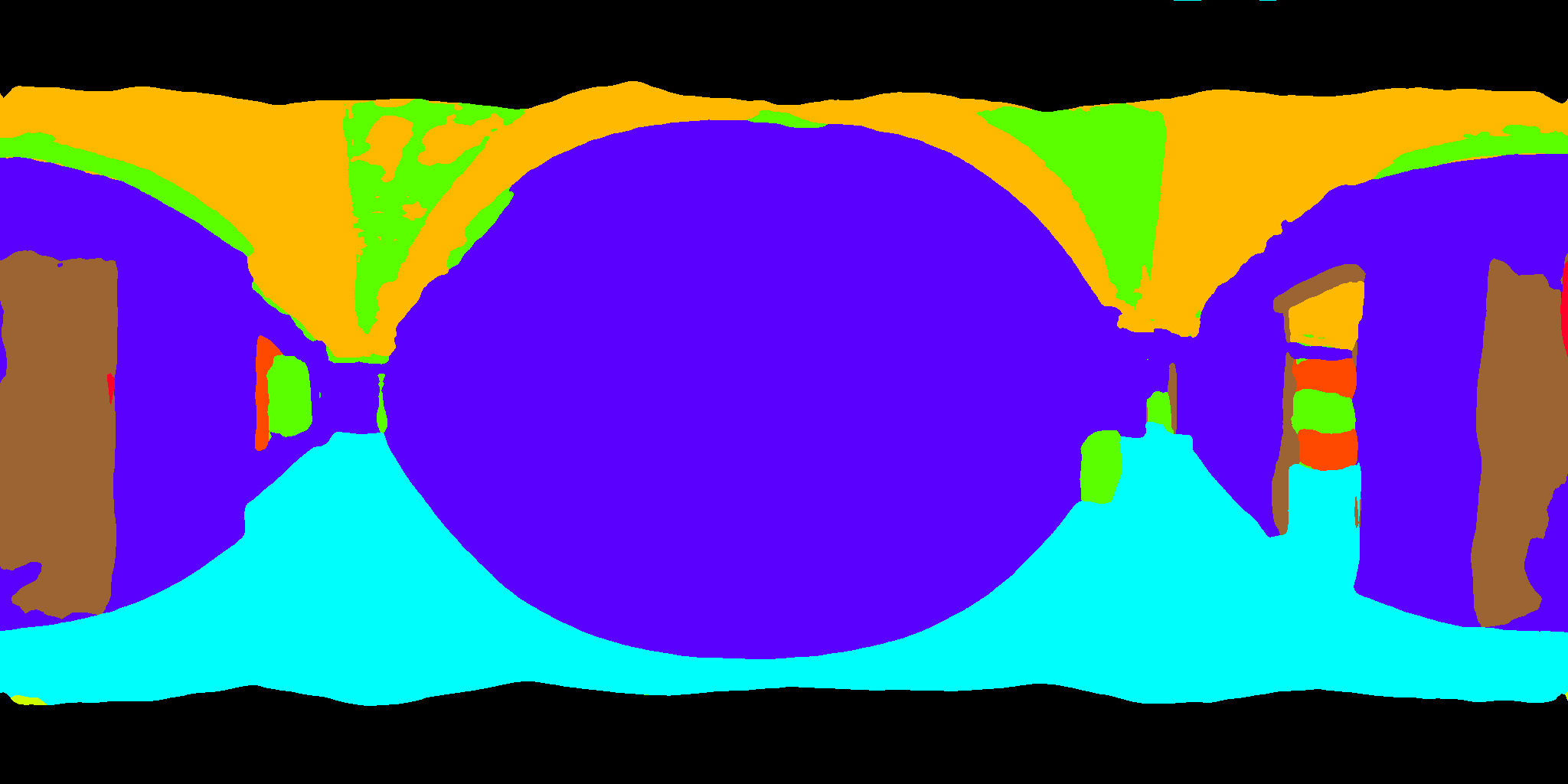}
        \caption{Our results}
        \label{sfig:source_ours_door1}
    \end{subfigure}
    
    \begin{subfigure}{0.23\linewidth}
        \centering
        \includegraphics[width=1\linewidth]{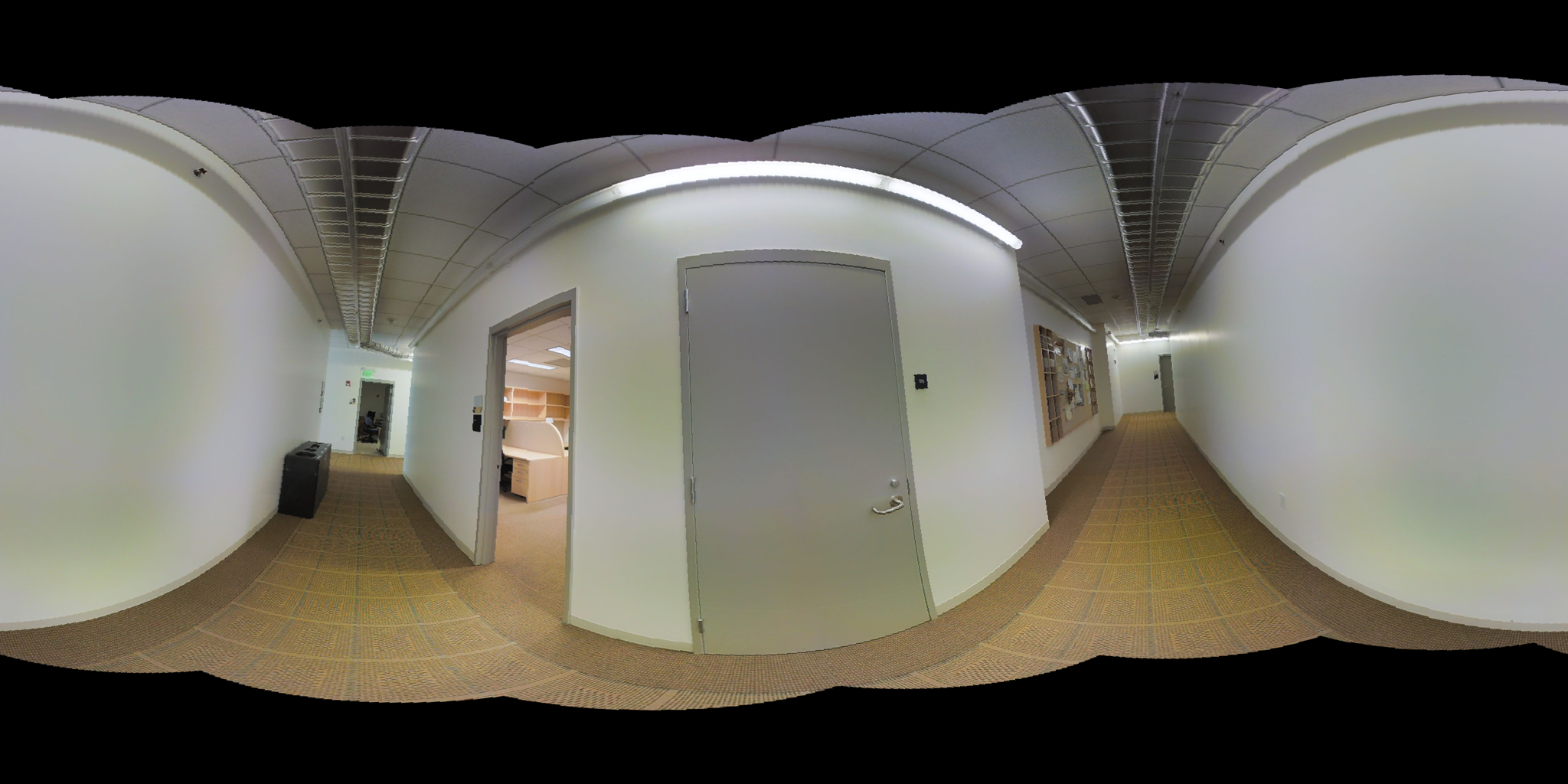}
        \caption{Rotated original picture}
        \label{sfig:rotated_door1}
    \end{subfigure}
    \begin{subfigure}{0.23\linewidth}
        \centering
        \includegraphics[width=1\linewidth]{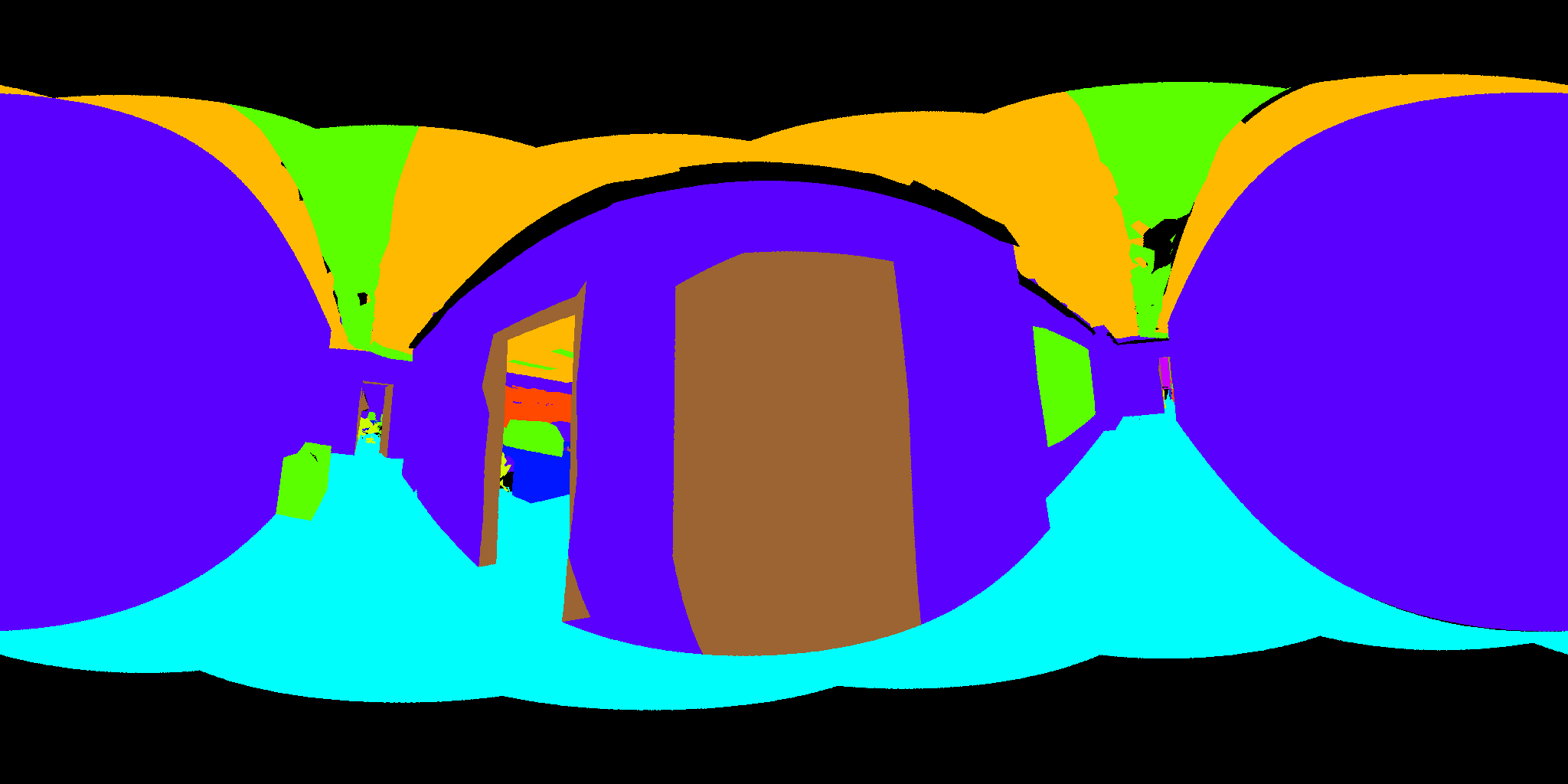}
        \caption{Rotated label}
        \label{sfig:rotated_GT_door1}
    \end{subfigure}
    \begin{subfigure}{0.23\linewidth}
        \centering
        \includegraphics[width=1\linewidth]{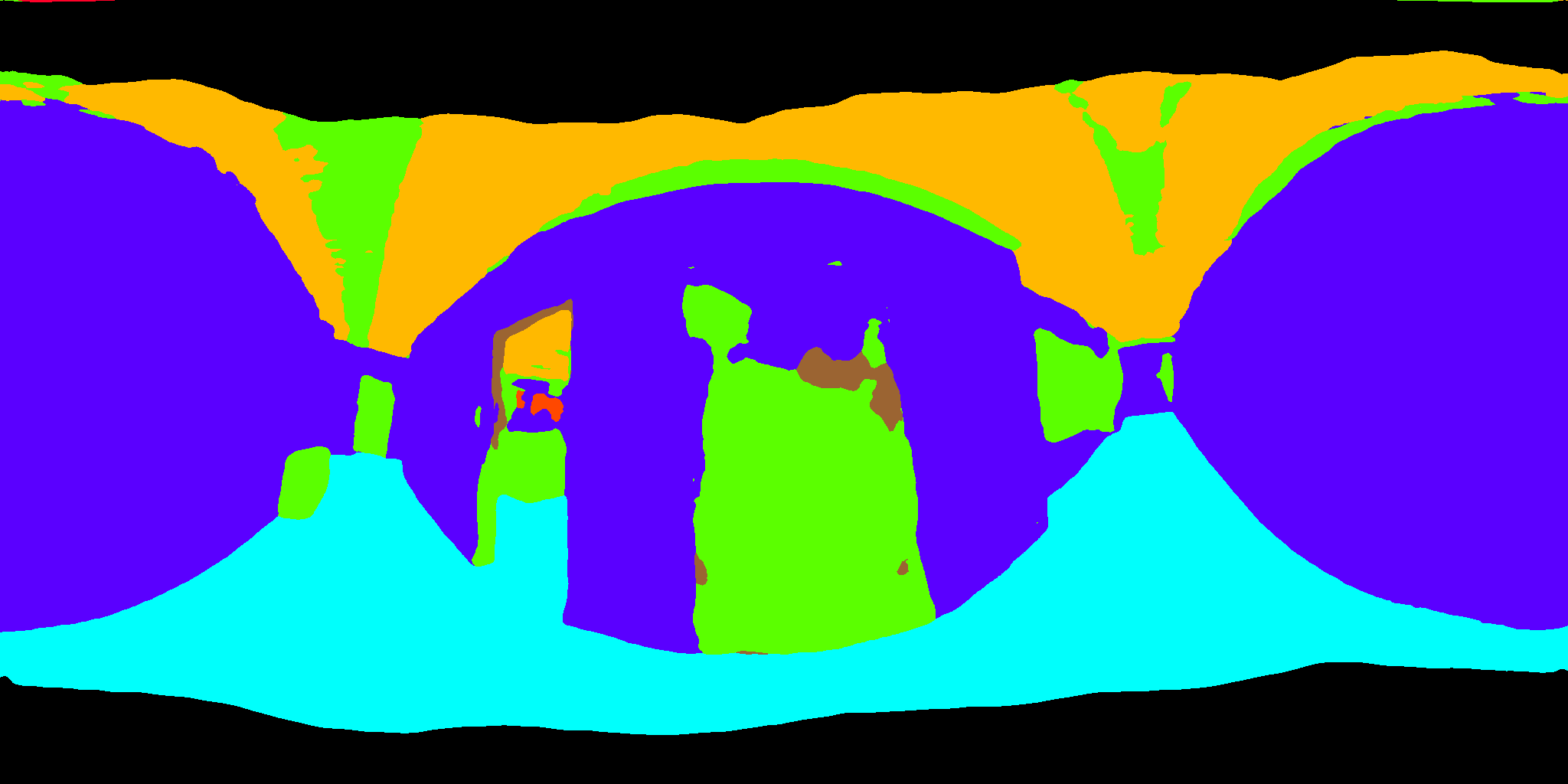}
        \caption{Baseline rotated results}
        \label{sfig:rotated_BL_door1}
    \end{subfigure}
    \begin{subfigure}{0.23\linewidth}
        \centering
        \includegraphics[width=1\linewidth]{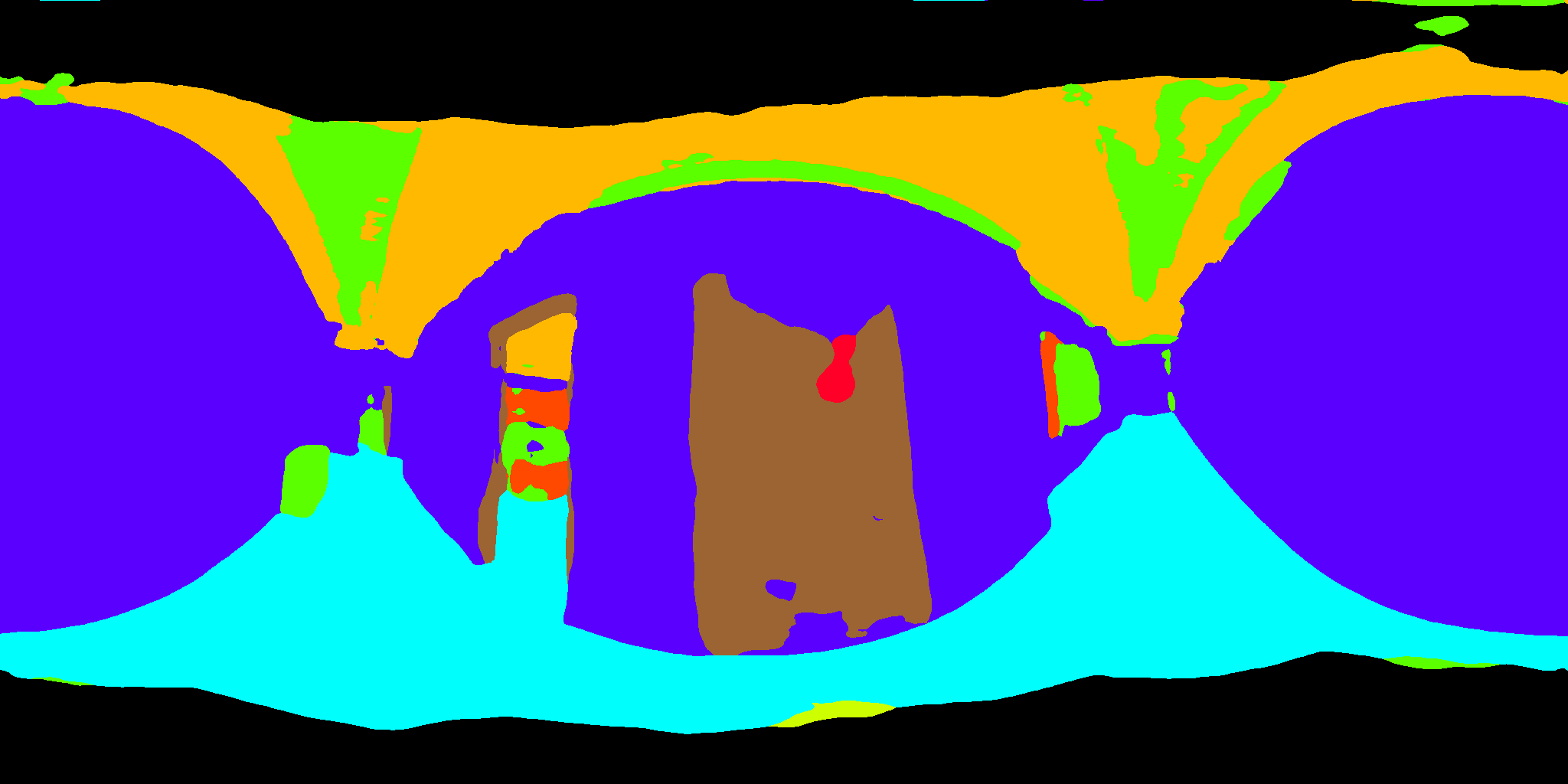}
        \caption{Our rotated results}
        \label{sfig:rotated_ours_door1}
    \end{subfigure}
    \caption{
    Visualization comparison of \ours and Trans4PASS+. 
    The rotation of the pitch / roll / yaw axis is $5^{\circ}$ / $5^{\circ}$ / $180^{\circ}$. 
    \ours gains the better results of semantic class ``door'' . 
    }
    \label{fig:Visualization_sup_door1}
    \centering
\end{figure*}

\begin{figure*}[tb]
    \centering
    \begin{subfigure}{0.23\linewidth}
        \centering
        \includegraphics[width=1\linewidth]{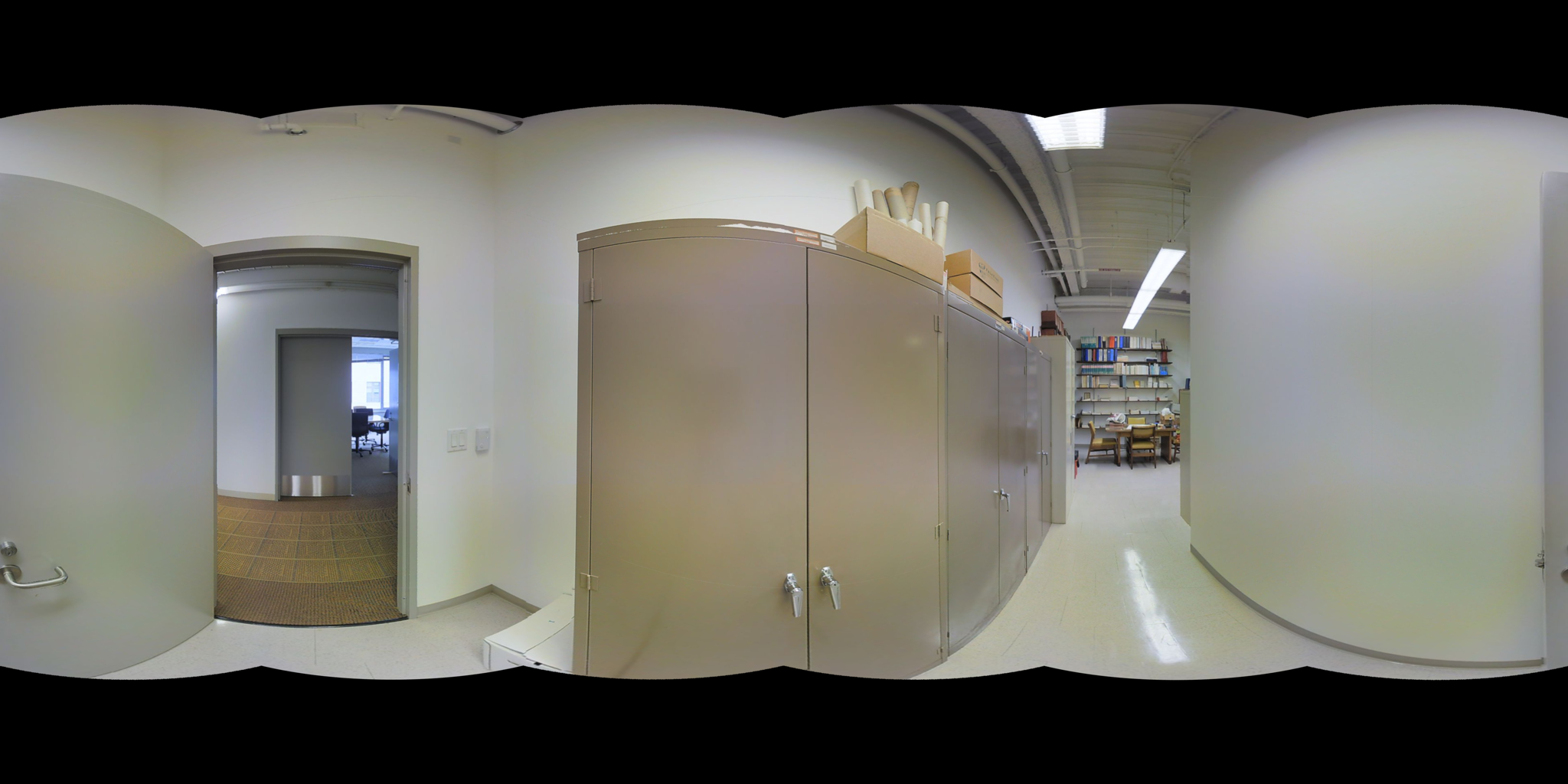}
        \caption{Original picture}
        \label{sfig:source_door2}
    \end{subfigure}
    \begin{subfigure}{0.23\linewidth}
        \centering
        \includegraphics[width=1\linewidth]{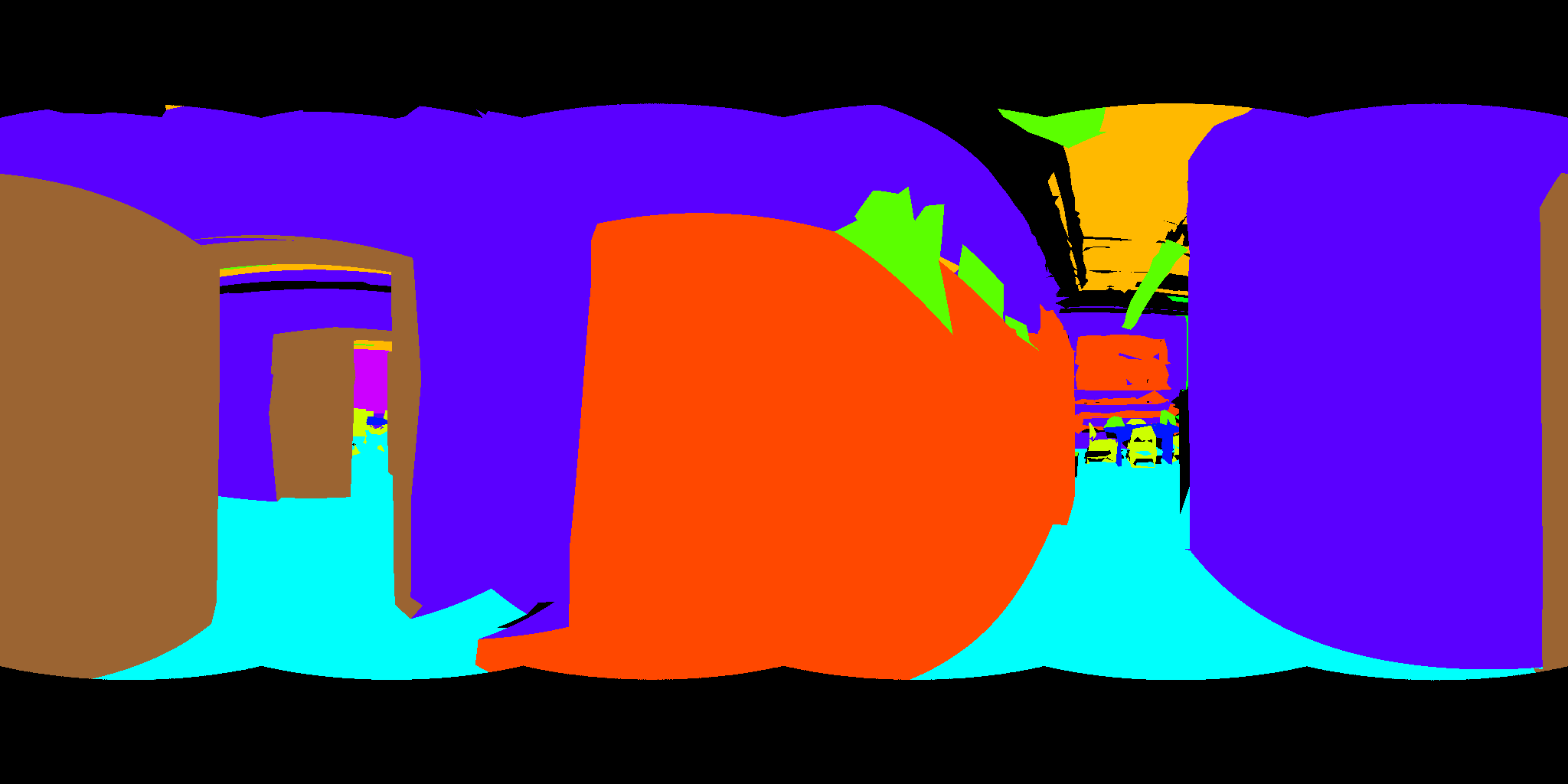}
        \caption{Label}
        \label{sfig:source_GT_door2}
    \end{subfigure}
    \begin{subfigure}{0.23\linewidth}
        \centering
        \includegraphics[width=1\linewidth]{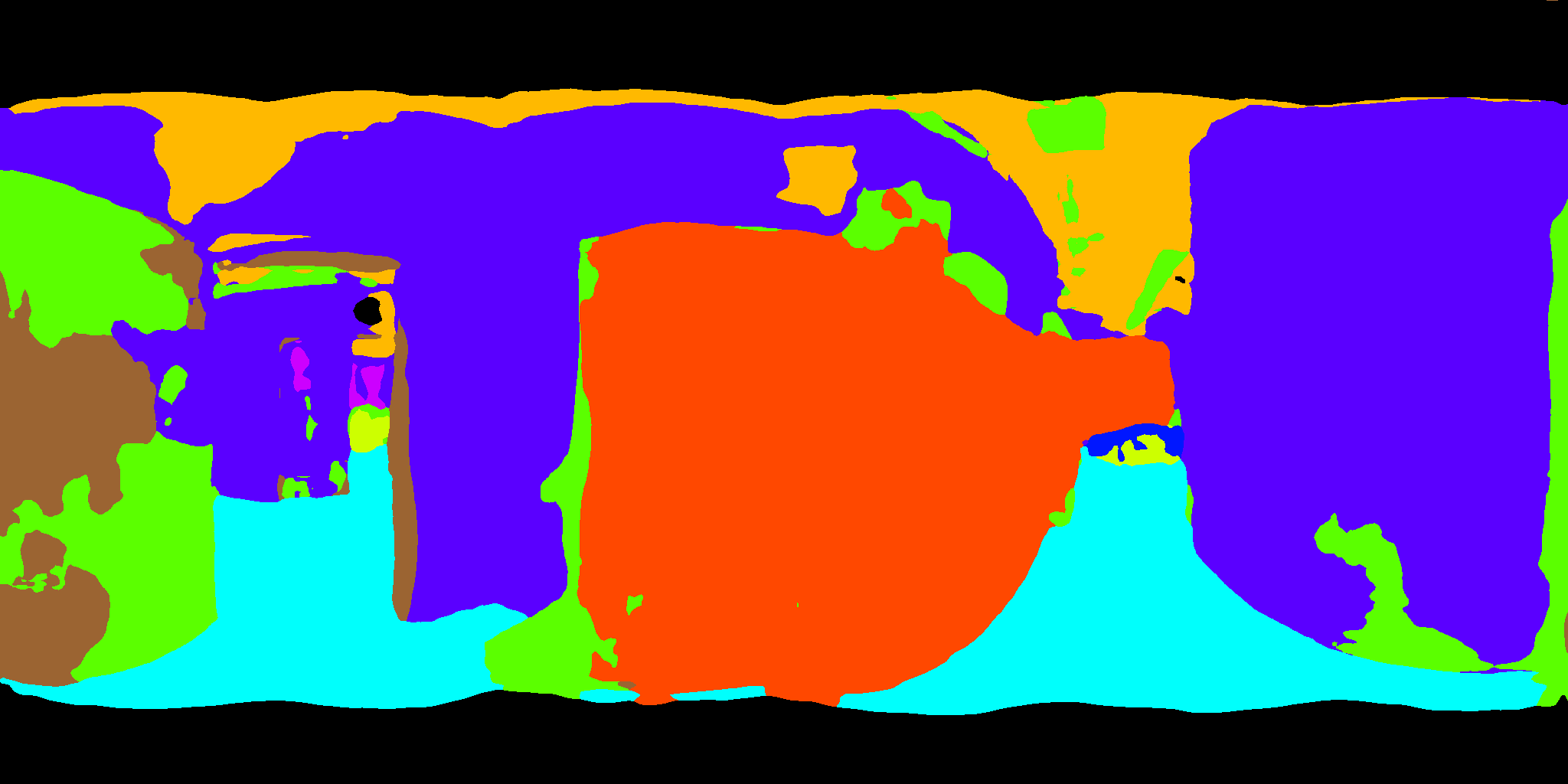}
        \caption{Baseline results}
        \label{sfig:source_BL_door2}
    \end{subfigure}
    \begin{subfigure}{0.23\linewidth}
        \centering
        \includegraphics[width=1\linewidth]{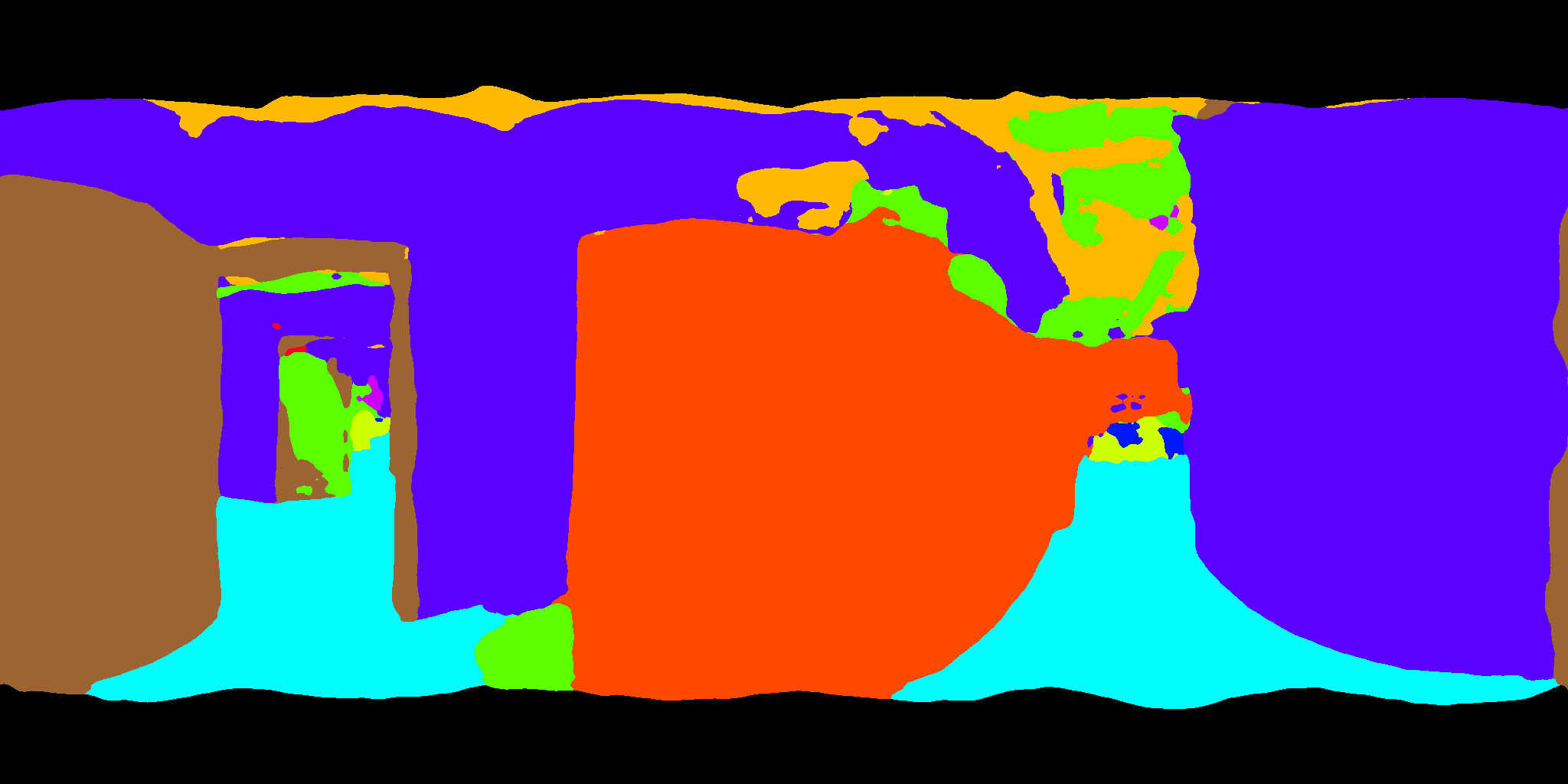}
        \caption{Our results}
        \label{sfig:source_ours_door2}
    \end{subfigure}
    
    \begin{subfigure}{0.23\linewidth}
        \centering
        \includegraphics[width=1\linewidth]{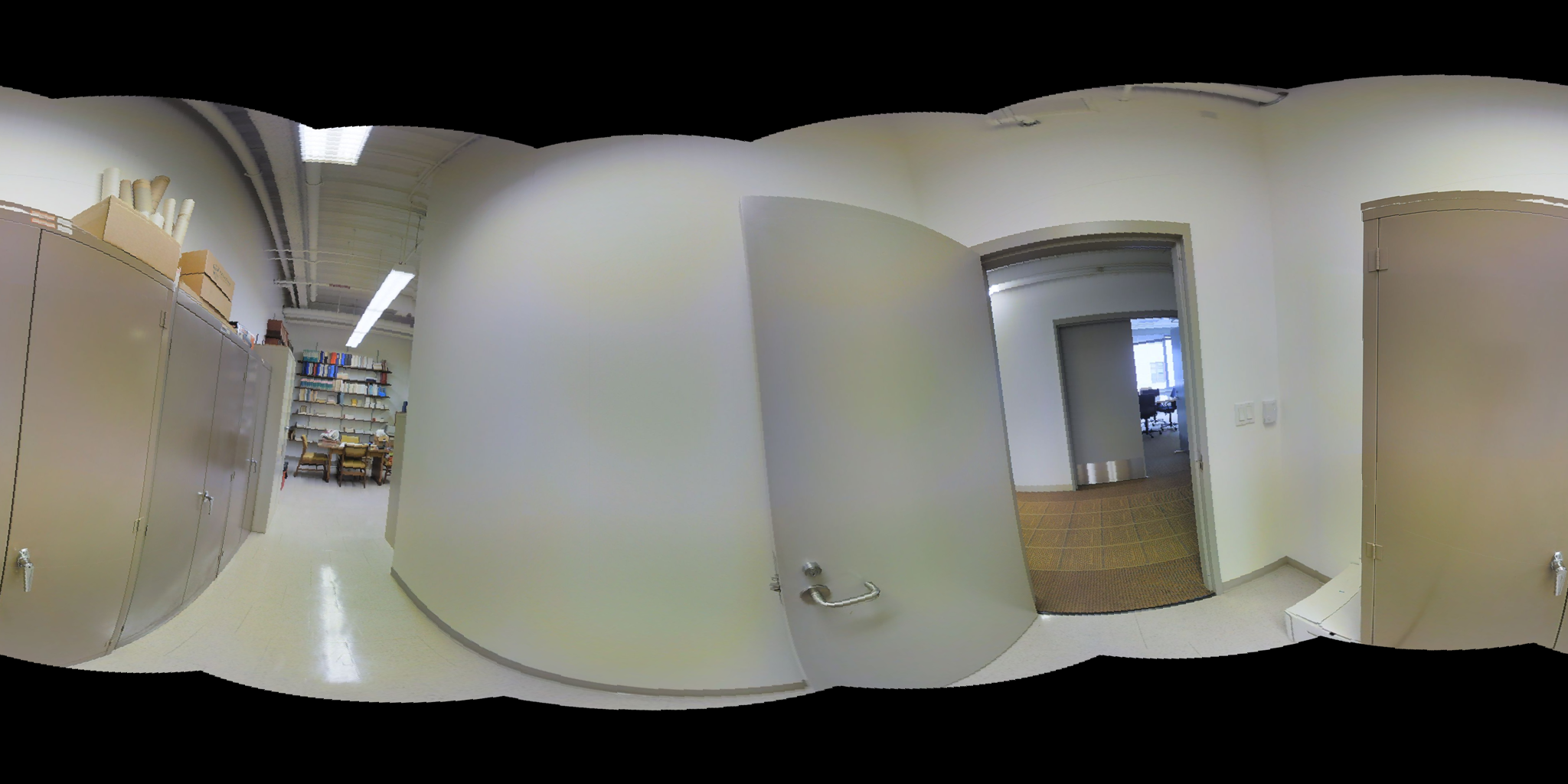}
        \caption{Rotated original picture}
        \label{sfig:rotated_door2}
    \end{subfigure}
    \begin{subfigure}{0.23\linewidth}
        \centering
        \includegraphics[width=1\linewidth]{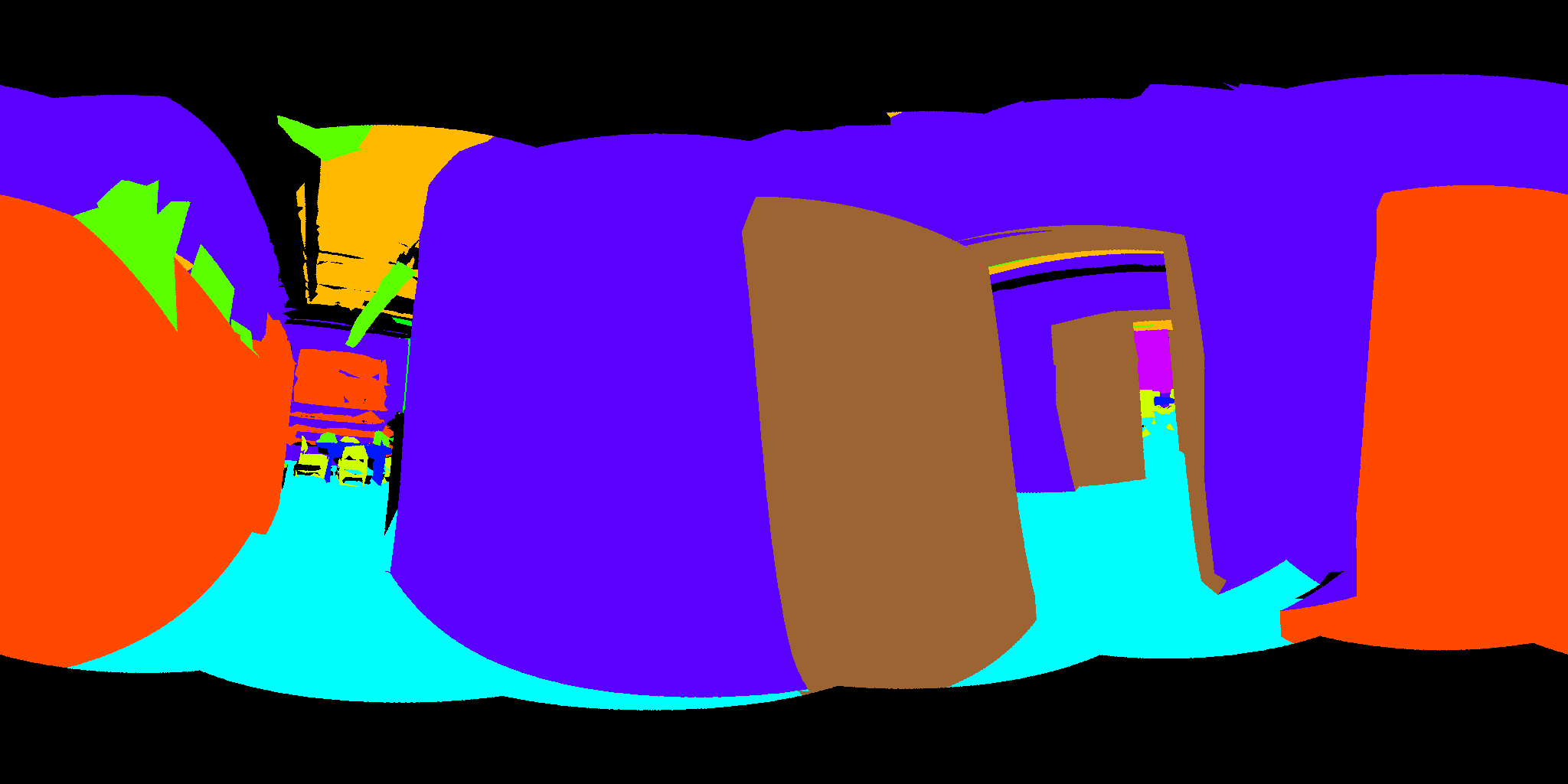}
        \caption{Rotated label}
        \label{sfig:rotated_GT_door2}
    \end{subfigure}
    \begin{subfigure}{0.23\linewidth}
        \centering
        \includegraphics[width=1\linewidth]{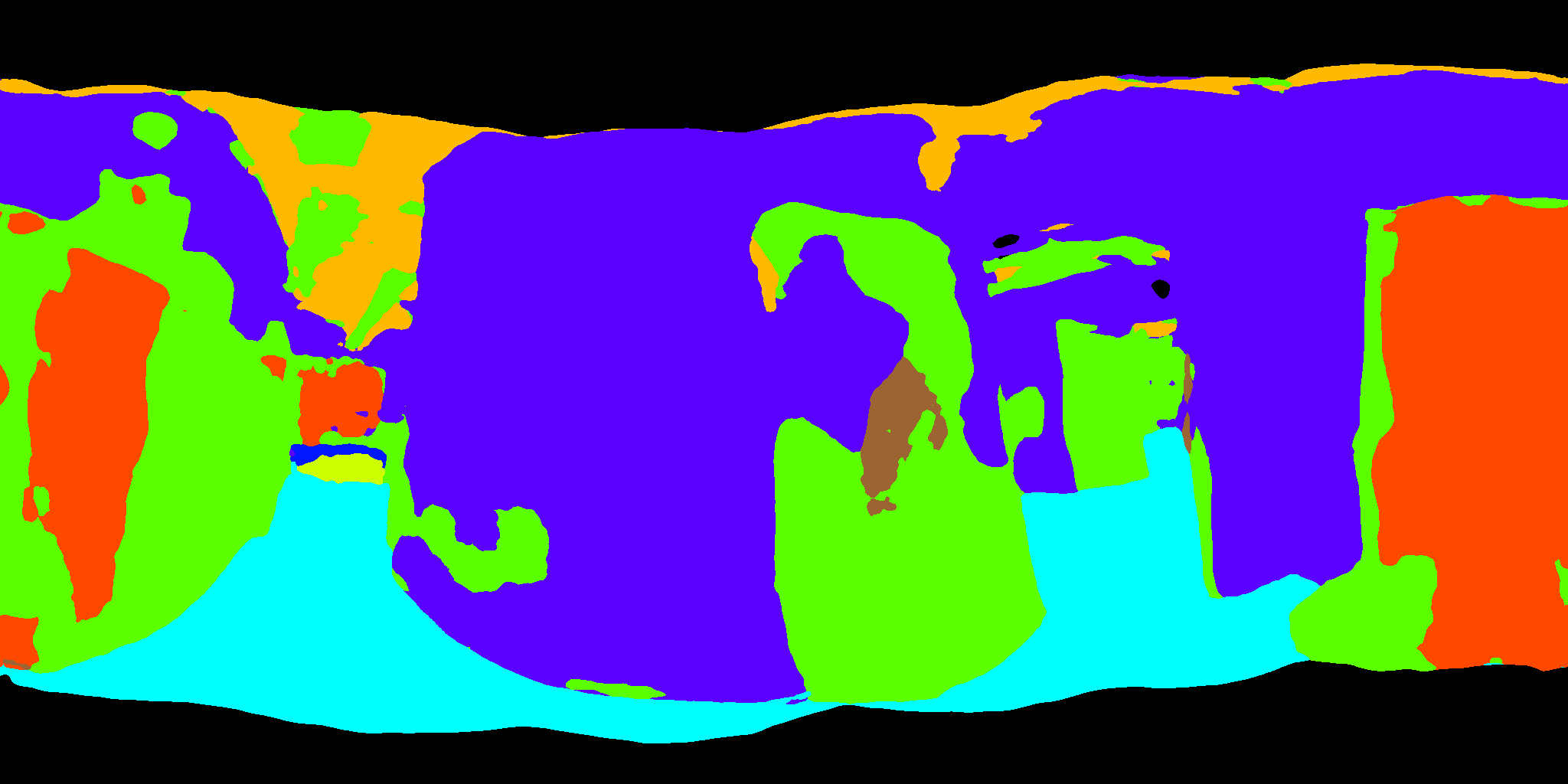}
        \caption{Baseline rotated results}
        \label{sfig:rotated_BL_door2}
    \end{subfigure}
    \begin{subfigure}{0.23\linewidth}
        \centering
        \includegraphics[width=1\linewidth]{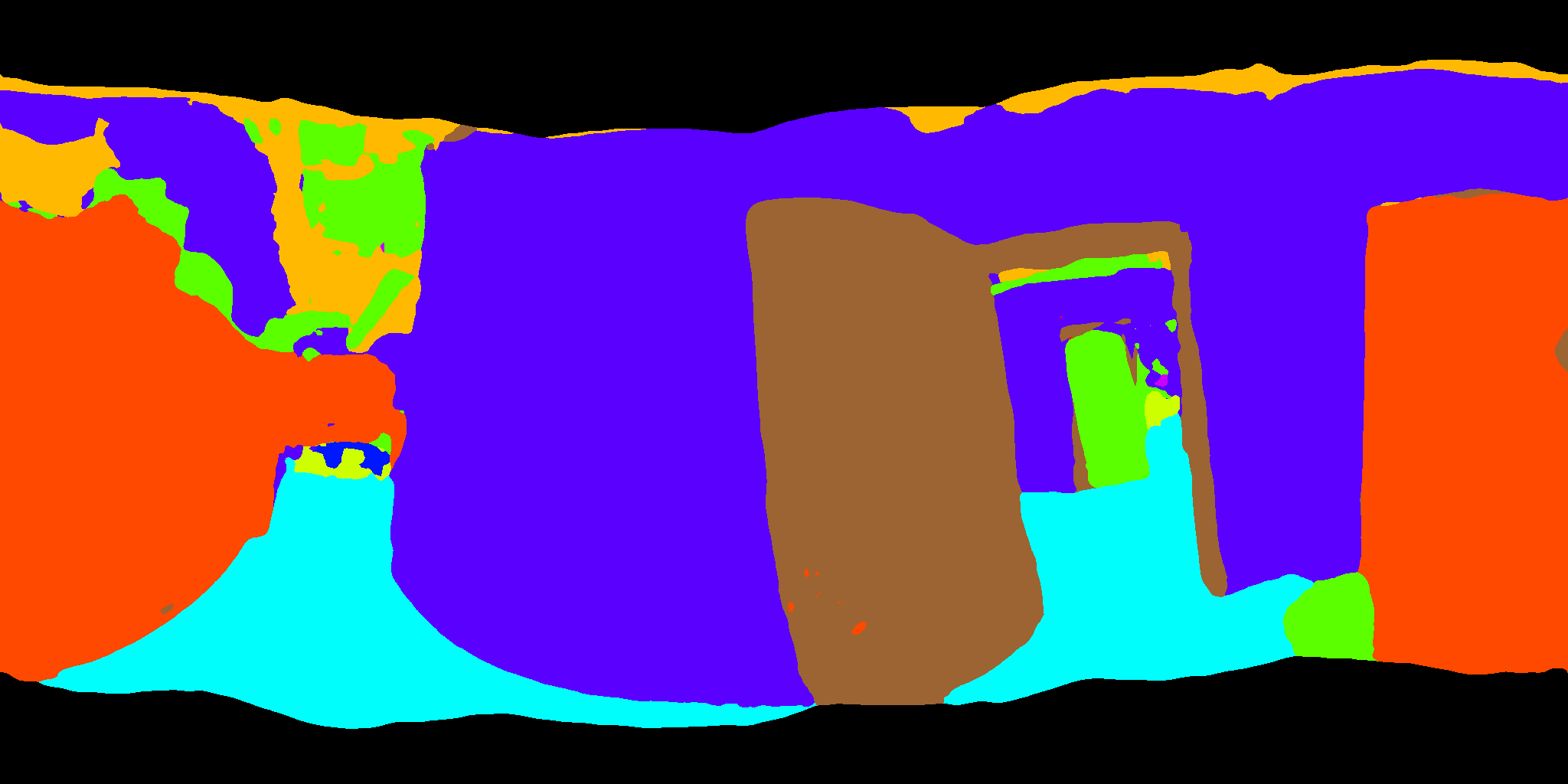}
        \caption{Our rotated results}
        \label{sfig:rotated_ours_door2}
    \end{subfigure}
    \caption{
    Visualization comparison of \ours and Trans4PASS+. 
    The rotation of the pitch / roll / yaw axis is $5^{\circ}$ / $5^{\circ}$ / $180^{\circ}$. 
    \ours gains the best results of semantic classes ``door'' and ``bookcase''. 
    }
    \label{fig:Visualization_sup_door2}
    \centering
\end{figure*}

\begin{figure*}[tb]
    \centering
    \begin{subfigure}{0.23\linewidth}
        \centering
        \includegraphics[width=1\linewidth]{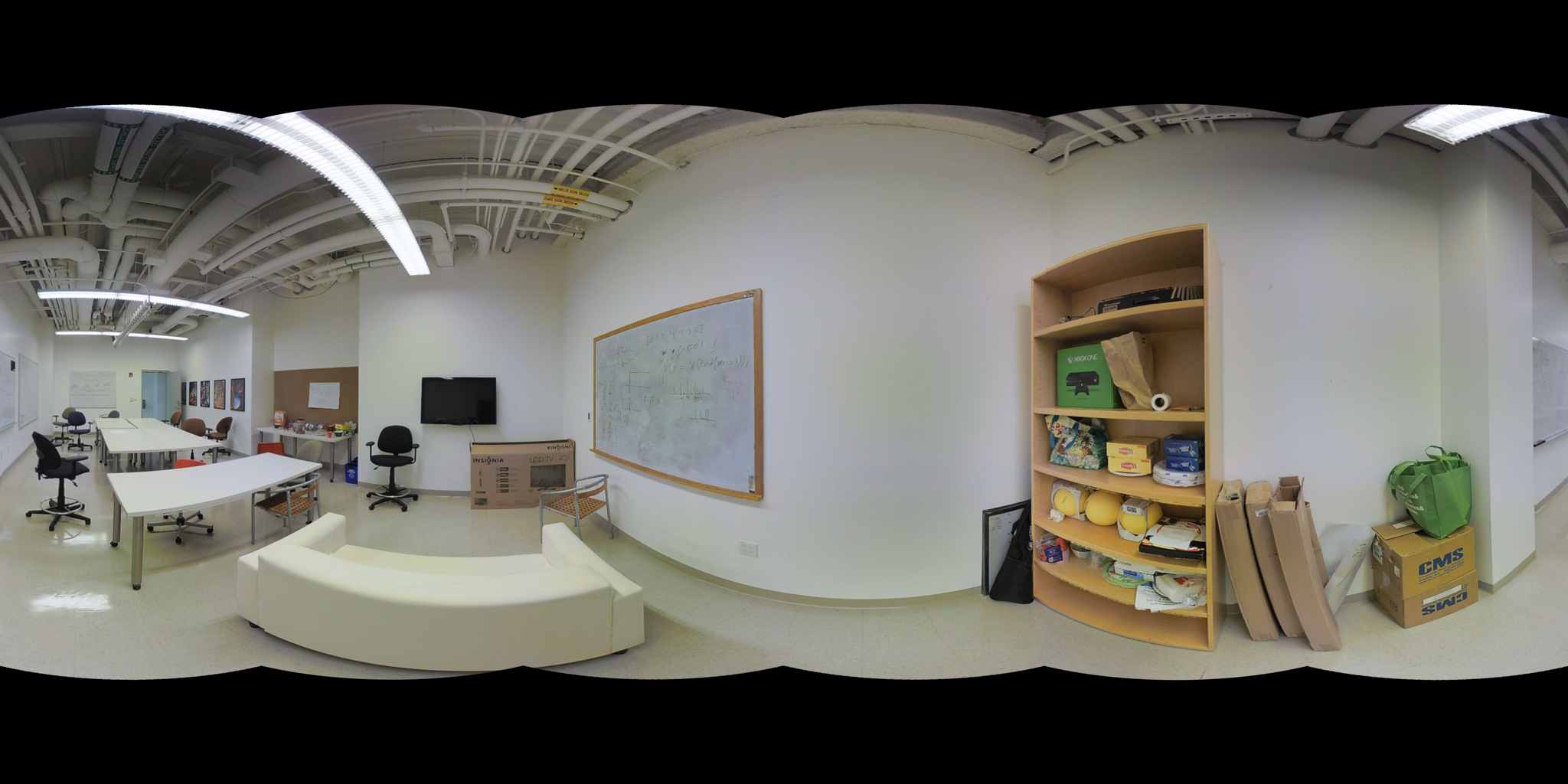}
        \caption{Original picture}
        \label{sfig:source_sofa}
    \end{subfigure}
    \begin{subfigure}{0.23\linewidth}
        \centering
        \includegraphics[width=1\linewidth]{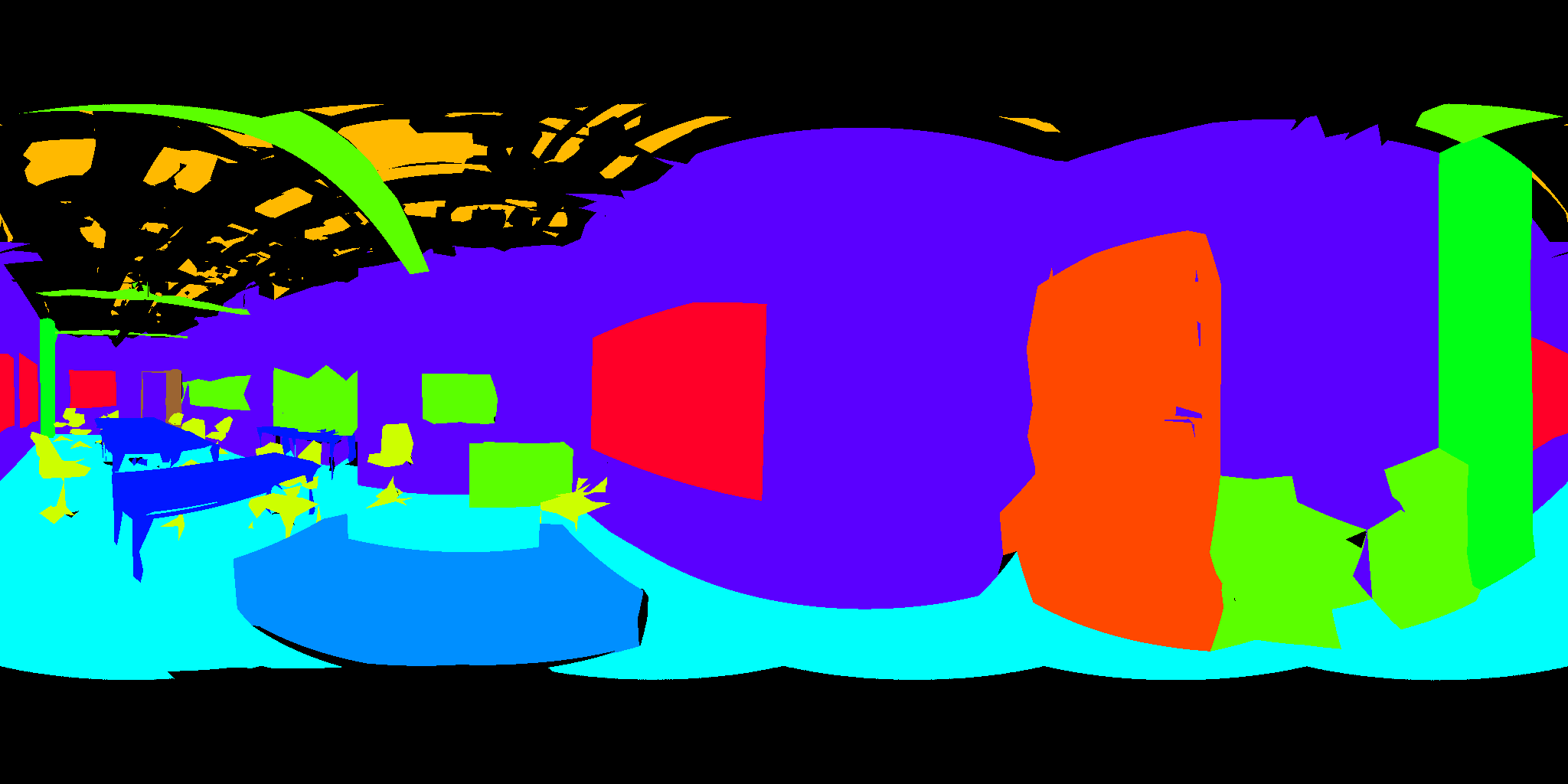}
        \caption{Label}
        \label{sfig:source_GT_sofa}
    \end{subfigure}
    \begin{subfigure}{0.23\linewidth}
        \centering
        \includegraphics[width=1\linewidth]{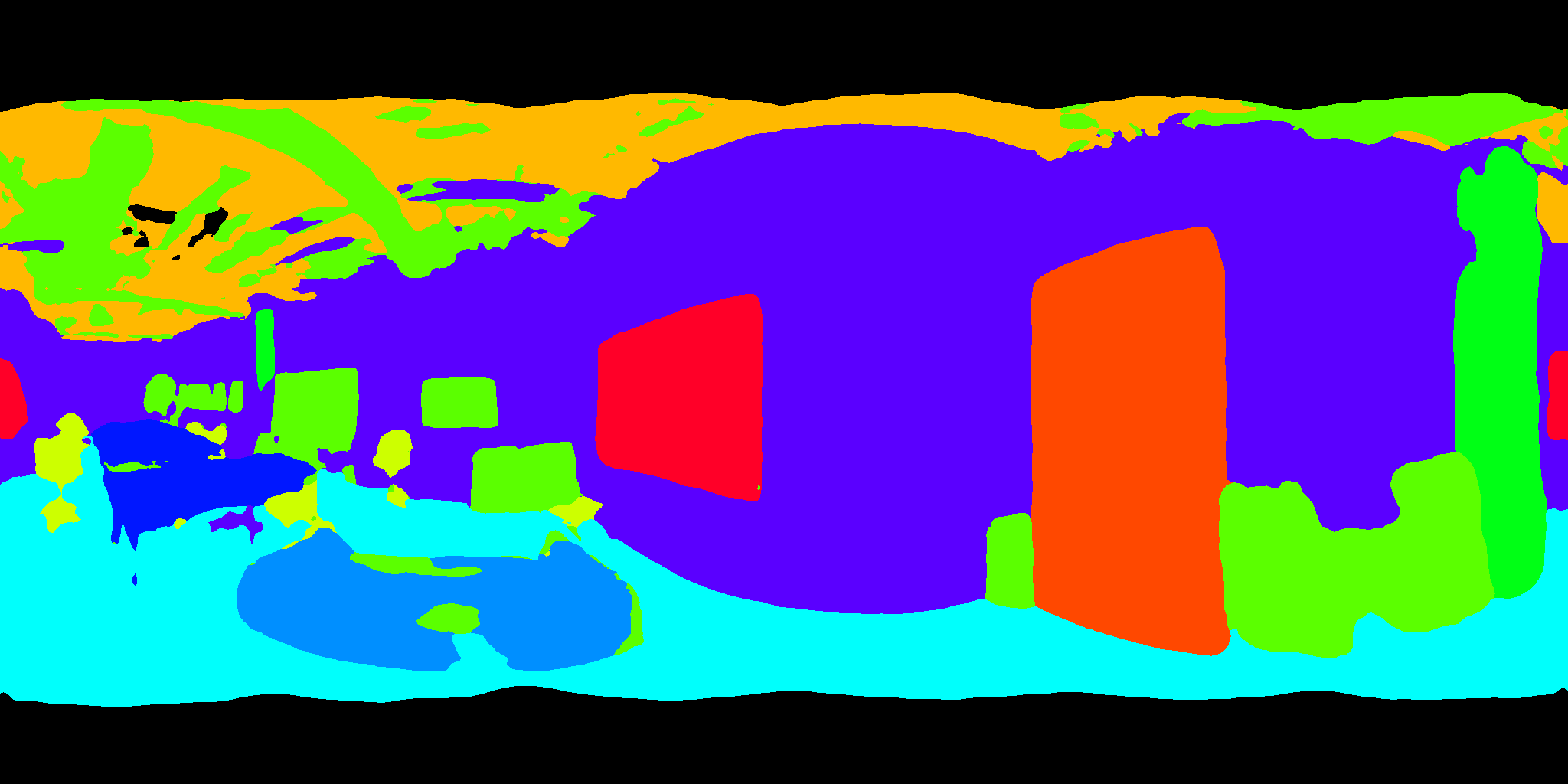}
        \caption{Baseline results}
        \label{sfig:source_BL_sofa}
    \end{subfigure}
    \begin{subfigure}{0.23\linewidth}
        \centering
        \includegraphics[width=1\linewidth]{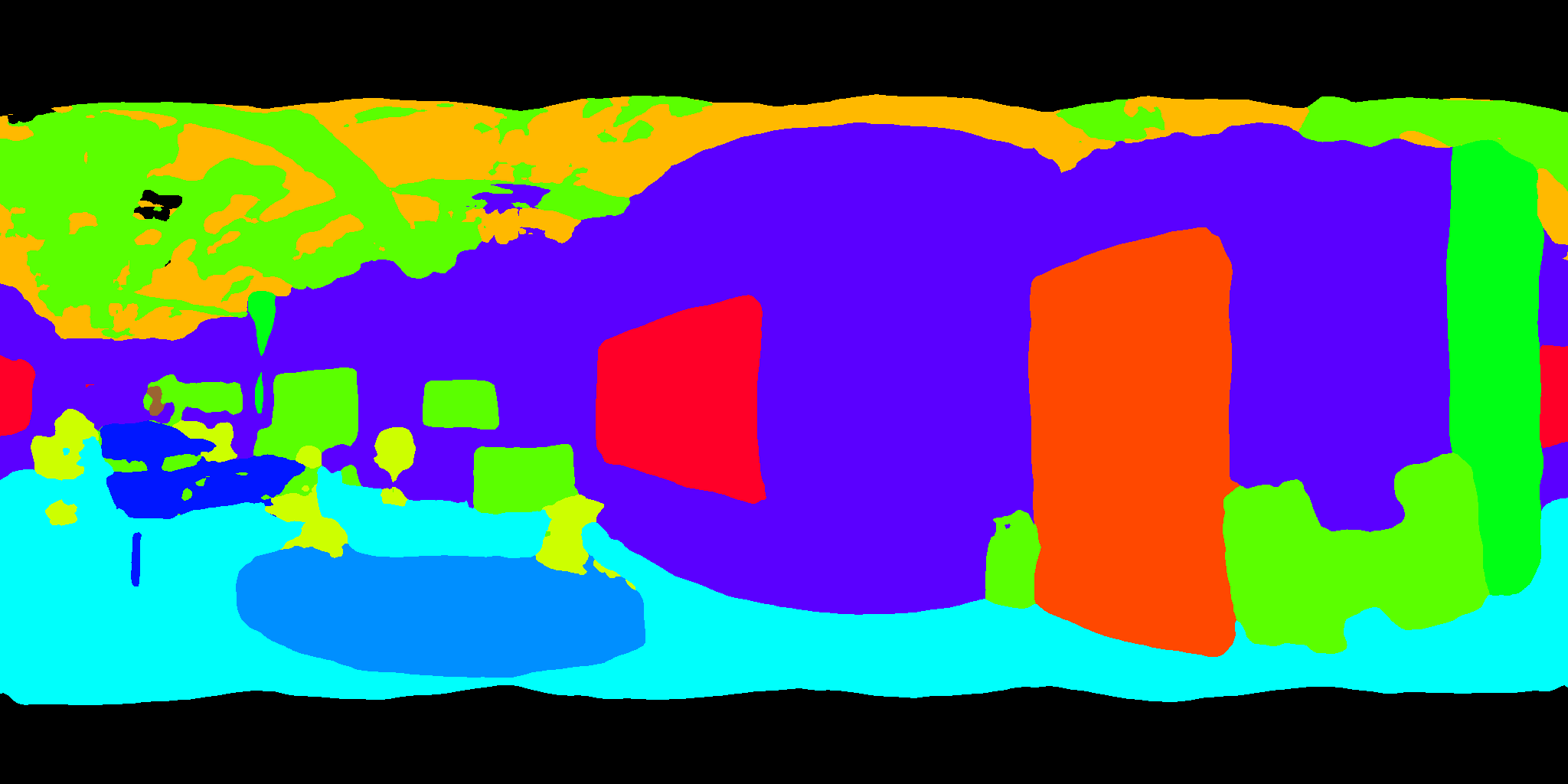}
        \caption{Our results}
        \label{sfig:source_ours_sofa}
    \end{subfigure}
    
    \begin{subfigure}{0.23\linewidth}
        \centering
        \includegraphics[width=1\linewidth]{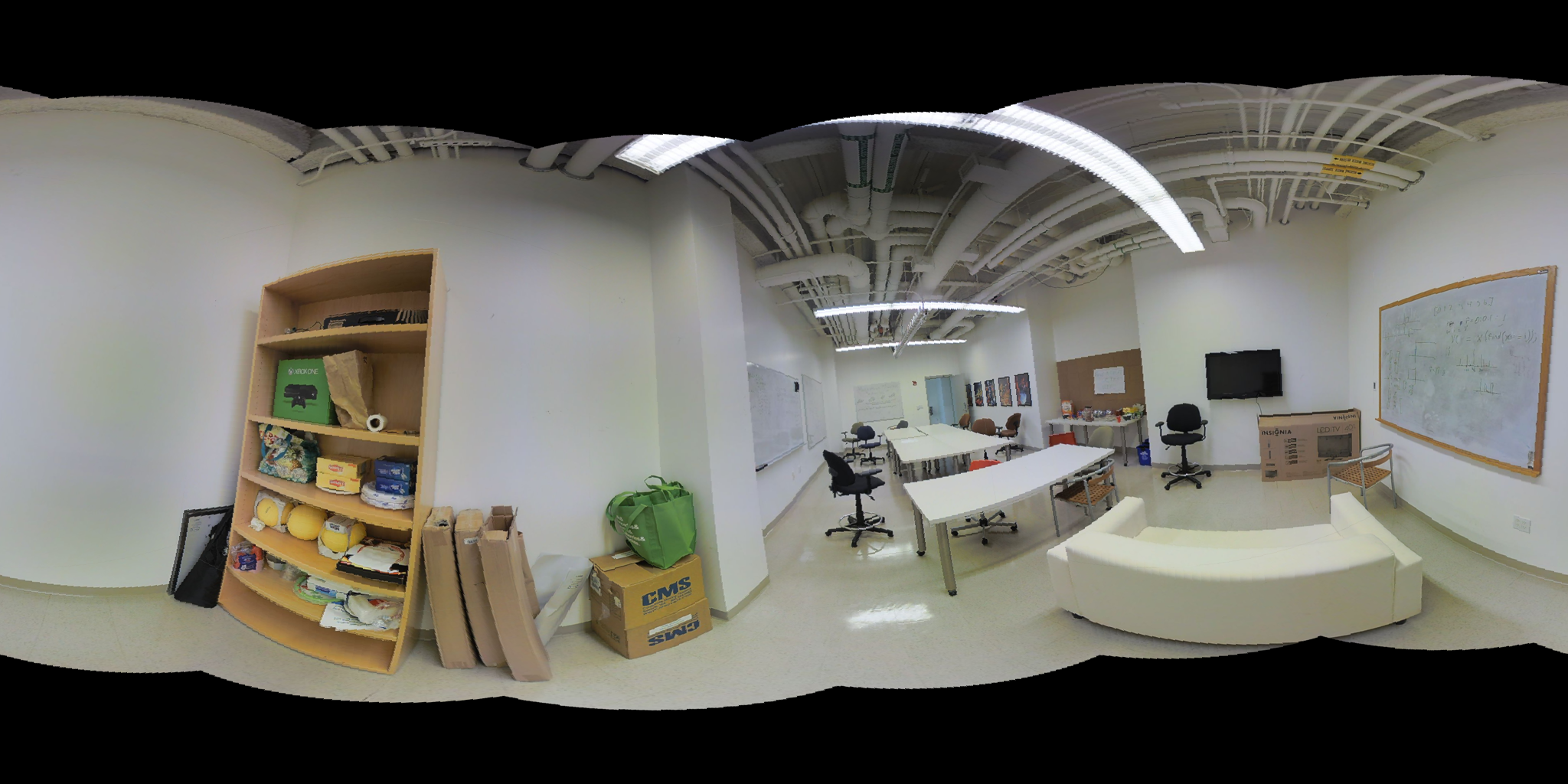}
        \caption{Rotated original picture}
        \label{sfig:rotated_sofa}
    \end{subfigure}
    \begin{subfigure}{0.23\linewidth}
        \centering
        \includegraphics[width=1\linewidth]{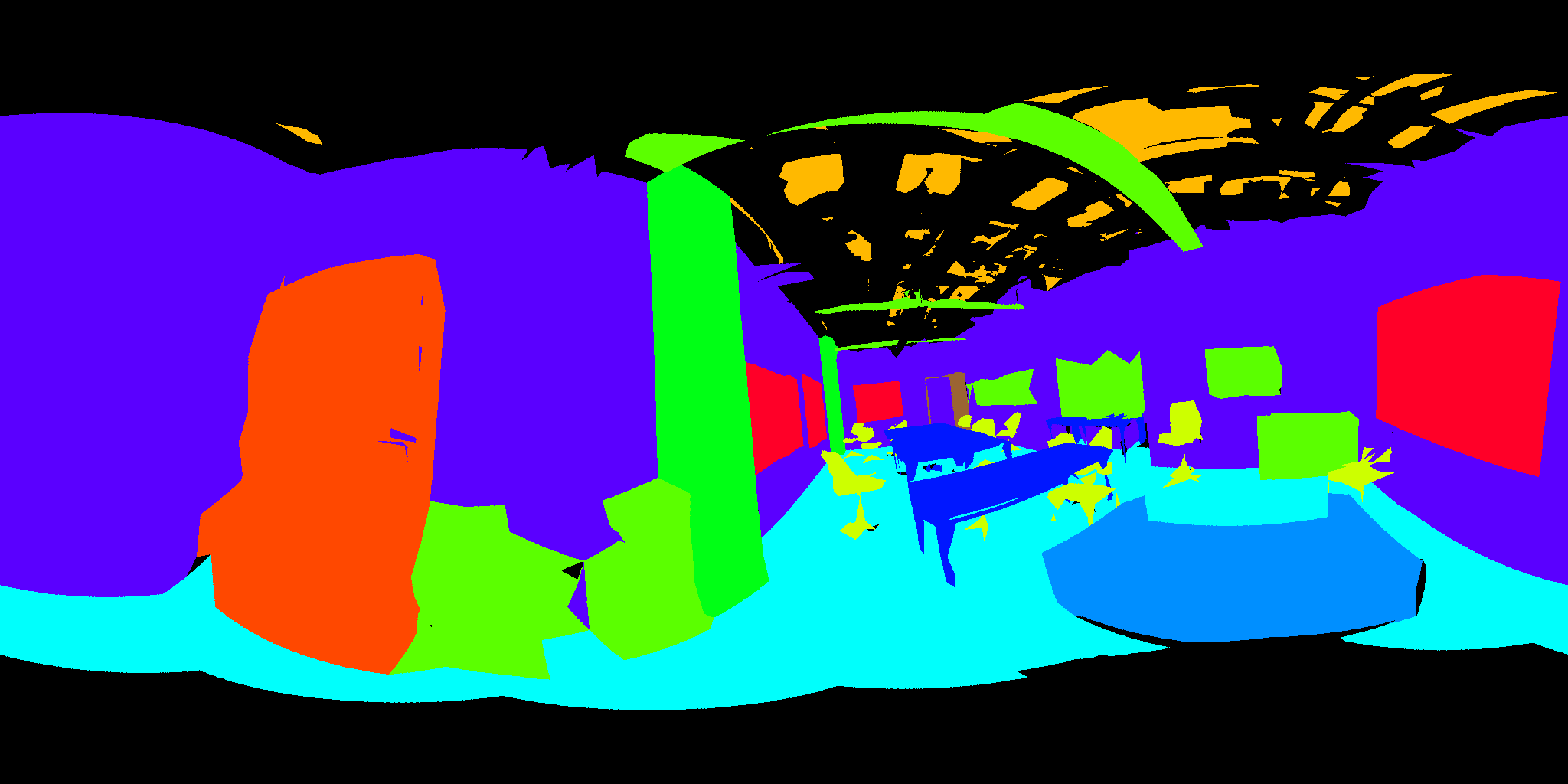}
        \caption{Rotated label}
        \label{sfig:rotated_GT_sofa}
    \end{subfigure}
    \begin{subfigure}{0.23\linewidth}
        \centering
        \includegraphics[width=1\linewidth]{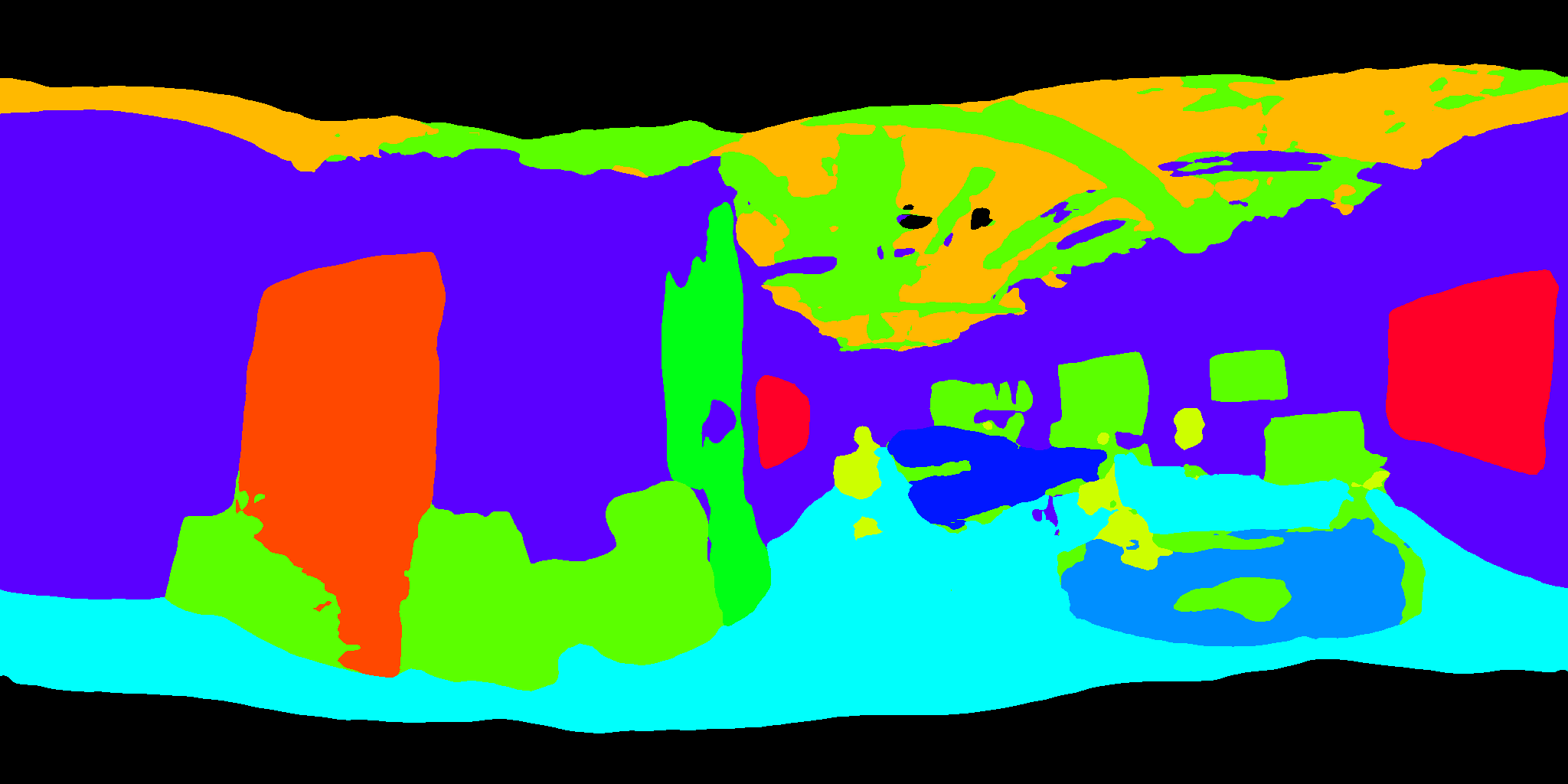}
        \caption{Baseline rotated results}
        \label{sfig:rotated_BL_sofa}
    \end{subfigure}
    \begin{subfigure}{0.23\linewidth}
        \centering
        \includegraphics[width=1\linewidth]{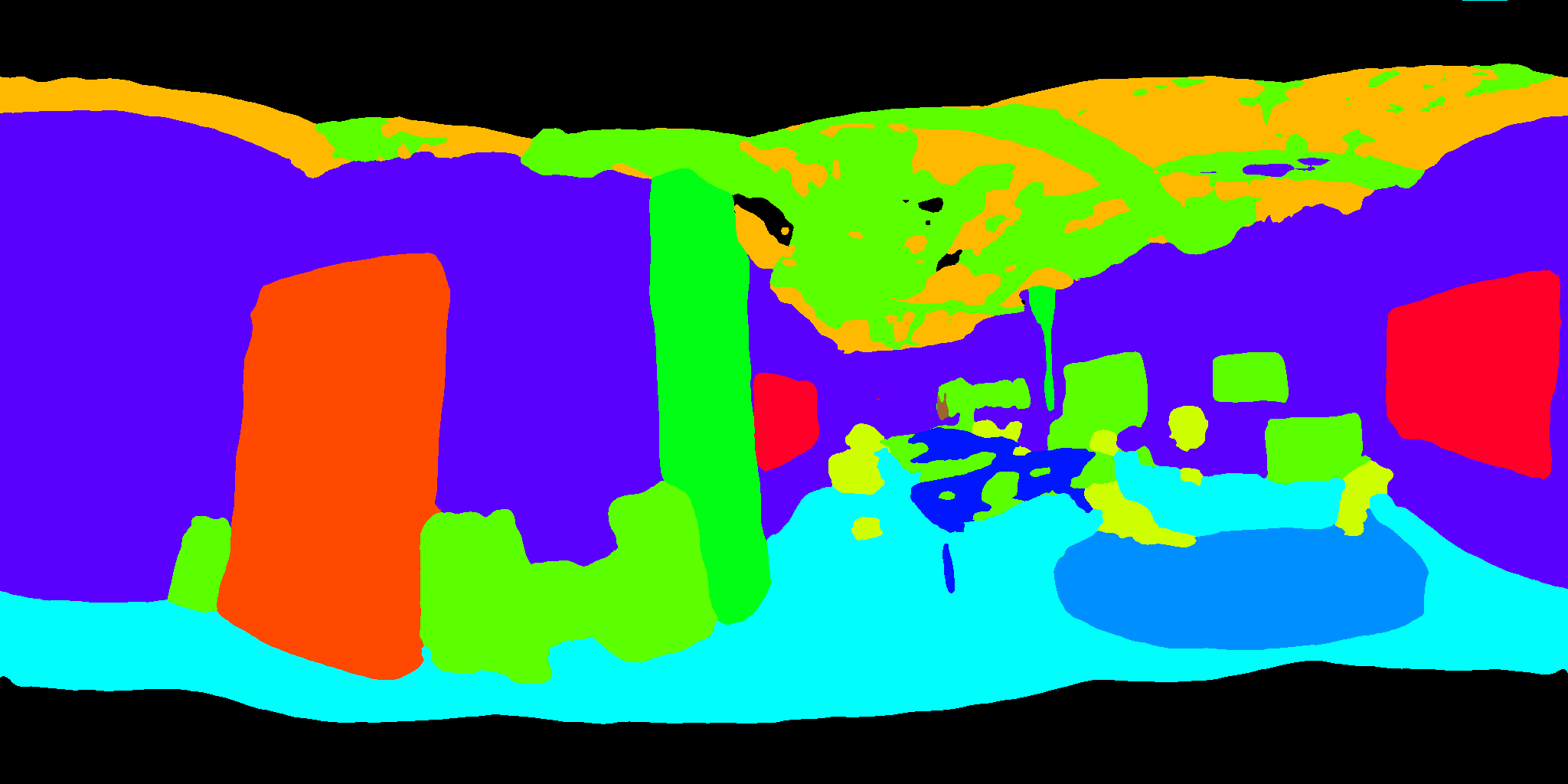}
        \caption{Our rotated results}
        \label{sfig:rotated_ours_sofa}
    \end{subfigure}
    \caption{
    Visualization comparison of \ours and Trans4PASS+. 
    The rotation of the pitch / roll / yaw axis is $5^{\circ}$ / $5^{\circ}$ / $180^{\circ}$. 
    \ours gains the better results of semantic class ``sofa'' . 
    }
    \label{fig:Visualization_sup_sofa}
    \centering
\end{figure*}

\begin{figure*}[tb]
    \centering
    \begin{subfigure}{0.23\linewidth}
        \centering
        \includegraphics[width=1\linewidth]{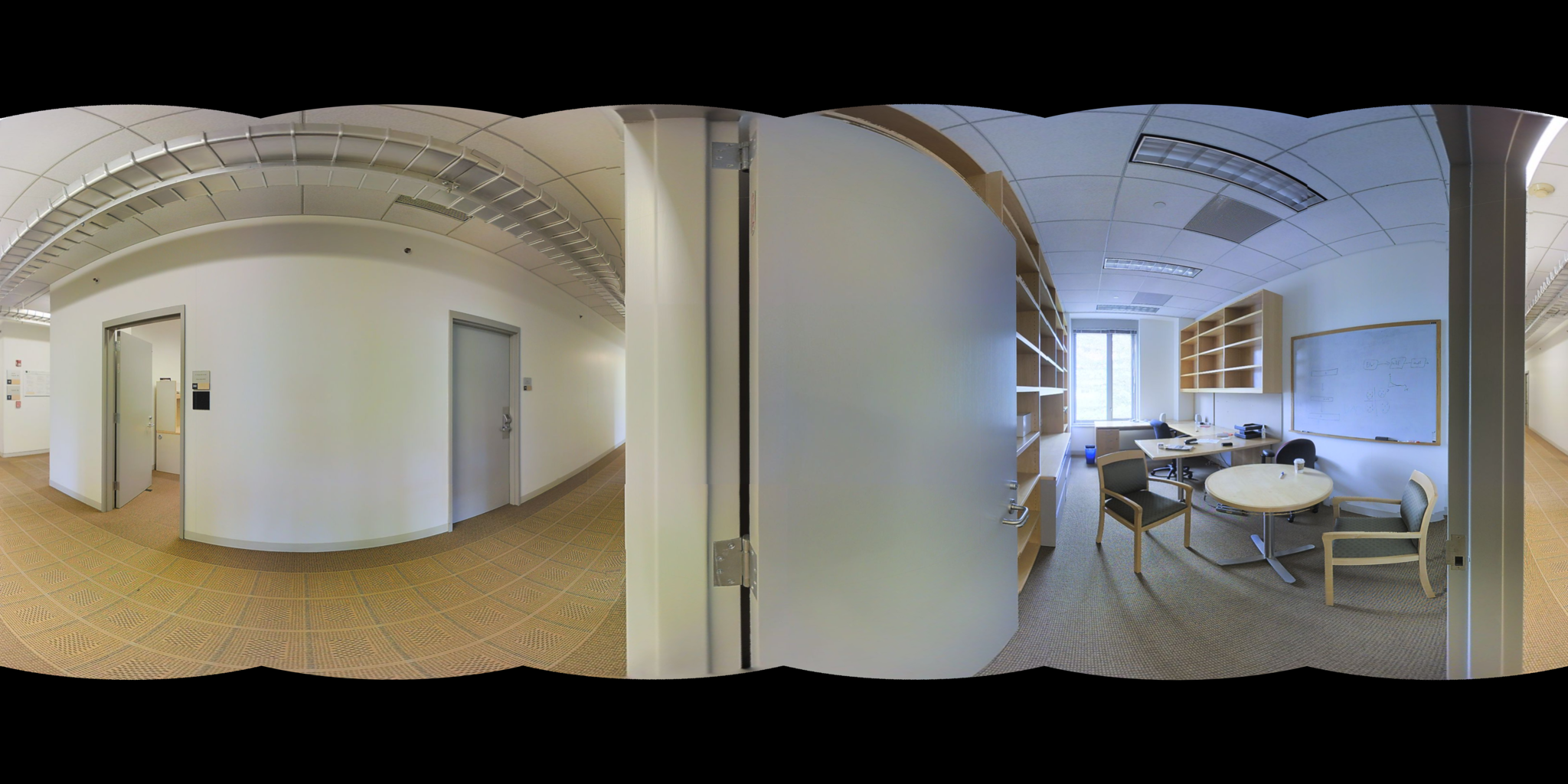}
        \caption{Original picture}
        \label{sfig:source_win}
    \end{subfigure}
    \begin{subfigure}{0.23\linewidth}
        \centering
        \includegraphics[width=1\linewidth]{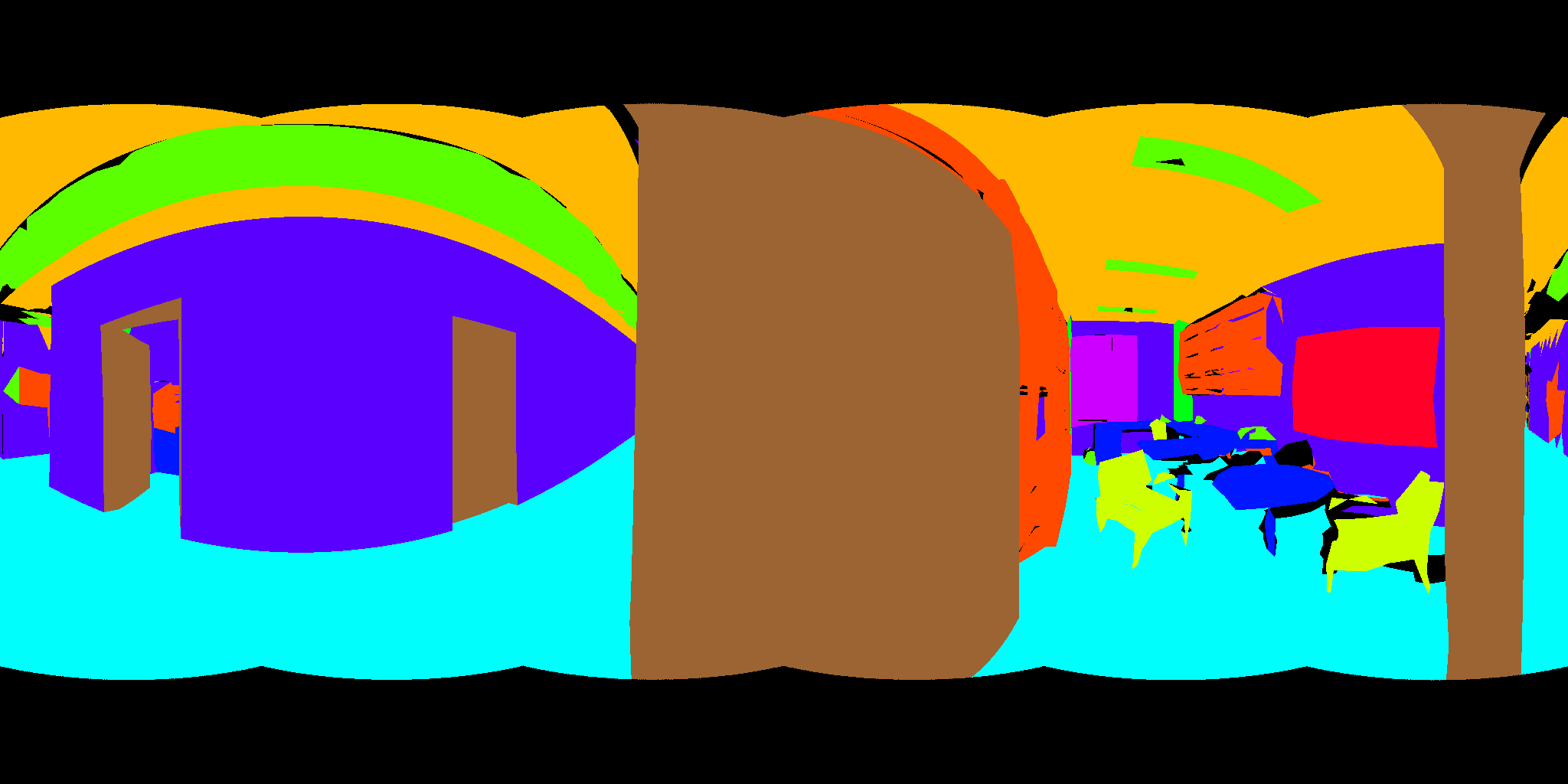}
        \caption{Label}
        \label{sfig:source_GT_win}
    \end{subfigure}
    \begin{subfigure}{0.23\linewidth}
        \centering
        \includegraphics[width=1\linewidth]{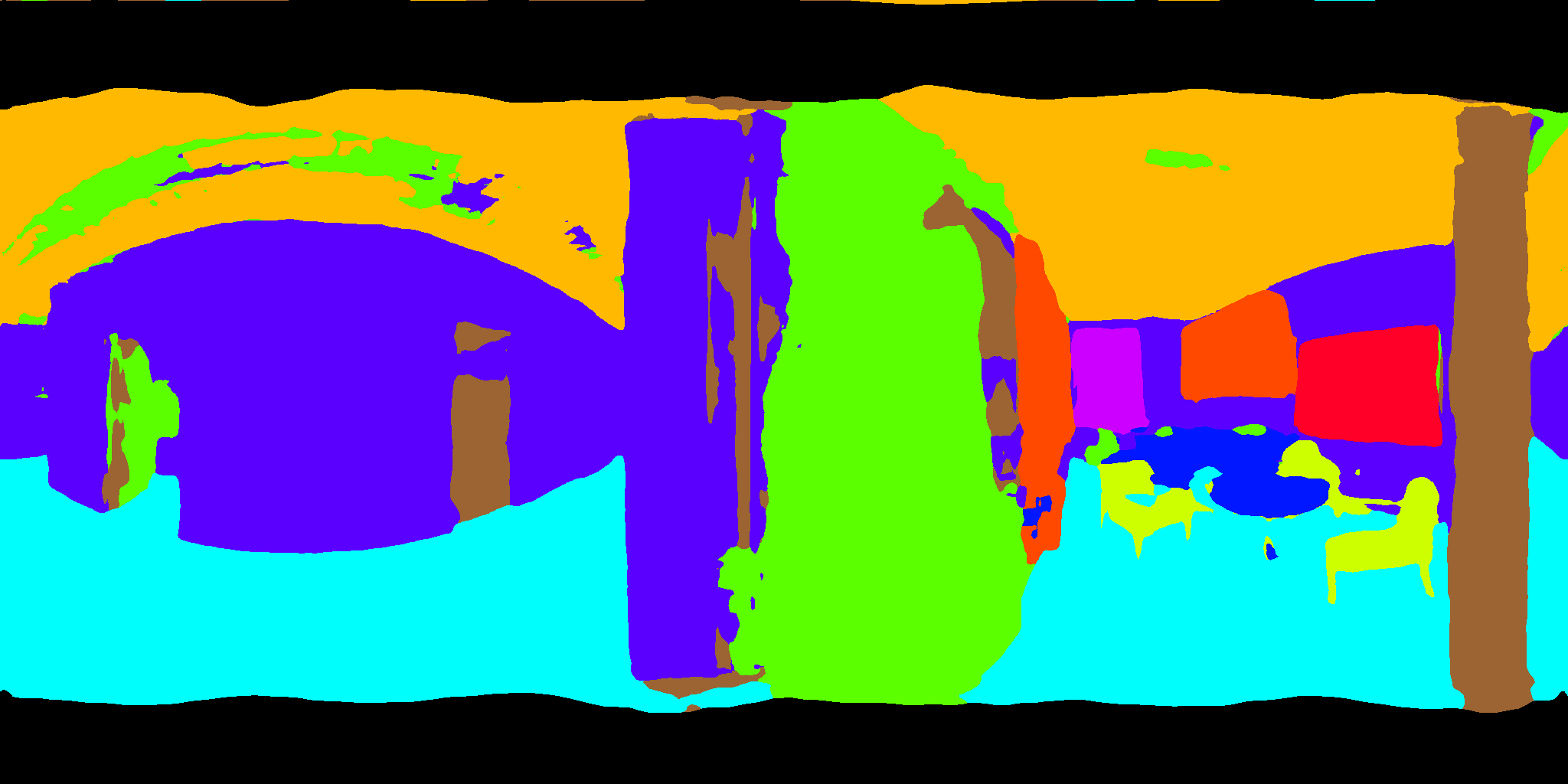}
        \caption{Baseline results}
        \label{sfig:source_BL_win}
    \end{subfigure}
    \begin{subfigure}{0.23\linewidth}
        \centering
        \includegraphics[width=1\linewidth]{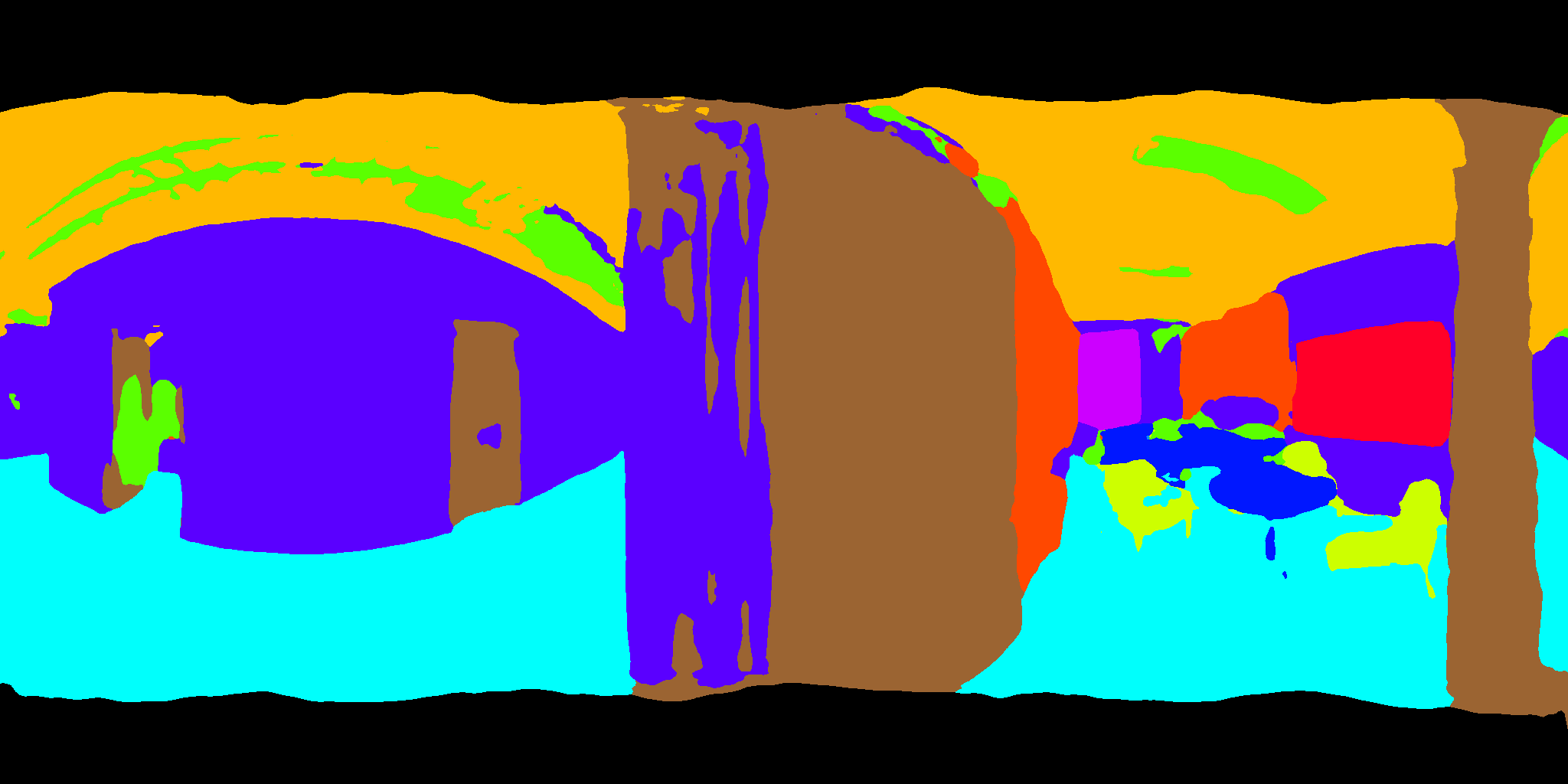}
        \caption{Our results}
        \label{sfig:source_ours_win}
    \end{subfigure}
    
    \begin{subfigure}{0.23\linewidth}
        \centering
        \includegraphics[width=1\linewidth]{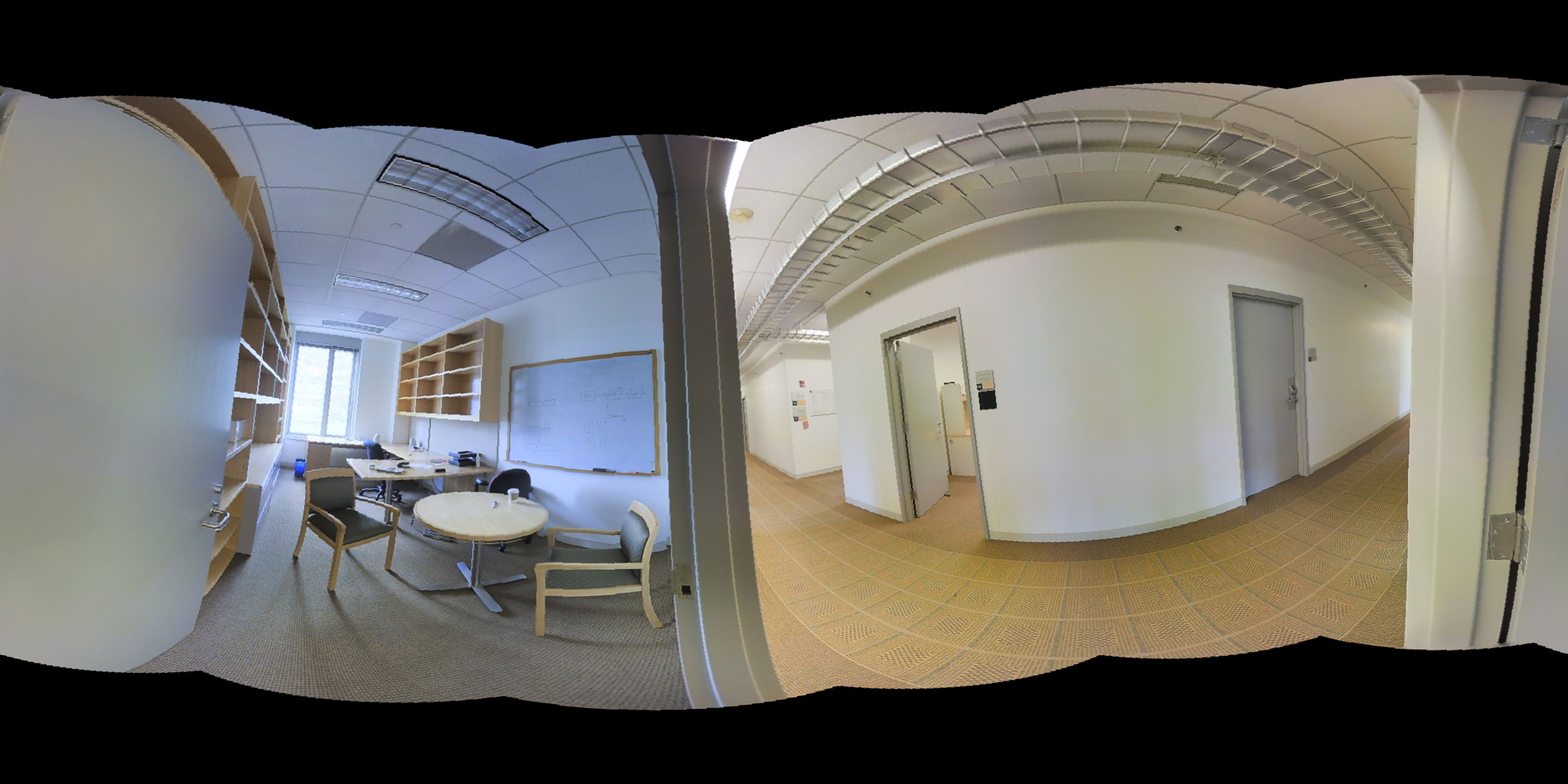}
        \caption{Rotated original picture}
        \label{sfig:rotated_win}
    \end{subfigure}
    \begin{subfigure}{0.23\linewidth}
        \centering
        \includegraphics[width=1\linewidth]{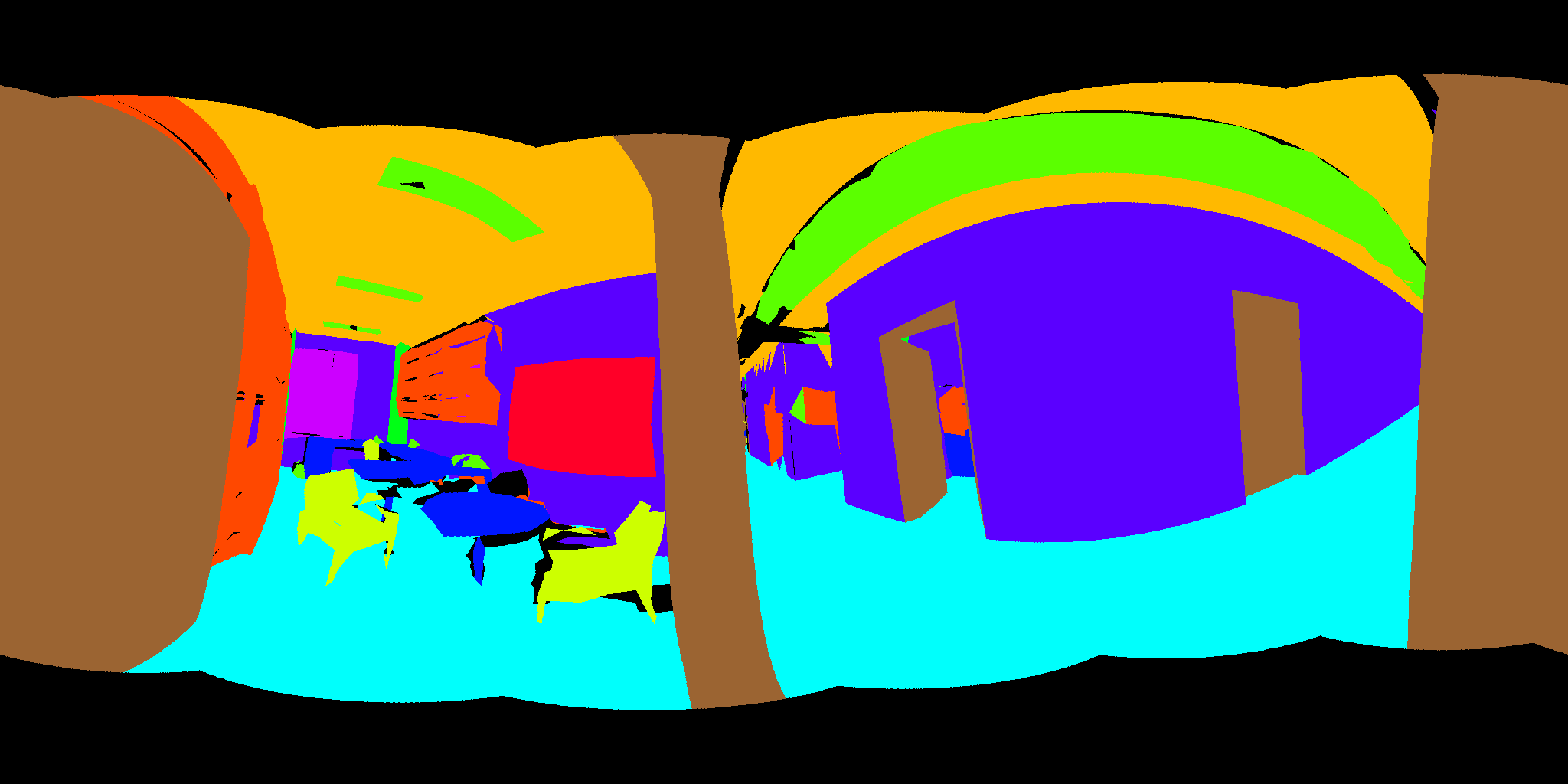}
        \caption{Rotated label}
        \label{sfig:rotated_GT_win}
    \end{subfigure}
    \begin{subfigure}{0.23\linewidth}
        \centering
        \includegraphics[width=1\linewidth]{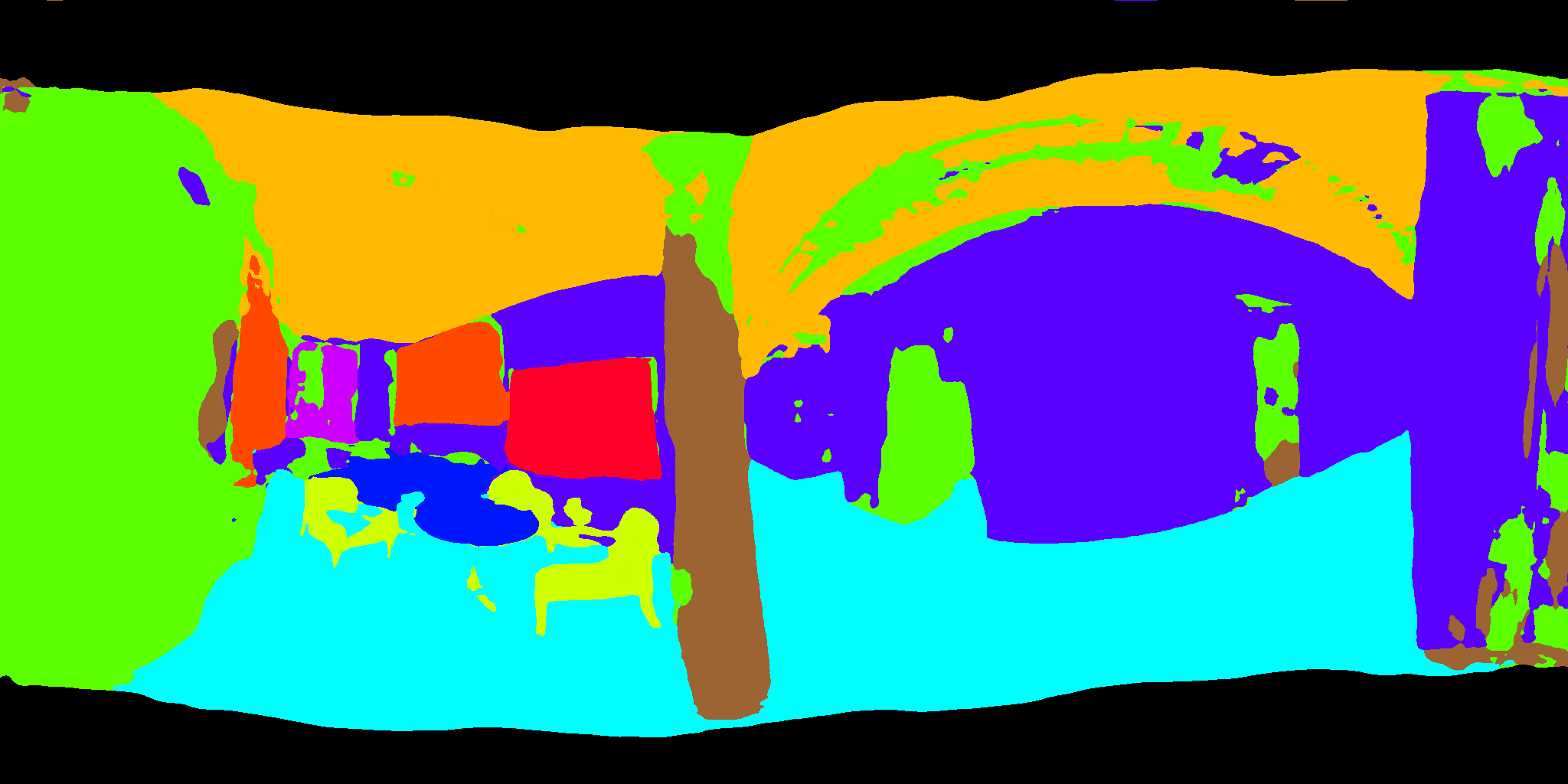}
        \caption{Baseline rotated results}
        \label{sfig:rotated_BL_win}
    \end{subfigure}
    \begin{subfigure}{0.23\linewidth}
        \centering
        \includegraphics[width=1\linewidth]{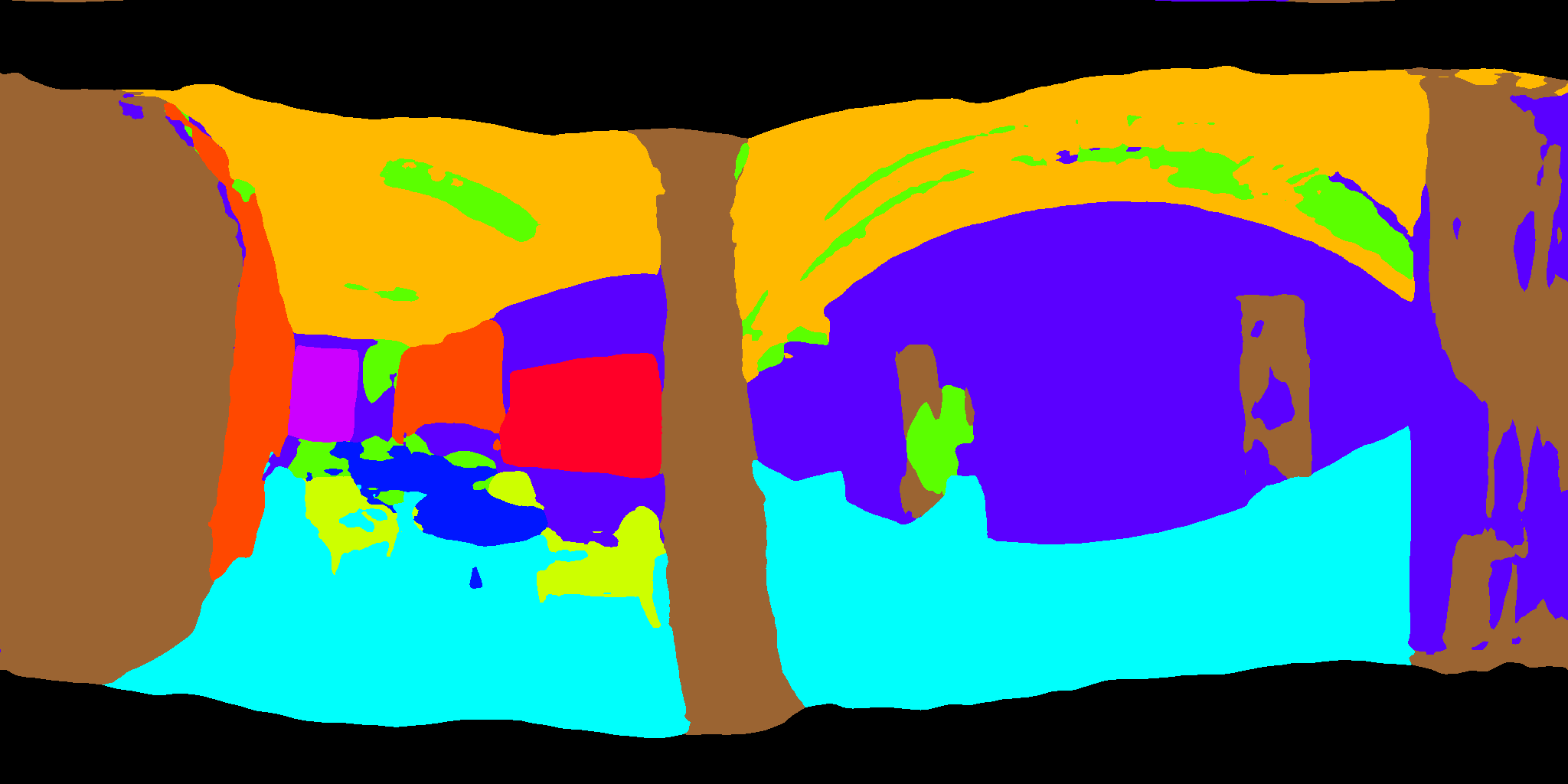}
        \caption{Our rotated results}
        \label{sfig:rotated_ours_win}
    \end{subfigure}
    \caption{
    Visualization comparison of \ours and Trans4PASS+. 
    The rotation of the pitch / roll / yaw axis is $5^{\circ}$ / $5^{\circ}$ / $180^{\circ}$. 
    \ours gains the better results of the semantic class ``windows'' . 
    }
    \label{fig:Visualization_sup_win}
    \centering
\end{figure*}

We show more visualizations of different samples to show the stability improvement of \ours. 
As shown in~\Cref{fig:Visualization_sup_door1}, the separated door in the original image becomes complete in the corresponding rotated image. 
\ours always predict the door in general right when the baseline fails in the corresponding rotated image. 
As shown in~\Cref{fig:Visualization_sup_door2}, the bookcase in the original image is separated in the corresponding rotated image. 
\ours always predict the bookcase right in general as well as the door near it when the baseline is missed in the corresponding rotated image. 
As shown in~\Cref{fig:Visualization_sup_sofa} /~\Cref{fig:Visualization_sup_win}, the location of the sofa / window is different in the original image and in the corresponding rotated image, and \ours has a more stable performance. 

\section{Details for SGA Image Projection}
In this section, we describe $T(\cdot)$ and $R(\cdot, \cdot, \cdot)$ in our SGA image projection in detail. 

Given an ERP-processed input panoramic image $I$ with width = $w$ and height = $h$, consider a point $A_{pan}$ in $I$ with coordinates $(i, j)$.  
$A_{pan}$ corresponds to the point $A$ in the sphere with latitude $A_{lat} = \pi \cdot i / h$ and longitude $A_{lon} = 2 \pi \cdot j / w$. 
Based on $A_{lat}$ and $A_{lon}$, the corresponding 3D vector $v_A$ is obtained. 
For each point of $I$, we obtain its $v_A$, and $V_I$ is the set of all $v_A$. 
We summarize this process as $I = T(V_I)$ and $V_{I} = T^{-1}(I)$. 

For a general rotation in three-dimensional space, the yaw, pitch, and roll angles are $\alpha_{\mathrm{use}}$, $\beta_{\mathrm{use}}$, and $\gamma_{\mathrm{use}}$, respectively. 
The corresponding rotation matrix $R$ can be obtained from these three using matrix multiplication. For example, the product:
\begin{equation}
  \label{eq:R_product}
  \begin{split}
    R(\alpha_{\mathrm{use}}, \beta_{\mathrm{use}}, \gamma_{\mathrm{use}}) &= R_z(\alpha_{\mathrm{use}}) \cdot R_y(\beta_{\mathrm{use}}) \cdot R_x(\gamma_{\mathrm{use}}), 
  \end{split}
\end{equation}
where $R_z(\alpha_{\mathrm{use}})$, $R_y(\beta_{\mathrm{use}})$, $R_x(\gamma_{\mathrm{use}})$ are single-axis 3D rotations and are calculated, respectively, as:
\begin{equation}
  \begin{split}
    \label{eq:3matrix}
      R_z(\alpha_{\mathrm{use}}) &= [[\cos \alpha_{\mathrm{use}}, -\sin \alpha_{\mathrm{use}}, 0], [\sin \alpha_{\mathrm{use}}, \cos \alpha_{\mathrm{use}}, 0], [0,0,1]] \\ 
      R_y(\beta_{\mathrm{use}}) &= [[\cos \beta_{\mathrm{use}}, 0, \sin \beta_{\mathrm{use}}], [0, 1, 0], [-\sin \beta_{\mathrm{use}}, 0, \cos \beta_{\mathrm{use}}]] \\
      R_x(\gamma_{\mathrm{use}}) &= [[1, 0, 0], [0, \cos \gamma_{\mathrm{use}}, -\sin \gamma_{\mathrm{use}}], [0, \sin \gamma_{\mathrm{use}},  \cos \gamma_{\mathrm{use}}]] \\
    \end{split}
  \end{equation}
Based on these operations, we build up our SGA iamge projection, $O_{3D}(I, \alpha_{\mathrm{use}}, \beta_{\mathrm{use}}, \gamma_{\mathrm{use}})$. 

\renewcommand{\arraystretch}{1.5}
\begin{table*}
    \footnotesize
    \centering
    \resizebox{\textwidth}{!}{
        \begin{tabular}{c|c|c|c|c|c|c|c}
        \toprule
        \multirow{2}{*}{($\beta$,$\gamma$,$\alpha$) ($^{\circ}$)} & \multicolumn{1}{l|}{BL mIoU / PAcc}   & \multirow{2}{*}{($\beta$,$\gamma$,$\alpha$) ($^{\circ}$)} & \multicolumn{1}{l|}{BL mIoU / PAcc}   & \multirow{2}{*}{($\beta$,$\gamma$,$\alpha$) ($^{\circ}$)} & \multicolumn{1}{l|}{BL mIoU / PAcc}   & \multirow{2}{*}{($\beta$,$\gamma$,$\alpha$) ($^{\circ}$)} & \multicolumn{1}{l}{BL mIoU / PAcc}   \\ \cline{2-2} \cline{4-4} \cline{6-6} \cline{8-8} 
                                             & \multicolumn{1}{l|}{Our mIoU / PAcc} &                                              & \multicolumn{1}{l|}{Our mIoU / PAcc} &                                              & \multicolumn{1}{l|}{Our mIoU / PAcc} &                                              & \multicolumn{1}{l}{Our mIoU / PAcc} \\ \midrule
        \multirow{2}{*}{(0,0,0)}                     & 53.617 / 81.483                              & \multirow{2}{*}{(0,1,0)}                     & 53.253 / 81.173                              & \multirow{2}{*}{(1,0,0)}                     & 53.436 / 81.231                              & \multirow{2}{*}{(1,1,0)}                     & 53.065 / 80.999                             \\ \cline{2-2} \cline{4-4} \cline{6-6} \cline{8-8} 
                                                    & 56.374 / 83.135                              &                                              & 56.141 / 83.003                              &                                              & 56.286 / 83.043                              &                                              & 56.205 / 83.020                              \\ \hline
        \multirow{2}{*}{(0,0,90)}                    & 53.918 / 81.590                               & \multirow{2}{*}{(0,1,90)}                    & 53.871 / 81.327                              & \multirow{2}{*}{(1,0,90)}                    & 53.506 / 81.267                              & \multirow{2}{*}{(1,1,90)}                    & 53.398 / 81.132                             \\ \cline{2-2} \cline{4-4} \cline{6-6} \cline{8-8} 
                                                    & 56.441 / 83.130                               &                                              & 56.294 / 83.038                              &                                              & 56.246 / 83.002                              &                                              & 56.268 / 83.021                             \\ \hline
        \multirow{2}{*}{(0,0,180)}                   & 53.587 / 81.476                              & \multirow{2}{*}{(0,1,180)}                   & 53.446 / 81.232                              & \multirow{2}{*}{(1,0,180)}                   & 53.559 / 81.218                              & \multirow{2}{*}{(1,1,180)}                   & 53.243 / 81.032                             \\ \cline{2-2} \cline{4-4} \cline{6-6} \cline{8-8} 
                                                    & 56.246 / 83.054                              &                                              & 56.119 / 82.986                              &                                              & 56.111 / 82.976                              &                                              & 56.047 / 83.002                             \\ \hline
        \multirow{2}{*}{(0,0,270)}                   & 53.669 / 81.459                              & \multirow{2}{*}{(0,1,270)}                   & 53.382 / 81.207                              & \multirow{2}{*}{(1,0,270)}                   & 53.557 / 81.194                              & \multirow{2}{*}{(1,1,270)}                   & 53.062 / 81.000                                 \\ \cline{2-2} \cline{4-4} \cline{6-6} \cline{8-8} 
                                                    & 56.223 / 83.051                              &                                              & 56.098 / 82.943                              &                                              & 56.153 / 82.969                              &                                              & 56.141 / 82.961                             \\ \bottomrule
        \end{tabular}
    }
    \caption{
    Detail performance comparison with Tran4PASS+ on Stanford2D3D Panoramic datasets official fold 1 with SGA metrics in $(\beta,\gamma,\alpha) =  (1^{\circ},1^{\circ},360^{\circ})$ setting. 
    This table shows the detailed performance of all 18 situations, and the analysis is in~\cref{tab:STA_SGAM_sup}. 
    ``BL'' means the baseline, Tran4PASS+. 
    ``PAcc'' means the pixel accuracy metric. 
    }
    \label{tab:SGAM}
\end{table*}
\renewcommand{\arraystretch}{1.5}
\begin{table*}
    \footnotesize
    \centering
    \resizebox{\textwidth}{!}{
        \begin{tabular}{c|c|c|c|c|c|c|c}
        \toprule
        \multirow{2}{*}{($\beta$,$\gamma$,$\alpha$) ($^{\circ}$)} & BL mIoU / PAcc   & \multirow{2}{*}{($\beta$,$\gamma$,$\alpha$) ($^{\circ}$)} & BL mIoU / PAcc   & \multirow{2}{*}{($\beta$,$\gamma$,$\alpha$) ($^{\circ}$)} & BL mIoU / PAcc   & \multirow{2}{*}{($\beta$,$\gamma$,$\alpha$) ($^{\circ}$)} & BL mIoU / PAcc   \\ \cline{2-2} \cline{4-4} \cline{6-6} \cline{8-8} 
                                                 & Our mIoU / PAcc &                                              & Our mIoU / PAcc &                                              & Our mIoU / PAcc &                                              & Our mIoU / PAcc \\ \midrule
        \multirow{2}{*}{(0,0,0)}   & 53.617/81.483 & \multirow{2}{*}{(0,3,0)}   & 51.632/79.878 & \multirow{2}{*}{(3,0,0)}   & 51.59/79.966  & \multirow{2}{*}{(3,3,0)}   & 50.172/78.864 \\  \cline{2-2} \cline{4-4} \cline{6-6} \cline{8-8} 
                                & 56.374/83.135 &                            & 56.105/82.98  &                            & 56.11/82.994  &                            & 56.021/82.952 \\ \hline
        \multirow{2}{*}{(0,0,90)}  & 53.918/81.59  & \multirow{2}{*}{(0,3,90)}  & 52.126/80.178 & \multirow{2}{*}{(3,0,90)}  & 51.804/79.946 & \multirow{2}{*}{(3,3,90)}  & 50.577/79.084 \\ \cline{2-2} \cline{4-4} \cline{6-6} \cline{8-8} 
                                & 56.441/83.13  &                            & 56.089/82.955 &                            & 56.173/82.982 &                            & 56.014/82.915 \\ \hline
        \multirow{2}{*}{(0,0,180)} & 53.587/81.476 & \multirow{2}{*}{(0,3,180)} & 51.543/79.817 & \multirow{2}{*}{(3,0,180)} & 51.777/80.09  & \multirow{2}{*}{(3,3,180)} & 50.424/79.002 \\ \cline{2-2} \cline{4-4} \cline{6-6} \cline{8-8} 
                                & 56.246/83.054 &                            & 56.05/82.965  &                            & 55.861/82.88  &                            & 55.838/82.86  \\ \hline
        \multirow{2}{*}{(0,0,270)} & 53.669/81.459 & \multirow{2}{*}{(0,3,270)} & 51.622/79.914 & \multirow{2}{*}{(3,0,270)} & 51.598/79.757 & \multirow{2}{*}{(3,3,270)} & 50.131/79.039 \\ \cline{2-2} \cline{4-4} \cline{6-6} \cline{8-8} 
                                & 56.223/83.051 &                            & 55.972/82.876 &                            & 56.088/82.938 &                            & 55.946/82.877 \\ \bottomrule
        \end{tabular}
    }
    \caption{
    Detail performance comparison with Tran4PASS+ on Stanford2D3D Panoramic datasets official fold 1 with SGA metrics in $(\beta,\gamma,\alpha) =  (3^{\circ},3^{\circ},360^{\circ})$ setting. 
    This table shows the detailed performance of all 18 situations, and the analysis is in~\cref{tab:STA_SGAM_sup}. 
    ``BL'' means the baseline, Tran4PASS+. 
    ``PAcc'' means the pixel accuracy metric. 
    }
    \label{tab:SGAMthree}
\end{table*}

\renewcommand{\arraystretch}{1.5}
\begin{table*}
    \footnotesize
    \centering
    \resizebox{\textwidth}{!}{
        \begin{tabular}{c|c|c|c|c|c|c|c}
        \toprule
        \multirow{2}{*}{($\beta$,$\gamma$,$\alpha$) ($^{\circ}$)} & BL mIoU / PAcc   & \multirow{2}{*}{($\beta$,$\gamma$,$\alpha$) ($^{\circ}$)} & BL mIoU / PAcc   & \multirow{2}{*}{($\beta$,$\gamma$,$\alpha$) ($^{\circ}$)} & BL mIoU / PAcc   & \multirow{2}{*}{($\beta$,$\gamma$,$\alpha$) ($^{\circ}$)} & BL mIoU / PAcc   \\ \cline{2-2} \cline{4-4} \cline{6-6} \cline{8-8} 
                                                 & Our mIoU / PAcc &                                              & Our mIoU / PAcc &                                              & Our mIoU / PAcc &                                              & Our mIoU / PAcc \\ \midrule
        \multirow{2}{*}{(0,0,0)}   & 53.617/81.483 & \multirow{2}{*}{(0,10,0)}   & 45.823/76.249 & \multirow{2}{*}{(10,0,0)}   & 45.408/76.165 & \multirow{2}{*}{(10,10,0)}   & 43.383/74.803 \\  \cline{2-2} \cline{4-4} \cline{6-6} \cline{8-8} 
                                & 56.374/83.135 &                             & 55.317/82.461 &                             & 55.507/82.572 &                              & 54.557/81.99  \\ \hline
        \multirow{2}{*}{(0,0,90)}  & 53.918/81.59  & \multirow{2}{*}{(0,10,90)}  & 46.105/76.338 & \multirow{2}{*}{(10,0,90)}  & 45.868/76.293 & \multirow{2}{*}{(10,10,90)}  & 44.017/75.222 \\ \cline{2-2} \cline{4-4} \cline{6-6} \cline{8-8} 
                                & 56.441/83.13  &                             & 55.286/82.502 &                             & 55.555/82.525 &                              & 54.57/81.965  \\ \hline
        \multirow{2}{*}{(0,0,180)} & 53.587/81.476 & \multirow{2}{*}{(0,10,180)} & 46.234/76.513 & \multirow{2}{*}{(10,0,180)} & 45.847/76.309 & \multirow{2}{*}{(10,10,180)} & 44.054/75.105 \\ \cline{2-2} \cline{4-4} \cline{6-6} \cline{8-8} 
                                & 56.246/83.054 &                             & 55.571/82.588 &                             & 55.175/82.468 &                              & 54.595/82.03  \\ \hline
        \multirow{2}{*}{(0,0,270)} & 53.669/81.459 & \multirow{2}{*}{(0,10,270)} & 45.462/76.125 & \multirow{2}{*}{(10,0,270)} & 46.242/76.456 & \multirow{2}{*}{(10,10,270)} & 43.84/75.251  \\ \cline{2-2} \cline{4-4} \cline{6-6} \cline{8-8} 
                                & 56.223/83.051 &                             & 55.451/82.448 &                             & 55.525/82.554 &                              & 54.724/82.02  \\ \bottomrule
        \end{tabular}
    }
    \caption{
    Detail performance comparison with Tran4PASS+ on Stanford2D3D Panoramic datasets official fold 1 with SGA metrics in $(\beta,\gamma,\alpha) =  (10^{\circ},10^{\circ},360^{\circ})$ setting. 
    This table shows the detailed performance of all 18 situations, and the analysis is in~\cref{tab:STA_SGAM_sup}. 
    ``BL'' means the baseline, Tran4PASS+. 
    ``PAcc'' means the pixel accuracy metric. 
    }
    \label{tab:SGAMten}
\end{table*}


\ifCLASSOPTIONcaptionsoff
  \newpage
\fi